\documentclass[10pt]{article} % For LaTeX2e
\usepackage[preprint]{tmlr}

\usepackage[ruled,vlined,linesnumbered]{algorithm2e}
\renewcommand{\cite}[1]{\citep{#1}}

\usepackage[utf8]{inputenc} % allow utf-8 input
\usepackage[T1]{fontenc}    % use 8-bit T1 fonts
\usepackage[bookmarksnumbered=true]{hyperref}       % hyperlinks
\usepackage{url}            % simple URL typesetting
\usepackage{booktabs}       % professional-quality tables
\usepackage{amsfonts}       % blackboard math symbols
\usepackage{nicefrac}       % compact symbols for 1/2, etc.
\usepackage{microtype}      % microtypography
\usepackage{xcolor}         % colors

\usepackage{mathtools}
\usepackage{amsmath}
\usepackage{amssymb}
\usepackage{commath}
\usepackage[inkscapelatex=false]{svg}

\usepackage[percent]{overpic}
\usepackage{sidecap}
\usepackage{comment}

\usepackage[toc,page,header]{appendix}
\usepackage{minitoc}

\usepackage{amsthm}
\usepackage{mathrsfs}
\usepackage{enumitem}
\usepackage[ruled,vlined,linesnumbered]{algorithm2e}

\usepackage{wrapfig}

\usepackage{microtype}
\usepackage{graphicx}
\usepackage{booktabs} % for professional tables
\usepackage{multirow}
\usepackage{longtable} % added by KAZU images
\usepackage{subcaption}
\usepackage{xcolor}         % colors

\usepackage{tabularx}
\usepackage{array}
\usepackage{makecell} % 改行しやすくする
\newtheorem{theorem}{Theorem} 
\newtheorem{lemma}{Lemma}
\newtheorem{proposition}{Proposition}

\theoremstyle{definition}
\newtheorem{definition}{Definition}

\theoremstyle{remark}

\usepackage[many]{tcolorbox}


\newtcolorbox{graybox}{
  colback=gray!10,    % 背景色（gray!10 は薄いグレー）
  boxrule=0pt,        % 枠線なし
  colframe=gray!10,   % 枠色＝背景（見えなくする）
  left=6pt, right=6pt, top=6pt, bottom=6pt, % 内側余白
  sharp corners,      % 角を丸めない
  breakable           % ページまたぎOK
}
\DeclareMathOperator*{\argmin}{arg\,min}

\DeclareMathAlphabet{\mathpzc}{OT1}{pzc}{m}{it}

\newcommand{\reg}{\varepsilon}

\newcommand{\RR}{\mathbb{R}}

\newcommand{\tens}[1]{\mathbf{#1}}

\newcommand{\tP}{\tens{P}}
\newcommand{\tQ}{\tens{Q}}
\newcommand{\tR}{\tens{R}}

\newcommand{\mats}[1]{\mathbf{#1}}
\newcommand{\mQ}{\mats{Q}}

\newcommand{\mP}{\mats{P}}
\newcommand{\mK}{\mats{K}}
\newcommand{\mR}{\mats{R}}
\newcommand{\mM}{\mats{M}}
\newcommand{\mW}{\mats{W}}
\newcommand{\cost}{\mM}

\newcommand{\ba}{\boldsymbol{a}}
\newcommand{\bb}{\boldsymbol{b}}

\newcommand{\bx}{\boldsymbol{x}} 
\newcommand{\by}{\boldsymbol{y}} 
\newcommand{\bz}{\boldsymbol{z}} 
\newcommand{\bmtheta}{\boldsymbol{\theta}} 
\newcommand{\bmeta}{\boldsymbol{\eta}} 

\newcommand{\FIM}{\boldsymbol{\mathbf{G}}}

\newcommand{\mani}[1]{\mathscr{#1}}

\newcommand{\family}{\mani{H}}
\newcommand{\subfamily}{\mani{E}}

\DeclareMathOperator{\Exp}{Exp}
\DeclareMathOperator{\Log}{Log}
\newcommand{\qExp}[1]{\Exp_{q}\left\lbrack #1 \right\rbrack }
\newcommand{\mExp}[1]{\Exp_{m}\left\lbrack #1 \right\rbrack }
\newcommand{\kExp}[1]{\Exp_{\kappa}\left\lbrack #1 \right\rbrack }
\newcommand{\kExpp}[1]{\Exp^{\prime}_{\kappa}\left\lbrack #1 \right\rbrack }
\newcommand{\cExp}[1]{\Exp_{\chi}\left\lbrack #1 \right\rbrack}
\newcommand{\cExpp}[1]{\Exp^{\prime}_{\chi}\left\lbrack #1 \right\rbrack }
\newcommand{\cExppp}[1]{\Exp^{\prime\prime}_{\chi}\left\lbrack #1 \right\rbrack }
\newcommand{\mExppp}[1]{\Exp^{\prime\prime}_{m}\left\lbrack #1 \right\rbrack }
\newcommand{\qExppp}[1]{\Exp^{\prime\prime}_{q}\left\lbrack #1 \right\rbrack }
\newcommand{\qLog}[1]{\Log_{q}\left\lbrack #1 \right\rbrack }
\newcommand{\mLog}[1]{\Log_{m}\left\lbrack #1 \right\rbrack }
\newcommand{\kLog}[1]{\Log_{\kappa}\left\lbrack #1 \right\rbrack }
\newcommand{\cLog}[1]{\Log_{\chi}\left\lbrack #1 \right\rbrack }

\newcommand{\invcLog}[1]{\Log_{\chi}^{-1}\left\lbrack #1 \right\rbrack }
\newcommand{\cEsc}[1]{\Tilde{\chi}\left\lbrack #1 \right\rbrack }
\newcommand{\mEsc}[1]{\Tilde{\chi}_m\left\lbrack #1 \right\rbrack }
\newcommand{\qEsc}[1]{\Tilde{\chi}_q\left\lbrack #1 \right\rbrack }
\newcommand{\Esc}[1]{\chi\left\lbrack #1 \right\rbrack }

\newcommand{\Escp}[1]{\chi^{\prime}\left\lbrack #1 \right\rbrack }
\newcommand{\invEsc}[1]{\Tilde{\chi}^{-1}\left\lbrack #1 \right\rbrack }
\newcommand{\chif}[1]{\chi\left\lbrack #1 \right\rbrack }
\newcommand{\chifp}[1]{\chi^{\prime}\left\lbrack #1 \right\rbrack }

\newcommand{\dtimes}{\circledast}
\newcommand{\inner}[2]{\langle #1, #2\rangle}
\newcommand{\csimplex}[2]{\Tilde{\boldsymbol{\Pi}}_{\chi}(#1, #2)}
\newcommand{\simplex}[2]{{\boldsymbol{\Pi}}(#1, #2)}
\newcommand{\dualdomain}[1]{\hat{\boldsymbol{\Pi}}_{\chi}(#1)}

\newcommand{\GK}[1]{\textcolor{black}{#1}}

\title{Escort Entropic Optimal Transport\\for Controllable Tail Decay}

\author{\name Kazu Ghalamkari \email kazfu@dtu.dk \\
      \addr Department of Applied Mathematics and Computer Science\\
      Technical University of Denmark
            \AND
    \name Petr Taborsky \email petr@dtu.dk \\
\addr Department of Applied Mathematics and Computer Science \\
Technical University of Denmark
      \AND
      \name Morten Mørup \email mmor@dtu.dk\\
      \addr Department of Applied Mathematics and Computer Science \\
      Technical University of Denmark
}

\def\month{4}  % Insert correct month for camera-ready version
\def\year{2026} % Insert correct year for camera-ready version
\def\openreview{\url{https://openreview.net/forum?id=954CjhXSXL}} % Insert correct link to OpenReview for camera-ready version

\begin{document}

%%%%%%%%% TOC %%%%%%%%%
\doparttoc % Tell to minitoc to generate a toc for the parts
\faketableofcontents % Run a fake tableofcontents command for the partocs
%\part{} % Start the document part
%%%%%%%%%%%%%%%%%%%%%%%%

\maketitle
\begin{abstract}

We discuss the flexible decay tunability of the optimal transport plan brought by escort-entropic regularization. Existing variations of escort entropy allow only sub-exponential decay or introduce a cutoff function in the functional form of the optimal plan, which may destabilize the barycentric projection, a widely used downstream task of optimal transport. To balance further flexibility in the decay tunability and noise robustness in the barycentric projection, we introduce an alternative variant of the escort entropic regularization that allows both sub- and super-exponential decay without a cutoff function. Our first-order perturbation analysis qualitatively reveals the mechanism by which the escort-entropic regularization induces noise robustness and local adaptability of the transport plan. Furthermore, we develop a unified algorithm for various escort-entropic regularization and demonstrate the usefulness of decay tunability in numerical experiments. 
\end{abstract}
\section{Introduction}\label{sec:intro}

Regularized optimal transport aims to find a transport plan $\mR$ that minimizes 
%\begin{align*}
   $\inner{\cost}{\mR} + \reg\Omega(\mR)$
%\end{align*}
subject to the marginal constraints $\sum_j \mR_{ij}=a_i$ and $\sum_i \mR_{ij}=b_j$ under the cost matrix $\cost$, appropriate regularizer $\Omega(\mR)$, and imposed regularization strength $\reg >0$. Optimal transport has become an indispensable tool that has evolved beyond its original motivation in logistics and infrastructure and is now widely used in machine learning~\citep{peyre2019computational}, computer vision~\citep{Bonneel2023survey}, and bioinformatics~\citep{klein2023mapping,yachimura2024scegot}.
%Optimal transport has become an indispensable tool that  beyond its motivational use in logistics and infrastructure today serves important applications also within machine learning~\citep{peyre2019computational}, computer vision~\citep{Bonneel2023survey}, and Bioinformatics~\citep{klein2023mapping, yachimura2024scegot}. 
The regularizer $\Omega(\mR)$ characterizes both the resulting plan and the algorithm that computes it. For instance, the negative Shannon entropy
%It is well-studied that the regularizer $\Omega(\mR)$ controls a property of the transport plan $\mR$ and/or associated algorithms, e.g., 
%the $\mathrm{L}^2$ norm $\Omega(\mR)=\sum_{ij}\mR_{ij}^2$ guarantees the uniqueness of the optimal plan, 
%the negative Shannon's entropy 
$\Omega(\mR)=\sum_{ij}\mR_{ij}\log{\mR_{ij}}$ and particular types of $f$-divergence enable efficient Sinkhorn updates~\citep{Cuturi2013Sinkhorn, Optimal2022Terj}, deformed $q$-entropy $\Omega(\mR)\propto\sum_{ij}\mR_{ij}^q$ controls the sparsity of the plan~\citep{Bao2022Sparse}, and mutual information offers robust transport plans to outliers~\citep{infoOT2023Chuang}.

We investigate the practical benefits of appropriately designed regularization that controls the decay of the entries of the transport plan $\mR_{ij}$ as the corresponding transport costs $\cost_{ij}$ increase.
%We investigate practical benefits obtained by appropriately designed regularization that controls the decay profile of the transport plan with respect to the cost matrix $\cost$. 
%the decay profile of the entries of the transport plan against the cost matrix $\cost$. 
%when sorted in decreasing order. 
%Inspired by the well-studied robustness induced by tunable distributional decay in statistical models, we investigate whether analogous benefits can be achieved in optimal transport by tuning the decay profile of transport-plan entries sorted in decreasing order. 
%Although the advantages of the various kinds of distributional decay in statistical models have been well discussed in the machine learning community, the decay of the transport plans has not been systematically discussed yet. 
%For this purpose, we focus on the $\chi$-escort entropic optimal transport ($\chi$-ET) developed by~\citet{kurose2022optimal}, where the escort entropic regularization $\varphi_\chi$, generated by a monotonically increasing function $\chi:\RR_{>0}\rightarrow\RR_{>0}$ (denoted the escort function), transforms the transport plan into a generalized exponential form. Appropriate choice of the escort function $\chi$ enables explicit control of the decay of the functional form of the transport plan beyond the well-discussed exponential decay $\mR_{ij}\propto \exp{[-\cost_{ij}]}$, induced by the negative Shannon's entropy. 
For this purpose, we focus on the $\chi$-escort entropic optimal transport ($\chi$-ET) developed by~\citet{kurose2022optimal}. Its regularizer $\varphi_\chi$ is generated by a monotonically increasing function $\chi:\RR_{>0}\rightarrow\RR_{>0}$, referred to as the \emph{escort function}, and transforms the functional form of the transport plan into a derivative of the generalized exponential form. Consequently, an appropriate choice of the escort function $\chi$ enables explicit control over the decay of the transport plan beyond the well-studied exponential decay $\mR_{ij}\propto \exp{[-\cost_{ij}]}$, induced by negative Shannon entropy~\citep{Benamou2015Iterative}. 
Notably, the mathematical work by~\citet{kurose2022optimal} aims to reveal the dually flat structure underlying optimal transport in the context of information geometry. Accordingly, practical implementations such as feasible algorithms and principles for designing escort functions, and the benefits of controlling the decay of the optimal transport plan remain unexplored. 
Thus, their pioneering work is, in our view, underappreciated in the machine learning community despite its potential for practical usefulness with flexible decay tuning for the optimal plan.
%Nevertheless, given its strong flexibility of the decay of the optimal transport plan, we believe that their pioneering work deserves greater attention from the machine learning community.

\begin{figure*}[t]
  \centering
  %\includesvg[width=0.99\linewidth]{figs/headline.svg}
  \includegraphics[width=0.99\linewidth]{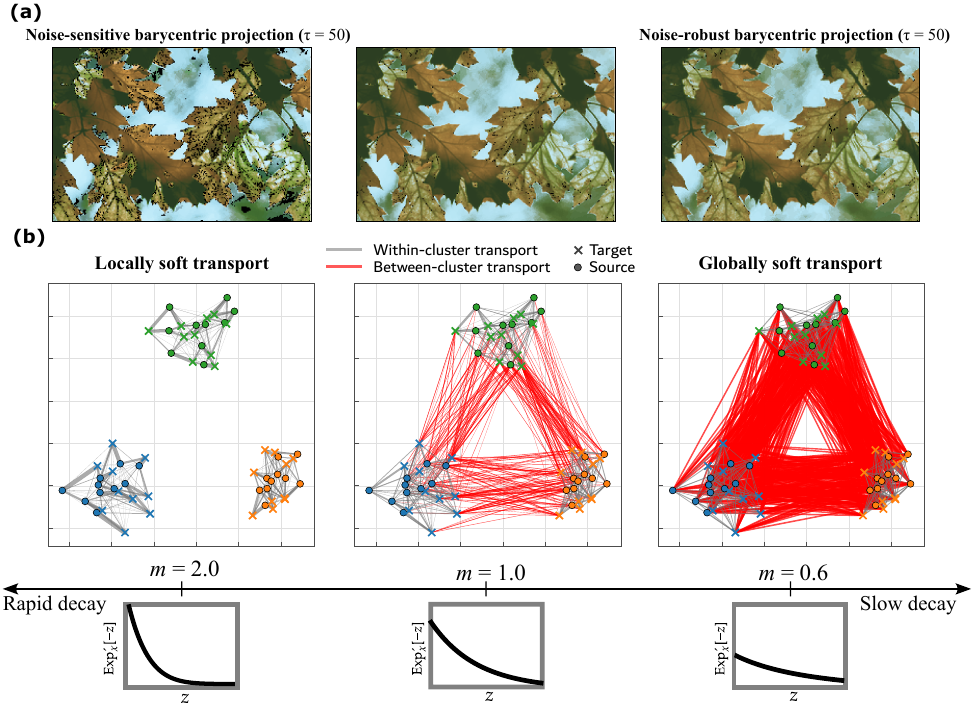}
  \caption{%Noise robustness by providing a globally soft as opposed to locally soft transportation plan. 
  %\GK{A single hyperparameter $m$ of the proposed Weibull-ET controls the tail of the derivative of the corresponding generalized exponential $\mExp{\ \cdot \ }$, which leads to a flexible decay property of the transport plan from rapid ($m>1$) to slow ($m<1$) decay. }
  A single hyperparameter $m$ of the proposed Weibull-ET controls the \GK{tail} of the functional form of the transport plan from rapid ($m>1$) to slow ($m<1$) decay. 
  This control allows us to adjust the robustness of the resulting barycentric projection to noise and the balance between local and global softness of the transport plan. Detailed experimental setup and the discussion of these results can be found in Section~\ref{sec:experiments}. (\textbf{a}) Color transported images by barycentric projection with Dirichlet noise ($\tau=50$), which demonstrates improved noise robustness when $m<1$. The value of $\reg$ is tuned to minimize the centroid drift shift $\Delta^{\tau}_{m\mathrm{-ET}}$. (\textbf{b}) The $m$-value dependency of the transportation from the three synthesized clustered sources to the targets, fixing $\reg=0.07$. Gray and red lines represent intra- and inter-cluster transport paths, respectively, and line width corresponds to the amount of transported mass. A larger $m>1$ prevents leakage of mass into inter-cluster transport while maintaining the softness of transport within the cluster. %For clarity, only the top 50\% of the transported mass is shown. 
  %Details of the experimental setup and quantitative evaluation when varying both $m$ and $\reg$ are provided in Section~\ref{sec:experiments}.
  }
  \label{fig:headline}
\end{figure*}

We presently fill this gap between the principled information-geometric representation of optimal transport and its practical implications. 
First, based on the dually flat structure of the feasible set discovered by~\citet{kurose2022optimal}, we present a practical algorithm to solve $\chi$-ET as a Bregman divergence minimization problem. 
%, which ensures global optimality. 
%without a cutoff function, while optimal transports with existing escort functions cover only the slow decay regime or include a cutoff function. % while the naive implementation of the Tsallis-ET, an instance of $\chi$-ET discussed in~\citep{kurose2022optimal}, is infeasible due to its domain issues, which has not been pointed out clearly before in the context of the optimal transport. 
%Second, we investigate the practical benefits of tuning the decay profile of the optimal plan. 
Second, we investigate how the escort function brings 
practical benefits through a tunable decay profile of the optimal plan in $\chi$-ET. 
Our first-order perturbation analysis qualitatively reveals that an appropriately tuned decay of the optimal plan may partially cancel the effect of perturbation in marginals by averaging, thereby stabilizing the barycentric projection, a fundamental downstream task of optimal transport. 
In addition to the averaging effect, we also discuss the selective effect of rapid decay, which promotes adaptive softness, bringing flexibility to the optimal plan, such as local softness. 
%preventing transport between clusters but allowing soft transport within clusters.
%Our first-order perturbation analysis shows how the escort function modulates the response of the barycentric projection, a fundamental downstream task in optimal transport, to perturbations in the input marginals. More specifically, we show qualitatively that an appropriately tuned decay of the optimal plan can facilitate partial cancellation of the perturbation by averaging over transport paths, thereby stabilizing the barycentric projection. 
%More specifically, we show qualitatively that an appropriately tuned decay of the optimal plan facilitates reducing the perturbation effects by averaging over transport paths, resulting in partial cancellation of the perturbation also stabilizing the barycentric projection. 

While the favorable decay profile depends on the downstream task, the data, and the magnitude of the perturbation, existing regularizers, including $\chi$-ET with the existing escort functions, yield exponential or slower decay unless they introduce a cutoff function, which sets the entries beyond a cost threshold exactly to zero and may limit the averaging. Thus, as a third contribution, we introduce a novel escort function $\chi_m$, whose hyperparameter $m$ continuously controls the decay of the transport plan from rapid decay (faster than exponential) to slow decay (slower than exponential) without a cutoff function.

%Although how much averaging is favorable depends on the data and the magnitude of the perturbation, existing regularizers, including $\chi$-ET with the existing escort functions, yield only slower decay than exponential and decay faster than exponential is reached only through a cutoff function, which sets the entries beyond a cost threshold exactly to zero and may limit the averaging. To cover both sides, as a third contribution, we introduce a novel escort function $\chi_m$, whose hyperparameter $m$ continuously controls the decay of the transport plan from rapid decay (faster than exponential) to slow decay (slower than exponential) without a cutoff function.
%However, existing approaches, including $\chi$-ET with existing escort functions, often introduce a cutoff function into the functional form of the optimal plan, which may limit the averaging and thereby produce unstable barycentric projections. Thus, as a third contribution, we introduce a novel escort function $\chi_m$, whose hyperparameter $m$ continuously controls the decay of the transport plan, from rapid decay (super-exponential tail) to slow decay (sub-exponential tail) without a cutoff function. 

%, while our proposed $\chi_m$ enables decay tuning without cutoff function. 

We empirically verify that $\chi$-ET with our novel escort function brings the above benefits in color transformation by demonstrating the stability of the barycentric projections in a noisy setting (Figure~\ref{fig:headline}(a)) and in synthetic data by demonstrating the controllable adaptive softness (Figure~\ref{fig:headline}(b)), and we also confirm the decay profile of the obtained optimal plan. Finally, we demonstrate that the ordinary entropic regularization cannot bring these benefits, due to its fixed decay ratio, even when we tune the strength of the regularization $\reg$. 

The remainder of the paper is organized as follows. We describe related work in Section~\ref{sec:relatework}, and subsequently review general escort functions $\chi$ and introduce the proposed escort function $\chi_m$ together with the associated generalized exponential functions in Section~\ref{sec:escort}. The problem setup is summarized in Section~\ref{sec:ET} and we present an algorithm that solves $\chi$-ET in Section~\ref{sec:algorithm}. We proceed and discuss the decay property of $\chi$-ET in Section~\ref{sec:tail} whereas section~\ref{sec:optim} provides the derivation of the algorithm introduced in Section~\ref{sec:algorithm}, following a summary of the necessary background within information geometry. Finally, Section~\ref{sec:experiments} presents numerical evaluations that demonstrate the benefit of the flexible decay of $\chi$-ET. We conclude the paper in  Section~\ref{sec:conclude}.

\textbf{Terminological remark \ \ } We define a transport plan as having \emph{full-support} if all its entries are strictly positive. 
The decay of a plan refers to how its entries decrease as the transport cost increases. Throughout the paper, rapid and slow decay refer to decays respectively faster and slower than exponential decay. Decay shapes with a cutoff function are distinguished from both rapid and slow decay. 
%The decay of the plan $\mR\in\RR^{I\times J}$ indicates the tail shape of a sequence of all entries in $\mR$ sorted in descending order, i.e., $(\mR_{(1)},\mR_{(2)},\dots,\mR_{(IJ)})$ where $\mR_{(\ell)}$ indicates the $\ell$-th largest entry in $\mR$. Throughout the paper, rapid and slow decay refer to decays respectively faster and slower than exponential decay. Tail shapes with a cutoff function are distinguished both from rapid and slow decay. 
Following the existing literature, we say that a regularizer $\Omega(\mR)$ has a \emph{boundary barrier} if its derivative diverges as an entry $\mR_{ij}$ approaches the boundary of its domain, i.e., $\mR_{ij}\rightarrow0$. This property discourages boundary solutions and leads to transport plans in the relative interior of the transport polytope~\citep{rockafellar1997convex}.

\begin{table}[t]
\centering
\caption{Regularized optimal transport methods with their scope of decay of plans}
\label{tab:comparison}
\renewcommand{\arraystretch}{1.1} 
\setlength{\tabcolsep}{5pt}         
\begin{tabular}{@{}lccc@{}}
\toprule
Method & $\Omega(\mR)$ & Global opt. & Decay profile\\
\midrule
\textbf{Weibull-ET (ours)}  & $\varphi_m\left(\tilde\chi_m^{-1}[\mR]\right)$                & Yes & Rapid ($m>1$) \& Slow ($m<1$) \\
Tsallis-ET                  & $\varphi_q \left(\tilde\chi_q^{-1}[\mR]\right)$               & Yes & Slow ($q>1$) \&  Cutoff ($q<1$) \\
Kaniadakis-ET               & $\varphi_\kappa \left(\tilde\chi_\kappa^{-1}[\mR]\right)$     & Yes & Slow \\
Sinkhorn~\citep{Cuturi2013Sinkhorn}               & $\sum_{ij}\mR_{ij}\log{\mR_{ij}}$                    & Yes & Exponential \\
L$^2$-OT~\citep{lorenz2021quadratically}      & $\sum_{ij}\mR_{ij}^2$                                     & Yes & Cutoff \\
$q$-DOT~\citep{Bao2022Sparse}                  & $\sum_{ij}\mR_{ij}\qLog{\mR_{ij}}$                          & Yes & Cutoff \\
TEM~\citep{Amid2024optimal}                   & $D_t(\tilde\mR , \ba\bb^\top)$              & No  & Cutoff \\
TROT~\citep{Muzellec2017Tsallis}       & $\sum_{ij}\mR_{ij}^q/(1-q)$                                  & No in cutoff regime & Slow \& Cutoff \\
EMD~\citep{kantorovich1942translocation}                        & $0$                       & Yes & Sparse \\
\bottomrule
\end{tabular}
\end{table}

\subsection{Related work}\label{sec:relatework}

We note that existing deformed $q$-entropic optimal transport ($q$-DOT)~\citep{Bao2022Sparse} and $\mathrm{L}^2$-regularized optimal transport provide optimal plans with a cutoff rather than a rapid decay with full support. While Tsallis regularized OT~(TROT)~\citep{Muzellec2017Tsallis} and Tempered Exponential-based OT~(TEM)~\citep{Amid2024optimal} both support slow decay, the gradient of the TROT objective is not Lipschitz continuous; thus, standard convergence guarantees do not directly apply, and the feasible set of TEM is not convex; thus, the global optimality of its solution is not guaranteed. 
%the objective of TROT is not Lipschitz, and the feasible set in TEM is not convex, and consequently neither guarantees global optimality. 
While existing $q$-DOT and L$^2$-OT also enhance local softness leveraging their cutoff function, these methods are inherently unable to provide a rapid decay plan, and the advantage of our proposed Weibull-ET is that a single hyperparameter $m$ enables covering from rapid to slow decay while preserving global optimization and full support without the use of a cutoff function. 
%Our contribution of novel escort offers flexible tuning from lighter and heavier 

Technically, when a boundary barrier exists, i.e., the derivative of the regularization diverges at the boundary of the support, the optimal plan stays internal to the transport polytope, and the complementarity condition of the Karush-Kuhn-Tucker (KKT) conditions~\citep{kuhn1951nonlinear} would not be activated, resulting in a cutoff that does not appear; however, such a boundary barrier destroys the Lipschitz continuity of the derivative of the objective function, so the standard convergence guarantees of gradient-based methods in the optimal plan do not directly apply. In contrast, when the boundary barrier is absent, the optimal plan on the domain boundary may lead to a cutoff function while the Lipschitz continuity of the derivative of the objective function holds. 
Our method avoids this trade-off by updating not the transport plan itself, but rather an escorted plan. This corresponds to the dual formulation of $\chi$-ET, which yields a smooth objective as is standard in convex analysis~\cite{Blondel2018Smooth}, thereby achieving both guaranteed convergence to the global optimum and full support without cutoff. The above-mentioned existing methods are summarized in Table~\ref{tab:comparison}.

Although unbalanced optimal transport~\citep{Benamou2003} achieves robustness to noisy mass histograms by relaxing the marginalization constraints using additional hyperparameters, the decay tunability of $\chi$-ET offers robustness in downstream barycentric projections without compromising the strict feasibility of the marginalization constraints. %This setting is relevant when the observed histograms must be treated as prescribed constraints, but small perturbations of them should not cause drastic changes in the transport structure, as in supply--demand matching over time. %, histogram matching, and barycentric projection with fixed marginals.
%While \citet{Optimal2022Terj} provide systematic analyses of $f$-divergence regularization in optimal transport, we reformulate escort-entropic optimal transport as a Bregman divergence minimization. The reformulating optimally transports as a Bregman divergence minimization enables the shape of the optimal plan~\citep{Arnaud2018Regularized}. 
%
%While \citet{Optimal2022Terj} provide systematic analyses of 
%$f$-divergence regularization in optimal transport, we reformulate escort-entropic optimal transport as a Bregman divergence minimization, a reduction established for smooth convex regularizers by \citet{Arnaud2018Regularized}, who also observe that the choice of regularizer changes the shape of the optimal plan; however, none of these works explicitly yield a plan that is both full-support and tunable from slow to rapid decay. 
%
While the $\chi$-ET is formulated as a $\chi$-divergence minimization, a related relaxation of regularized optimal transport into a Bregman divergence minimization problem is systematically discussed by~\citet{Dessein2018Regularized}. However, their alternate scaling algorithm relies on the separability of the regularizer, which the escort entropic regularizer does not guarantee to satisfy.

\section{Methods}\label{sec:method}
%Since the proposed framework transports probablity mass functions by escort distirbution, we review the escort and related deformed exponential and logarithm functions in Section~\ref{sec:escort}. Then, we define the $\chi$-entropic escort transport ($\chi$-ET) in Section~\ref{sec:escort}, following the proposed algorithm to exactly solve it, which is our main result. The derivation of the algorithm is presented in Section~\ref{sec:optim}. 
As is common in the literature~\citep{Cuturi2013Sinkhorn, Frogner2015Learning, Amari2019Information, Ghalamkari2022Fast}, we identify the discrete joint distribution $p(x, y)$ on the sample space $\mathcal{X}=[I]\times[J]$ as a non-negative normalized matrix $\tP$ where each entry $\mP_{ij}$ corresponds to $p(x=i,y=j)$. 
For a positive integer $n$, let $[n]$ be the set of natural numbers up to $n$, that is, $[n]=\{1,2,\dots,n\}$. The symbols $\mathbb{R}$ and $\mathbb{R}_{> 0}$ denote the sets of real numbers and positive real numbers, respectively. For a function $f$, we write $f^\prime$ and $f^{\prime\prime}$ for its first and second derivatives, respectively.
%We also denote by $\mathbb{R}$ and $\mathbb{R}_{> 0}$ the sets of real numbers and positive real numbers, respectively. 
Throughout the paper, we assume without loss of generality that the masses $\ba=(a_1,\dots,a_I)$ and $\bb=(b_1,\dots,b_J)$ are positive normalized vectors, i.e., $\sum_i a_i = \sum_j b_j =1$. We also provide a notation table in Table~\ref{tb:notation} in the Appendix for convenience.

%\subsection{Problem setup}
%We presently consider how to define the optimal transport plan under adversarial attacks in which the optimal plan is a trade-off between loss in goods by adversarial attacks $\ell^{\text{AA}}$ by damaging the $N$ most critical transportation paths in the learned optimal transportation plan $\mP$ and the transportation cost $\ell^{\text{TC}}$ where
%\begin{align}
% \ell^{\text{AA}}_N= \sum_{n=1}^N s_n,\quad
% \ell^{\text{TC}}= \inner{\cost}{\mP},
%\end{align}
%where $s_n$ is the $n$-th largest entry of $\mP$. Consequently, we define optimal transport under adversarial attack as the plan $\mP$ that produces the most favorable trade-off between $\ell^{\text{AA}}_N$ and $\ell^{\text{TC}}$, i.e., keeping both losses at a minimum. We further define the Pareto front by tracing the plans $\mP^{*\text{AA}}$ having the minimal $\ell^{\text{AA}}_N$ for any given specification of the transportation cost $\ell^{\text{TC}}$. We note that this Pareto front is naturally bounded by $\ell^{\text{TC}}_{min}=\inner{\cost}{\mP^*}$ defining the plan with minimal transportation cost given in \eqref{eq:OT} and the  "uniform" plan $\mP=\frac{1}{\|\mathbf{a}\|_1}\mathbf{ab}$ producing minimal adversarial loss $\ell^{\text{AA}}_N=???$ but maximal transportation cost on the Pareto front $\ell^{\text{TC}}_{max}=\inner{\cost}{\frac{1}{\|\mathbf{a}\|_1}\mathbf{ab}}$. Consequently, the aim is to find transportation plans with optimal resilience (i.e., lowest $\ell^{\text{AA}}_N$) for any given cost in the regime $[\ell^{\text{TC}}_{min}:\ell^{\text{TC}}_{max}]$.

\subsection{The \texorpdfstring{$\chi$}{Chi}-escort distribution and deformed logarithm and exponential functions}\label{sec:escort}

Since $\chi$-ET is a generalization of entropic optimal transport using escort functions and their associated generalized logarithm functions, we review these definitions here, as a preliminary to the discussion of $\chi$-ET in Section~\ref{sec:ET}.

Let $\chi:\mathbb{R}_{>0}\to\mathbb{R}_{>0}$ be a differentiable increasing function, which we refer to as an \emph{escort function}. For a distribution $\mathbf{P}$, we define the $\chi$-escort distribution by 
\begin{align}\label{eq:escort_dist}
\cEsc{\mP}_{ij} \coloneqq \frac{\Esc{\mP_{ij}}}{\sum_{ij}\Esc{\mP_{ij}}}.    
\end{align}
The escort function $\chi$ nonlinearly reweights $\mP$, thereby reallocating probability mass so that specific regions of the distribution, such as tails or peaks, are emphasized or downweighted. For instance, the Tsallis escort $\chi_q(t)=t^q$ with $q> 1$ emphasizes larger values and downweights smaller values via a \emph{global} power-law transformation~\citep{naudts2004estimators}. Another example is the Kaniadakis escort, defined by $\chi_\kappa(t)=t/\cosh\bigl(\kappa\log t\bigr)$, which induces an \emph{adaptive} reweighting whose effective exponent depends on $t$~\citep{kaniadakis2020kappa}. For small $t$, it behaves approximately as $\chi_\kappa(t) \simeq t^{1+\kappa}$, while for large $t$ it approaches $\chi_\kappa(t) \simeq t^{1-\kappa}$, and the transform is close to the identity around $t \simeq 1$~\citep{Hristopulos2022Kani, Wada2023Kani}.
%for small $t$ it behaves approximately as $\chi(t) \simeq t^{1+\kappa}$, whereas near $\chi(t) \simeq 1$ the transformation becomes closer to the identity. 
In these examples, hyperparameters $q$ and $\kappa$ control the strength and the regime of the transformation. The escort distribution $\cEsc{\mP}$ recovers the original distribution $\tP$ in the limits  $q\rightarrow 1$ and $\kappa \rightarrow 0$, where the escort function reduces to the identity, i.e., $\chi_q(t)\rightarrow t \ (q \rightarrow 1)$ and $\chi_\kappa(t)\rightarrow t \ (\kappa \rightarrow 0)$. %In the Tsallis escort, we consider only $q\geq 1$, for the reason discussed in Section~\ref{sec:method}.

%For any differentiable increasing non-negative function $\chi:\RR_{>0} \rightarrow \RR_{>0}$, we define the $\chi$-escort of distribution $\mP$ as $\cEsc{\mP_{ij}} \coloneqq \Esc{\mP_{ij}}/\sum_{ij}\Esc{\mP_{ij}}$. The escort function $\chi$ can emphasize the feature in the distribution by distorting its shape. For instance, the Tsallis escort $\chi_q(t)=t^q$ with $q \geq 1$ emphasize larger value understate smaller value by powering these values $q$. 

%We mainly discuss two kinds of escort, Tsallis escort $\chi_q(t)=t^q$ and Kaniadakis escort $\chi_\kappa(t)=2t^{\kappa+1}/(t^{2\kappa}+1)$ where $q$ and $\kappa$ are hyper-parameter controlling the behavior of escorts. 

As seen in Section~\ref{sec:optim}, the $\chi$-escort distribution generates a coordinate system for a $\chi$-deformed exponential family where the following  $\chi$-deformed logarithm and its inverse, the  $\chi$-deformed exponential function, play a central role~\citep{Nauduts2002Deformed, Korbel2019Information}.
\begin{align}\label{eq:clog_cExp}
    \cLog{t} = \int_{1}^{t} \frac{ds}{\Esc{s}}, \quad
    \cExp{s} = \invcLog{s},
\end{align}
for $t\in(0,\infty)$ and $s\in(\cLog{0^+},\cLog{\infty})$. The positivity of the escort function $\Esc{t}>0$ is required to guarantee the monotonicity of $\cLog{t}$ and $\cExp{t}$. The monotonicity of the escort function $\Escp{t}>0$ is required to guarantee the concavity of $\cLog{t}$ and convexity of $\cExp{t}$. For the Tsallis escort, the corresponding deformed logarithm and exponential functions are given as
\begin{align}\label{eq:q_log_q_exp}
    \qLog{t}=\frac{t^{1-q}-1}{1-q}, \quad \qExp{t} = [1+(1-q)t]^{\frac{1}{1-q}}_+,
\end{align}
respectively, where the cutoff function is defined as $[\,\cdot\,]_{+}=\max(x,0)$. For the Kaniadakis escort, the correspondence is given as
\begin{align*}
    \kLog{t}=\sinh(\kappa \log t)/\kappa, \quad \kExp{t} = \left(
    \kappa t +\sqrt{1+\kappa^2 t^2}
    \right)^{1/\kappa},
\end{align*}
respectively~\citep{Scarfone2018Information}. 
We note that $q\rightarrow1$ and $\kappa \rightarrow 0$ recover the ordinary exponential function $e^t$ and logarithm function $\log{t}$. As described in Sections~\ref{sec:ET} and~\ref{sec:tail}, these deformed exponential functions enable us to tune the decay of the transport plan. However, Kaniadakis's escort inherently covers only slow decay, and the cutoff function in Equation~\eqref{eq:q_log_q_exp} makes the Tsallis's escort unable to offer the rapid decay full support plan with $q<1$. 

Therefore, to presently acquire a wider tunable decay regime, we propose an alternative escort function $\chi_m$ as
\begin{align}\label{eq:m_escort}
    \chi_m(t) = m (t+1) t \left[\log\left(1+\frac{1}{t}\right)\right]^{1-1/m},
\end{align}
where the positive real-valued hyperparameter $m$ continuously controls the decay profile of the transport plan from rapid ($m>1$) to slow ($m<1$) while ensuring the transport plan has full support. 
%where the positive real-valued hyperparameter $m$ continuously controls the decay from slow ($m<1$) to rapid ($m>1$) while retaining full support. 
We name this escort in Equation~\eqref{eq:m_escort} as \emph{Weibull escort} because it causes Weibull decay asymptotically as discussed in Section~\ref{sec:tail}. Its corresponding deformed logarithm and exponential functions have a closed-form:
\begin{align}\label{eq:m_log_exp}
    \mLog{t} = c_m-\left[\log\left(1+\frac{1}{t}\right)\right]^{1/m}, \quad
    \mExp{t} = \frac{1}{\exp{[(c_m-t)^m}]-1},
\end{align}
where $c_m=(\log{2})^{1/m}$. Any positive hyperparameter $m$ in Equation~\eqref{eq:m_log_exp} does not recover the ordinary logarithm and exponential functions. We here note favorable properties of the proposed $\chi_m$-function: 
(i) the decay of the optimal plan is controllable in both regimes, from rapid ($m>1$) to slow ($m<1$) decay, by tuning a single hyperparameter $m>0$,
(ii) the non-bounded below deformed logarithm function $\mLog{t}$ guarantees the full support of the optimal plan contributing to the noise robustness in barycentric projections, and 
(iii) the corresponding $\mLog{t}$ and $\mExp{t}$ have closed-form without special functions nor numerical approximation. The above properties (i) and (ii) are detailed in Section~\ref{sec:tail}. The plots of these escort functions can be found in Figure~\ref{fig:full_plots}(\textbf{a}) in Appendix~\ref{app:remarks}.

\subsection{\texorpdfstring{$\chi$}{Chi}-escort entropic optimal transport}\label{sec:ET}

We here consider $\chi$-escort entropic optimal transport ($\chi$-ET) with regularizer $\Omega(\mR)=\varphi_\chi( \invEsc{\mR})$ developed by~\citet{kurose2022optimal} as
\begin{align}\label{eq:tradional_OT}
    \mR = \argmin_{\mR\in\simplex{\ba}{\bb}} 
    \inner{\cost}{\mR} 
    + \reg\varphi_\chi( \invEsc{\mR}),
\end{align}
where the transportation polytope $\simplex{\ba}{\bb}$ and the $\chi$-entropy $\varphi_\chi$ are defined as
\begin{align}\label{eq:chi_entropy}
    \simplex{\ba}{\bb} = \set{\mR \mid 
    \sum\nolimits_j \mR_{ij}=a_i, \ 
    \sum\nolimits_i \mR_{ij}=b_j, \ \mR_{ij} \geq 0},\quad
\varphi_\chi(\mP)=\sum\nolimits_{ij}\cEsc{\mP}_{ij}\cLog{\mP_{ij}},
\end{align}
respectively. The function $\Tilde{\chi}^{-1}$ is the inverse of $\chi$-escort distribution defined in Equation~\eqref{eq:escort_dist}. The positive hyperparameter $\reg$ controls the strength of the regularization. As we see in Section~\ref{sec:tail}, the $\chi$-entropy controls the decay of the plan $\mR$. To leverage the information geometric optimization methods, we focus on an alternative form of Equation~\eqref{eq:tradional_OT} as follows.
%To take advantage of the decay-tunable properties of escort distributions in optimal transport problems, we here review $\chi$-entropic optimal transport.

%While ordinary optimal transport finds an optimal transport plan $\tP$ between the distributions given a cost matrix $\cost$, the proposed escort optimal transport finds a plan between their corresponding escort distributions. Thus, we here define $\chi$-entropic regularized escort transport where the regularization $\varphi_\chi(\mP)=\sum_{ij}\cEsc{\mP_{ij}}\cLog{\mP_{ij}}$ relaxed the problem convex optimization. 

\begin{graybox}
\textbf{$\chi$-Escort Entropic Optimal Transport ($\chi$-ET)~\citep{kurose2022optimal} \ } Given cost matrix $\cost\in\RR_{>0}^{I\times J}$ and vectors $\ba\in\RR_{>0}^I$ and $\bb\in\RR_{> 0}^J$, the optimal $\chi$-escort plan $\mP$ is defined as 
\begin{align}\label{eq:ET}
    \mP = \argmin_{\mP\in\csimplex{\ba}{\bb}} 
    \inner{\cost}{\cEsc{\mP}} 
    + \reg \varphi_\chi(\mP),
\end{align}
for the $\chi$-transportation polytope
\begin{align}\label{eq:simplex}
    \csimplex{\ba}{\bb} = \set{\mP \mid 
    \sum\nolimits_j \cEsc{\mP}_{ij}=a_i, \ 
    \sum\nolimits_i \cEsc{\mP}_{ij}=b_j, \ \mP_{ij} \geq 0}.
    %\sum_{ij} \mP_{ij}=1
    %}
\end{align}
%and $\chi$-entropy 
%\begin{align}\label{eq:chi_entropy}
%\varphi_\chi(\mP)=\sum\nolimits_{ij}\cEsc{\mP}_{ij}\cLog{\mP_{ij}}.
%\end{align}
\end{graybox}
%By learning a plan $\mP$ such that its escort distribution optimizes the total cost, we can take advantage of the favorable structural properties of escort distributions. 
%Assuming the escort map $\Tilde{\chi}$ is a bijection on the simplex and 
Defining $\mR=\cEsc{\mP}$, the problem in Equation~\eqref{eq:ET} can be written as the original form of $\chi$-ET in Equation~\eqref{eq:tradional_OT}. This problem in both Equations~\eqref{eq:tradional_OT} and~\eqref{eq:ET} reduces to the ordinary entropic optimal transport when $\chi(t)=t$ because the regularizer $\varphi_\chi(\mP)$ recovers the negative Shannon entropy $\varphi(\mP)=\sum_{ij}\mP_{ij}\log{\mP_{ij}}$. %Thus, $\chi$-ET generalizes the entropic regularized optimal transport. 
%Moreover, regardless of the choice of escort function $\chi$, Equation~\eqref{eq:ET} reduces the original optimal transport in Equation~\eqref{eq:OT} in the limit of $\reg \rightarrow 0$. 
We refer to the constraint in Equation~\eqref{eq:simplex} that defines the feasible manifold $\csimplex{\ba}{\bb}$ as the \emph{$\chi$-balancing condition}. Depending on the choice of the escort function $\chi$, we refer to the resulting $\chi$-ET as the Tsallis-ET, Kaniadakis-ET, or Weibull-ET. Notably, since the normalizing condition of $\mP$ ensures $\mP_{ij} < 1$, only the $\chi_\kappa(t) \simeq t^{1+\kappa}$ branch is reachable and the effect of Kaniadakis-escort becomes similar to the simpler power Tsallis-escort with $q = 1 + \kappa$, thus we presently focus mainly on the Tsallis and the proposed Weibull-ET.
%The original optimal transport without regularization $\Omega(\mR)=0$ is recovered in the limit of $\reg\rightarrow0$ regardless of the choice of escort function $\chi$. 

As seen in Section~\ref{sec:optim}, we leverage the projection theory of deformed information geometry via reformulating $\chi$-ET as 
%The benefit of introducing $\varphi_\chi(\mP)$ instead of the Shannon entropy $\varphi(\mP)$ is to leverage the projection theory of information geometry via reformulating $\chi$-ET as 
\begin{align}\label{eq:reform}
    %\argmin_{\mP\in\csimplex{\ba}{\bb}} \inner{\mM}{\cEsc{\mP}} + \varepsilon \varphi_\chi(\mP) 
    \text{Equation~\eqref{eq:ET}}
    = \argmin_{\mP\in\csimplex{\ba}{\bb}}  \varepsilon D_{\chi}[\mP; \mK],
\end{align}
where the $\chi$-divergence~\citep{amari2012geometry, Matsuzoe2017Sequential} and the normalized $\chi$-deformed kernel matrix $\mK$ is defined as
\begin{align}\label{eq:chi_div}
    D_\chi[\mP;\mK] = \sum_{ij} \cEsc{\mP}_{ij}\left\{ \cLog{\mP_{ij}} - \cLog{\mK_{ij}} \right\}, \quad \ \ %\\
    \mK_{ij} = \cExp{-\frac{\cost_{ij}}{\reg}+g}, %\label{eq:Kernel}.
\end{align}
respectively. \GK{The $\chi$-divergence is a subfamily of Bregman divergences, as discussed in Section~\ref{sec:optim}.} The real-valued constant $g$ is uniquely identified by the normalization of $\mK$, i.e., $\sum_{ij}\mK_{ij}=1$. 
%Due to the escort $\Tilde{\chi}[\mP]$, the $\chi$-divergence in Equation~\eqref{eq:chi_div} is not separable, for which \citet{Arnaud2018Regularized} provide no efficient projection scheme. 
Although we optimize Equation~\eqref{eq:ET} with respect to $\mP$ in the following discussion, the transport of the mass $\ba$ to $\bb$ is performed by $\mR=\cEsc{\mP}$. The $\chi$-divergence for the Tsallis escort $\chi_q(t)=t^q$ is referred to as the Tsallis divergence~\citep{tsallis1999nonextensive} and its property is reported in~\citep{Poczos2011on, ghalamkari2026em}. %, and its optimization is equivalent to Amari's $\alpha$-divergence. 
 %The proof is based on our finding that the constant $g$ does not change the optimal $\mP$ as shown in Theorem~\ref{th:solution_invariance} in the Appendix. 
The above reformulation is formally proven in Proposition~\ref{prop:ET_and_divergence} in Appendix~\ref{sec:proofs}, which is based on the fact that the constant $g$ does not change the optimal $\mP$. 
%In the following, we show that the projection theory in information geometry derives our algorithm introduced in Section~\ref{sec:algorithm}, leveraging the flatness of submanifolds in the coordinate system of the deformed exponential family.
If we naively normalize the kernel matrix $\mK$ by dividing all its elements by the total sum of $\mK$, then the optimal solution $\mP$ is not guaranteed to be that of the original objective in Equation~\eqref{eq:reform}. As discussed in Section~\ref{sec:optim}, the $\chi$-transportation polytope $\csimplex{\ba}{\bb}$ is a flat manifold; therefore, $\chi$-ET optimizing Equation~\eqref{eq:chi_div} is a convex optimization problem, upon choosing an appropriate coordinate system.
%ensuring the global optimality of the proposed algorithm. % despite the non-linear reweighting of the plan $\mP$.

%Although the $\chi$-divergence $D_{\chi}[\mP; \mK]$ is a convex function with respect to $\mP$, the convexity of the feasible set $\csimplex{\ba}{\bb}$ is nontrivial. Thus, to formulate the optimization in Equation~\eqref{eq:reform} as a convex optimization, we employ alternative coordinate systems in the following sections.

This reformulation in Equation~\eqref{eq:reform} and the algorithm proposed below assume that the appropriate normalizer $g$ in Equation~\eqref{eq:chi_div} does not let $-\mM_{ij}/\reg+g$ be outside the domain of the $\chi$-exponential $\cExp{\,\cdot\,}$, i.e., 
\begin{align*}
   \cLog{0^+}<-\mM_{ij}/\reg+g<\cLog{\infty}.
\end{align*}
When the domain of the $\chi$-exponential is \emph{bounded below}, i.e., $\cLog{0^+}>-\infty$, 
this assumption can fail and cause domain issues. In fact, the domain of the Tsallis exponential is bounded below for $q<1$ as seen in Figure~\ref{fig:Tsallis_log} in Appendix~\ref{app:remarks}. As a result, the optimization 
%via formulation in Equation~\eqref{eq:ET} 
might be infeasible in the regime of $q<1$ as detailed in Appendix~\ref{sec:domain_issue}. If we consider the Tsallis-ET with $q<1$ in the original formulation in Equation~\eqref{eq:tradional_OT} directly, it is still a convex optimization problem since the regularizer is given as a convex function
\begin{align*}
    \varphi_q(\mR) = \frac{1}{1-q} \left( \left(\sum_{ij} \mR_{ij}^{1/q}\right)^q -1 \right).
\end{align*}
However, it removes the barrier at the boundary, i.e., $|\, \partial \Omega/\partial \mR_{ij}\,| < \infty$ for $\mR_{ij}\rightarrow0$ and $q<1$, which causes the cutoff function in the form of the transport plan $\mR$. As detailed in Section~\ref{sec:tail}, the cutoff function may cause instability against the noise in the given histogram in the barycentric projection, a downstream task of the optimal transport. In contrast, the domain of the proposed Weibull deformed exponential is \emph{unbounded below}, i.e., $\mLog{0^+}=-\infty$, which does not cause a domain issue over all $m\in\RR_{>0}$ and guarantees a full support plan.

As a remark, in the limit $\reg\rightarrow\infty$, the optimization in Equation~\eqref{eq:ET} becomes independent of the cost matrix $\cost$. This cost-free limit admits $\chi$-deformed rank-$1$ form, $\mP_{ij}\rightarrow X_i \dtimes_\chi Y_j$, where $\chi$-deformed product $\dtimes_\chi$ is defined through the generalized exponential law $\cExp{x_i}\dtimes_\chi\cExp{y_j}=\cExp{x_i+y_j}$ as shown in Proposition~\ref{prop:limit} in Appendix~\ref{sec:proofs}~\citep{Matsuzoe2015Deformed,Ghalamkari2026Deformed}. While this fact includes the known rank-$1$ form of the plan $\mP_{ij} \rightarrow a_i b_j$ ($\reg \rightarrow\infty$) in the entropic optimal transport $\chi(t)=t$, there is no guarantee that factors $X_i$ and $Y_j$ can be given as a closed form for a general escort function $\chi$. Please also refer to Appendix~\ref{sec:deformed_product} for the explicit form of the $\chi$-deformed product. 

%The reformulation in Equation~\eqref{eq:reform} assumes that all entries of $\mK$ lie in the domain of the $\chi$-deformed logarithm. However, for the Tsallis deformation with $q<1$, the cut‑off function $[\,\cdot\,]_{+}$ in Table~\ref{tab:deformed_exp_log} may produce zero entries in $\mK$, which lie outside the domain of the Tsallis-deformed logarithm. Accordingly, we restrict our attention to $q\geq1$ in the Tsallis escort. In contrast, our proposed escort $\chi_m$ does not have a domain issue over all $m\in\RR_{>0}$.

%We also note that, using the inverse escort $\chi^{-1}$, we can reformulate the problem in Equation~\eqref{alg:ET} into regularized optimal transport 
%\begin{align*}
%    \mR = \argmin_{\mR\in\simplex{\ba}{\bb}} 
%    \inner{\cost}{\mR} 
%    + \reg\varphi_\chi( \invEsc{\mR})
%\end{align*}
%where we define $\mR=\cEsc{\mP}$. However, there is no guarantee we have a closed-form of the inverse, which we can naturally expect the naive formulation to have additional complexity.
\IncMargin{1em}
\begin{algorithm*}[t]
 \SetKwInOut{Input}{input}
 \SetKwInOut{Output}{output}
 \SetFuncSty{textrm}
 \SetCommentSty{textrm}
 \SetKwFunction{ManyBodyApproximation}{{\scshape EscortTransport}}
 \SetKwFunction{EM}{{\scshape EscortTransport}}
 \SetKwProg{myfunc}{}{}{}
 \myfunc{
 \ManyBodyApproximation{$\chi$, $\cost$, $\ba$, $\bb$, $\varepsilon$, $\delta$}
    %\tcp*[f]{$\bmtheta^{t=0}$ is the initial value of $\bmtheta^{\tP}$}
    }
    {
        %$\mK_{ij} \leftarrow \cExp{-\cost_{ij}/\varepsilon+g}$ with normalizing factor $g$; \\
        Compute $\hat{\bmeta}_1$ using Eq.~\eqref{eq:onebody_etahat};\\
        Compute the two-body natural parameters of the Kernel matrix $\bmtheta_2^{\mK}$ using Eq.~\eqref{eq:update_thetaK}; \\
        Initialize the plan $\mP$ and compute its one-body natural parameters $\bmtheta_1$; \\
        \Repeat{$\| \bmeta_1 - \hat{\bmeta}_1 \| < \delta \ $\label{line:endloop}}{\label{line:startloop}
            Compute $\theta_{11}$ to satisfy  $\sum_{ij}\mP_{ij}=1$ using the bisection method;\\  
            Compute $\tP$ using the current parameter $\theta_{11}, \bmtheta_{1}$ with Eq.~\eqref{eq:parameterization} where $\bmtheta_2$ is fixed to $\bmtheta^{\mK}_2$;\\
            Compute $\bmeta_1$ using Eq.~\eqref{eq:parameterization};\\
            Compute $\chi$-Fisher information matrix $\mathbf{G}$ using Eq.~\eqref{eq:FIM};\\
            Estimate appropriate learning-rate $\beta$ by line search; \\
         $\bmtheta_1 \leftarrow \bmtheta_1 - \beta\FIM^{-1}(\bmeta_1 - \hat{\bmeta}_1)$;\\
    }
        \Return{{\normalfont Optimal plan} $\mR=\cEsc{\tP}$}
    }
\caption{$\chi$-Escort Entropic Optimal Transport ($\chi$-ET)}
\label{alg:ET}
\end{algorithm*}
\DecMargin{1em}

\subsection{Proposed algorithm}\label{sec:algorithm}
Given the cost matrix $\cost$, mass $\ba$ and $\bb$, escort function $\chi$, and the strength of the regularization $\reg$, we present the algorithm to obtain the optimal plan $\mR=\cEsc{\tP}$ that minimizes Equation~\eqref{eq:reform}. 

We parameterize an initial $\mP$ by the natural parameter $\bmtheta = (\theta_{ij})$ and expectation parameter $\bmeta = (\eta_{ij})$ as
\begin{align}\label{eq:parameterization}
    \mP_{ij} = \cExp{\sum\nolimits_{i'=1}^i\sum\nolimits_{j'=1}^j \theta_{i'j'}}, \quad
    \eta_{ij} = \sum\nolimits_{i'=i}^{I}\sum\nolimits_{j'=j}^{J}\cEsc{\mP_{i'j'}},
\end{align}
where the parameterizations are justified in Section~\ref{sec:optim}. As common in the literature in~\citep{Ghalamkari2021Fast, Ghalamkari2023Nonnegative, hu2026pseudononlinear}, we collect parameters for which either index $i$ or $j$ equals $1$ and define one-body natural parameters $\bmtheta_1 = (\theta_{21}, \theta_{31}, \dots, \theta_{I1}, \theta_{12}, \dots, \theta_{1J})^\top$, and one-body expectation parameters $\bmeta_1   = (\eta_{21}, \eta_{31}, \dots, \eta_{I1}, \eta_{12}, \dots, \eta_{1J})^\top$. Similarly, we define \emph{two-body natural parameters} as the natural parameters whose indices are both different from $1$, i.e., $\bmtheta_2 = (\theta_{22},\dots,\theta_{IJ})^\top$. %are referred to as \emph{two-body natural (expectation) parameters}. 
Here, $\theta_{11}$ and $\eta_{11}$ are automatically determined by the normalization of $\mP$ and are therefore not free parameters. Consequently, they are neither one-body nor two-body parameters.

We define the matrix $\mK$ by Equation~\eqref{eq:chi_div} and %with the normalizing factor $g$ obtained by the bisection method. % where the constant $g$ is uniquely identified so as to $\sum_{ij}\mK^e_{ij}=1$ by bisection method. 
obtain its natural parameter $\bmtheta^\mK$ of the matrix $\mK$ by the inclusion-exclusion principle~\citep{rota1964foundations}:
\begin{align}\label{eq:update_thetaK}
    \theta_{ij}^{\mK} = \cLog{\mK_{ij}} - \cLog{\mK_{i-1,j}} - \cLog{\mK_{i,j-1}} + \cLog{\mK_{i-1,j-1}}.
\end{align}
Putting Equation~\eqref{eq:chi_div} into Equation~\eqref{eq:update_thetaK}, we can obtain $\theta_{ij}^{\mK}$ without numerically evaluating the normalizing constant $g$, i.e., $\theta_{ij}^{\mK}=(-\cost_{ij}+\cost_{i-1,j}+\cost_{i,j-1}-\cost_{i-1,j-1})/\reg$. Throughout the algorithm, we fix the two-body natural parameters of $\mP$ to those of $\mK$, i.e., $\bmtheta_2=\bmtheta_2^{\mK}$. In each iteration, we update the one-body natural parameters $\bmtheta_1$ by
\begin{align}\label{eq:update_rule}
    \bmtheta_1 \gets \bmtheta_1 - \beta\FIM^{-1}(\bmeta_1 - \hat{\bmeta}_1),
\end{align}
where the one-body expectation parameters $\bmeta_1$ are precomputed by Equation~\eqref{eq:parameterization} and each element in $\hat{\bmeta}_1=(\hat{\eta}_{21}, \dots, \hat{\eta}_{I1}, \hat{\eta}_{12}, \dots, \hat{\eta}_{1J})^\top$ is given as
\begin{align}\label{eq:onebody_etahat}
    \hat{\eta}_{i1}=\sum\nolimits_{i'=i}^{I} a_{i'}, \quad 
    \hat{\eta}_{1j}=\sum\nolimits_{j'=j}^{J} b_{j'}.
\end{align}
The step size $\beta$ is obtained in each iteration by line search~\citep{armijo1966minimization}. The Jacobian $\FIM\in\RR^{(I+J-2)\times(I+J-2)}$ is given as
\begin{align}\label{eq:FIM}
    \FIM_{(ij),(vw)} = \frac{\partial \eta_{vw}}{\partial \theta_{ij}} = \sum\nolimits_{k=1}^I \sum\nolimits_{l=1}^J \cEsc{\mP}_{kl}\chifp{\mP_{kl}}
    \Delta_{ij}^{kl}\Delta_{vw}^{kl},
    %\left(
    %    \zeta_{ij}^{kl}-\eta_{ij}
    %\right)
    %\left(
    %    \zeta_{i'j'}^{kl}-\eta_{i'j'}
    %\right)
\end{align}
where the term $\Delta_{ij}^{kl}$ is defined as $\Delta_{ij}^{kl}=\zeta_{ij}^{kl}-\eta_{ij}$ and the zeta-function $\zeta:\mathcal{X}\times\mathcal{X}\rightarrow\{0,1\}$ is defined as 
$\zeta_{ij}^{kl}=1$ if $i \leq k, j \leq l$ and $0$ otherwise. % The parameter $\theta_{11}$ is then determined by a bisection method so that the normalization of $\mP$ is satisfied. 
Subscripts of the matrix $\FIM$ are indices of the one-body parameters, i.e., $(ij), (vw)\in\{12,13,\dots,1J, 21, 31, \dots, I1\}$. Then, we update $\tP$ using the obtained $\bmtheta$ in Equation~\eqref{eq:parameterization} where $\theta_{11}$ is determined by the bisection method to normalize $\mP$, i.e., $\sum_{ij}\mP_{ij}=1$. The above procedure is repeated until the residual error $\|\bmeta_1 - \hat{\bmeta}_1 \|$ is below a predefined threshold $\delta$. The entire procedure is summarized in Algorithm~\ref{alg:ET}. 

The convergence guarantee is discussed in Section~\ref{subsec:natural_gradient}. 
%This algorithm has no dependency on an initial plan as discussed in Section~\ref{subsec:natural_gradient}. % and guarantees a global optimal $\chi$-transport plan $\tP$, minimizing Equation~\eqref{eq:ET}. 
If we assume $\ba$ and $\bb$ are normalized uniform vectors and $\chi(t)=t$,  the above procedure recovers the matrix balancing algorithm proposed by~\citet{Sugiyama2017Tensor}.

\subsection{Decay of the optimal plan and barycentric stability induced by escort}\label{sec:tail}

The decay of the transport plan can be tuned by the escort function $\chi$. More specifically, the critical condition of the Lagrange function of Equation~\eqref{eq:ET} guarantees the solution form as $\mR_{ij}=\cEsc{\mP}_{ij} \propto \cExpp{-z_{ij}}$ where $z_{ij}=\mM_{ij}/\reg+\alpha_i+\beta_j$ for Lagrange multipliers $\boldsymbol{\alpha}=(\alpha_i)_i$ and $\boldsymbol{\beta}=(\beta_j)_j$, which is proven in Proposition~\ref{prop:decay} in Appendix~\ref{sec:proofs}. As a result, Tsallis-ET and Weibull-ET offer a transport plan that follows a power decay and a Weibull decay as:
\begin{align*}
    \qEsc{\mP}_{ij} \propto z_{ij}^{-q/(q-1)}  \ \ (q>1), \quad \mEsc{\mP}_{ij}\propto mz_{ij}^{m-1}\exp{\left[-z_{ij}^m\right]} \ \ (z_{ij} \gg c_m).
\end{align*}
Since the Tsallis logarithm with $q>1$ and Weibull logarithm with any $m\in\RR_+$ are unbounded below, 
%Since the $q$($>1$)- and $m$-deformed logarithms are unbounded below, i.e., $\mLog{0^+}=-\infty$, 
the regularizer has a boundary barrier $|\,\partial \Omega/\partial \mR_{ij}\,| \rightarrow \infty$ as $\mP_{ij}\rightarrow0$, which guarantees the full support of the optimal plan as proven in Proposition~\ref{prop:barrier} in Appendix~\ref{sec:proofs}.

%Given bounded above domains of the $q$($>1$)-exponential and $m$-exponential functions, their full-support of the optimal plan is guaranteed because of the zero barrier, i.e., $|\,\partial \Omega/\partial \mR_{ij}\,| \rightarrow \infty$ for $\mP_{ij}\rightarrow0$, which is formally proven in Proposition~\ref{prop:barrier} in the Appendix. 

Tsallis-ET with $q<1$ is bounded below, and it may cause a cutoff function. On the other hand, the proposed Weibull-ET ensures a full-support plan with both rapid decay ($m>1$) and slow decay ($m<1$) regimes. We also note that Kaniadakis-ET covers only slow decay plans, which is shown in Proposition~\ref{prop:Kani_tail} in Appendix~\ref{sec:proofs}. The larger $q$ in Tsallis-ET or a smaller $m$ in Weibull-ET enhances averaging effects by allowing transportation not only along low-cost paths. In contrast, larger $m$ in Weibull-ET promotes selective effects. This difference affects the robustness of its downstream barycentric projection under noise in a given marginalized condition, as seen below.

%providing an alternative to the Tsallis escort.

\begin{table}[t]
\centering
\caption{Three roles of the escort function in $\chi$-ET.}
\label{tab:escort_response}
\begin{tabular}{lll}
\toprule
 & Governor & Role of the escort function $\chi$ \\
\midrule
Decay of the transport plan      $\mR=\cEsc{\mP}$ & $\cExpp{-z}$                   & Controls the tail or decay profile. \\
Local decay rate $\rho_\chi(z)$ & $\chi^\prime \circ {\cExp{-z}}$ & Controls the local adaptability. \\
Response weights  $S^\chi(z)$ & $\cExppp{-z}$                   & First-order barycentric response to perturbations. \\
\bottomrule
\end{tabular}
\end{table}

We consider two point clouds, $X=(\bx^{(1)},\dots,\bx^{(I)})$ and $Y=(\by^{(1)},\dots,\by^{(J)})$. The points $\bx^{(i)}\in\RR^{N}$ and $\by^{(j)}\in\RR^{N}$ have masses $a_i$ and $b_j$, respectively, which form the mass vectors $\ba=(a_1,\dots,a_I)$ and $\bb=(b_1,\dots,b_J)$. The regularized optimal plan transporting points $X$ to $Y$ is given in Equation~\eqref{eq:tradional_OT} with cost matrix $\cost_{ij}=d(\bx^{(i)},\by^{(j)})$ for appropriately chosen \GK{function} $d:\RR^N \times \RR^N \rightarrow \RR$. Using the obtained plan $\mR$, the barycentric projection moves the points $X$ to $\hat{X}=(\hat{\bx}^{(1)},\dots,\hat{\bx}^{(I)})$ to minimize the total cost, i.e.,
\begin{align}\label{eq:Barycenter}
    \hat{\bx}^{(i)} = \argmin_{\bx\in\RR^N} \sum_{j=1}^J \mR_{ij} d(\bx,\by^{(j)}).
\end{align}
When the \GK{function $d$} is defined as the squared Euclidean norm $d(\bx,\by)=\| \bx-\by \|^2$, the projection has a closed form solution given as $\hat{\bx}^{(i)}=\sum_{j} w_{ij} \by^{(j)}$ for the weights $w_{ij}=\mR_{ij}/a_i$~\citep{seguy2018large, Blondel2018Smooth}. Since $\mR_{ij}\propto\cExpp{-z_{ij}}$ holds as discussed above, the weights follow a generalized softmax function for $\bz_i=(z_{i1},\dots,z_{iJ})$ as 
\begin{align}\label{eq:chi_softmax}
    w_{ij}^{[\chi]}(\bz_i)=\frac{\cExpp{-z_{ij}}}{\sum_{j=1}^J\cExpp{-z_{ij}}},
\end{align}
which recovers the power-softmax function for the Tsallis escort $\chi(t)=t^q$ and the ordinary softmax function for $\chi(t)=t$~\citep{Abe2003GeometryOE}.
%the latter $\boldsymbol{w}^{[1]}$ corresponds the weight for the ordinarily entropic optimal transport. 

%where we defined $z_{ij}=\cost_{ij}+\lambda_i + \mu_j$. 
We consider the effect of perturbation $\ba \rightarrow \ba + \Delta \ba$ to the barycentric projection in Equation~\eqref{eq:Barycenter} assuming $|\Delta a_i| \ll 1$. The perturbation $\Delta \ba$ shifts the weights $w_{ij}$ as $w_{ij} \rightarrow w_{ij} + \Delta w_{ij}$. %Under the perturbation, the Lagrange multipliers are affected by the perturbation, i.e,  $\lambda_i \rightarrow \lambda_i + \Delta \lambda_i$ and $\mu_j \rightarrow \mu_j+ \Delta \mu_j$. 
Our first-order perturbation analysis in Theorem~\ref{th:EGRF} in Appendix~\ref{sec:proofs} concludes that the perturbation effect on the barycenter $\Delta\hat{\bx}^{(i)}=\sum_j \Delta w_{ij} \by^{(j)}$ depends on the escort function $\chi$ as follows:
\begin{align}\label{eq:escort_Gibbs_response}
    \Delta\hat{\bx}^{(i)[\chi]}(\bz_i) =
    -\sum_{j=1}^J w_{ij}(\bz_i) \rho_\chi(z_{ij})(\by^{(j)}-\hat{\bx}^{(i)})\Delta z_{ij} + O(\|\Delta \bz\|^2),
    %\delta\hat{\bx}^{(i)[q]}  = -\frac{q}{q-1} \sum\nolimits_j \frac{w_{ij}^{[q]}}{z_{ij}} (\by^{(j)}-\hat{\bx}^{(i)})\delta z_{ij}, \quad 
    %\delta\hat{\bx}^{(i)[q=1]} =  -\frac{1}{\reg} \sum\nolimits_j w^{[q=1]}_{ij} (\by^{(j)}-\hat{\bx}^{(i)})\delta z_{ij}
\end{align}
where the effect of the perturbation on $z_{ij}=\mM_{ij}/\reg+\alpha_i+\beta_j$ is defined as $\Delta z_{ij} = \Delta\alpha_i + \Delta\beta_j$ and the local decay rate is given as 
\begin{align*}
    \rho_\chi(z_{ij}) \coloneqq -\frac{\mR'_{ij}(z_{ij})}{\mR_{ij}(z_{ij})}= \chi^\prime\circ{\cExp{-z_{ij}}},
\end{align*} 
as shown in Proposition~\ref{prop:decay_rate} in the Appendix. %The formal derivation of Equation~\eqref{eq:escort_Gibbs_response} can be found in Theorem~\ref{th:EGRF} in the Appendix. %As a remark, we regard Equation~\eqref{eq:escort_Gibbs_response} as an escort generalization of the Gibbs response formula since it recovers the Gibbs response formula with temperature $\reg$ for $\chi(t)=t$.
As seen in Equation~\eqref{eq:escort_Gibbs_response}, the first-order sensitivity of the barycentric projection against the perturbation is affected by the response weight $S^\chi$, which is defined as the product of the weight $w_{ij}$ and the local decay rate $\rho_\chi$ as
\begin{align*}
    S^\chi_{ij}(\bz_i) \coloneqq w_{ij}(\bz_i)\rho_\chi(z_{ij})=\frac{\cExppp{-z_{ij}}}{\sum_j \cExpp{-z_{ij}}},
\end{align*}
where we use Equation~\eqref{eq:deff_chi_relation} of Appendix~\ref{sec:proofs}.
%the product of the weight $w_{ij}$ and local decay rate $\rho_\chi$. 
The response weights for Tsallis-ET and Weibull-ET are proportional to
\begin{align*}
    \qExppp{-z}=q\left(1+(q-1)z\right)^{-\frac{2q-1}{q-1}}, \quad
    \mExppp{-z}=m^2z^{2m-2}\exp{[-z^m]},
\end{align*}
respectively. Therefore, for smaller $m$ and larger $q$, the response weights remain non-negligible over a broader range of $z_{ij}$, allowing centered, weakly correlated perturbations $\Delta z_{ij}$ to partially cancel through averaging over $j$ when sufficiently many paths contribute. A benefit of Weibull-ET is its broader tunability of the response-weight profile, which allows $m$ to be selected to facilitate effective averaging of the perturbations in Equation~\eqref{eq:escort_Gibbs_response}. In contrast, a cutoff reduces the number of contributing transport paths and may limit such averaging, potentially increasing the sensitivity of the barycentric projection to noisy marginals.

%Therefore, for smaller $m$ and larger $q$, the response weights remain non-negligible over a broader range of $z$, allowing perturbations $\Delta z_{ij}$ to partially cancel through averaging over $j$, assuming larger $J$ and centered $\sum_{ij}\Delta z_{ij} \simeq 1$. In contrast, the cut-off function in the form of the optimal plan reduces the effective summation in $\sum_j$, which prevents averaging the perturbations and may bring instability in the noisy Barycentric projection.

%plans with cutoff function reduce the number of contributing paths, limiting this cancellation and potentially making the barycentric response more sensitive to individual perturbations. 

%In slow-decay plans, e.g., $q$-ET with $q>1$ and $m$-ET with $m<1$, the weights $w_{ij}$ are not concentrated on a small number of low-cost paths, but are spread over a broader set of target points $j$. Therefore, perturbations $\delta z$ can more easily cancel out through the summation over $j$, which makes the barycentric projection stable under noisy $\ba$. In contrast, the selective effects $\boldsymbol{w}^{[q=1]}$ cannot cancel out the perturbation through the summation $\sum_{j=1}^J$, leading to an unstable response of $\delta\hat{\bx}$ to the noise. The cut-off function in the form of the optimal plan reduces the effective summation in $\sum_j$ to cancel the noise out, which also brings the instability in the noisy Barycentric projection.

In short, the decay of the optimal plan is governed by $\cExpp{t}$, its local decay is governed by $\chi^\prime \circ \cExp{t}$, and the response weights of the barycentric projection under perturbations are governed by $\cExppp{t}$ as summarized in Table~\ref{tab:escort_response}. The plots of these functions can be found in Figure~\ref{fig:full_plots} in Appendix~\ref{app:remarks}.

%The averaging effects of the long-tailed weights $\boldsymbol{w}^{[q]}$ with $q>1$ make the perturbation effect  $\delta z_{ij}$ cancel out through the summation $\sum_{j=1}^J$, thereby leading to stable barycentric projection to the noise in $\ba$, which is the consequence of the law of large numbers assuming large $J$ values. In contrast, the selective effects $\boldsymbol{w}^{[q=1]}$ cannot cancel out the perturbation through the summation $\sum_{j=1}^J$, leading to an unstable response of $\delta\hat{\bx}$ to the noise. The cut-off function in the form of the optimal plan reduces the effective summation in $\sum_j$ to cancel the noise out, which also brings the instability in the noisy Barycentric projection.

We note that Lagrange multipliers $\boldsymbol{\alpha}$ and $\boldsymbol{\beta}$ and their responses to the noise $\Delta\boldsymbol{\alpha}$ and $\Delta\boldsymbol{\beta}$ depend on hyperparameters, i.e., $\reg$, $q$, and $m$. Consequently, their closed-form updates are not straightforward. Thus, based on our analysis, we cannot quantitatively state that the Weibull-ET always provides better robustness than traditional entropic optimal transport. %\pt{if we show that $\mu$ and $\lambda$ are bounded that we could argue that, I think. However, there if tensor is so large, that even sparse solution will have $|\Omega[1]|$ large enough for CLT to kick in}
The contribution of the above qualitative discussion is to identify a mechanism for modulating robustness for the barycentric projection under perturbations in the input marginals. 

Finally, we also comment here on our analysis for standard entropic optimal transport. Its decay is fixed to a selective exponential curve, and the local decay ratio is constant as $\rho(z)=1$, thus it has no modulation that enables local adaptive sparseness, whereas Weibull-ET and Tsallis-ET have $z$-dependent local decay rate $\rho_\chi(z)$. L$^2$-OT and $q$-DOT also have $z$-dependent local decay rate; however, as discussed in Section~\ref{sec:relatework}, they inherit the cutoff function in their form of plan, which may cause instability in the barycentric projections under noisy settings. In summary, the novelty of Weibull-ET is that it simultaneously satisfies all three properties: 
(i) tunability from slow to rapid decay, 
(ii) full support without a cutoff function, and 
(iii) adaptive local decay rate $\rho_\chi$.

\begin{figure*}[tbp]
  \centering
  \includegraphics[width=0.9\linewidth]{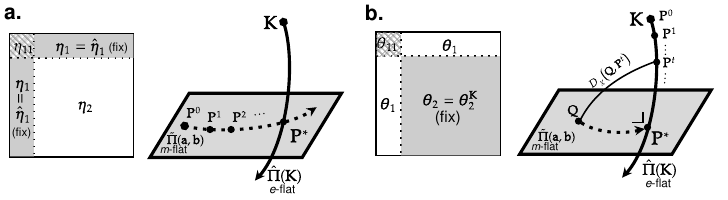}
  \caption{$\chi$-ET finds the intersection $\mP^*$ of two manifolds, $e_\chi$-flat manifold $\dualdomain{\mK}$ and $m_\chi$-flat manifold $\csimplex{\ba}{\bb}$. Compared with (\textbf{a}) learning the two-body expectation parameters $\bmeta_2$ while fixing the one-body expectation parameters $\bmeta_1$, (\textbf{b}) learning the one-body natural parameters $\bmtheta_1$ while fixing the two-body natural parameters $\bmtheta_2$ requires fewer parameters to be optimized. The solid arrows represent $e_\chi$-geodesics, while the dashed arrows represent $m_\chi$-geodesics.}
  \label{fig:sketch_optim}
\end{figure*}

\section{Algorithm derivation}\label{sec:optim}

In the following, we show that the projection theory in information geometry derives our algorithm introduced in Section~\ref{sec:algorithm} by leveraging the flatness of the feasible manifold $\csimplex{\ba}{\bb}$ in the coordinate system of the $\chi$-deformed exponential family.

%Optimizing the $\chi$-divergence is a central topic in the geometry of generalized exponential families, known as deformed information geometry, where various theories have already been discovered. We import these ideas and formulate entropic transport as a convex optimization problem. To this end, we first summarize the minimal results from deformed information geometry in Section~\ref{subsec:preliminaries_for_DIG}, and then derive the proposed algorithm in Section~\ref{subsec:natural_gradient}.

\subsection{Deformed information geometry on non-negative normalized matrices}\label{subsec:preliminaries_for_DIG}

%\textbf{Non-negative matrices as deformed exponential family}
We consider a parameterized probability distribution $p_{\bmtheta}(\bx)$ with random variable $\bx$ on the sample space $\mathcal{X}$, i.e., $p_{\bmtheta}:\mathcal{X}\rightarrow(0,1)$. 
For any escort function $\chi$ and appropriately defined tuple of functions $\boldsymbol{F}(\bx)=(F_1(\bx),\dots,F_N(\bx))^\top$ where each function $F_n$ is defined as $F_n:\mathcal{X}\rightarrow\mathbb{R}$, 
the family of distribution $p_{\bmtheta}(\bx)=\Exp_{\chi}\left[ \boldsymbol{F}(\bx)^\top\bmtheta -\psi_\chi(\bmtheta) \right]$ is called the $\chi$-deformed exponential family $\family_\chi$, where real-valued parameters $\bmtheta=(\theta_1,\dots,\theta_N)^\top$ are referred to as natural parameters~\citep{amari2021information}. The $\chi$-deformed free energy $\psi_\chi(\bmtheta)$ is a convex function with respect to the natural parameters $\bmtheta$ and are defined to normalize the distribution $p_{\bmtheta}$, i.e., $\int_{\bx\in\mathcal{X}} \Exp_{\chi}\left[ \boldsymbol{F}(\bx)^\top\bmtheta -\psi_\chi(\bmtheta) \right] d\bx=1$~\citep{naudts2011generalised}. As an example of a $\chi$-deformed exponential family in a discrete sample space $\mathcal{X}=[I]\times[J]$, assuming $\bx=(i,j)\in\mathcal{X}$, $\bmtheta=(\theta_{21},\dots,\theta_{IJ})$, $\theta_{11}=-\psi_\chi(\bmtheta)$, and $\mathbf{F}((i,j))=(\zeta^{ij}_{21},\dots,\zeta^{ij}_{IJ})^\top$, it recovers our model in Equation~\eqref{eq:parameterization}. Please refer to Section~\ref{sec:algorithm} for the definition of $\zeta_{kl}^{ij}$. Thus, the entire set of non-negative normalized matrices forms the $\chi$-deformed exponential family for any escort function $\chi$, which is consistent with the results in~\citep{amari2012geometry}. Throughout the following discussion, we assume the above discrete sample space $\mathcal{X}=[I]\times[J]$.

%\textbf{Dual coordinate system for non-negative matrices}
The Legendre transformation of the $\chi$-free energy $\psi_\chi(\bmtheta)$ is defined as $\phi_\chi(\bmeta)=\sup_{\bmtheta}\left(\bmtheta^\top\bmeta-\psi_\chi(\bmtheta)\right)$, where %the function $\varphi_\chi(\bmeta)$ and
parameter $\bmeta$ is referred to as %to as the $\chi$-entropy and 
expectation parameters, and its explicit form in our settings can be found in Equations%~\eqref{eq:chi_entropy} and
~\eqref{eq:parameterization}. %, respectively.
This transformation produces a dual relationship between natural parameters and expectation parameters, i.e., $\bmeta=\boldsymbol{\nabla}_{\bmtheta}\psi_\chi(\bmtheta)$ and $\bmtheta=\boldsymbol{\nabla}_{\bmeta}\phi_\chi(\bmeta)$ where $\boldsymbol{\nabla}_{\bmtheta}=(\partial_{\theta_{21}},\dots,\partial_{\theta_{IJ}})^\top$ and $\boldsymbol{\nabla}_{\bmeta}=(\partial_{\eta_{21}},\dots,\partial_{\eta_{IJ}})^\top$. %The explicit form of the natural parameter of our setting can be found in Equation~\eqref{eq:parameterization}. Given distribution $\mP$, its expectation parameter can be obtained by inclusion-exlcuiton principle, i.e., $\theta_{kl}=\cLog{\tP_{kl}}-\cLog{\tP_{k-1,l}}-\cLog{\tP_{k,l-1}}+\cLog{\tP_{k-1,l-1}}$. 
Any matrix $\mP$ in $\family_\chi$ can be identified by either natural parameters or expectation parameters; thus, they form coordinate systems of plan $\mP$, referred to as the $\bmtheta$-coordinate system and $\bmeta$-coordinate system, respectively.\footnote{Strictly speaking, the coordinate systems and parameters should be denoted by $\bmtheta_\chi$ and $\bmeta_\chi$, as they depend on the choice of $\chi$. For notational simplicity, however, we suppress the subscript $\chi$ of $\bmtheta$ and $\bmeta$.} We define the $\chi$-entropy $\varphi_\chi$ as $\varphi_\chi=\phi_\chi \circ H$, where the function $H:\family_\chi \rightarrow \RR_{\geq0}^{IJ-1}$ transforms $\mP$ to its $\bmeta$-representation, i.e., $H(\mP)=\bmeta^\mP$. The explicit form of the $\chi$-entropy can be found in Equation~\eqref{eq:chi_entropy}. While the Legendre dual $\phi_\chi$ is a convex function, the $\chi$-entropy $\varphi_\chi$ is not guaranteed to be a convex function due to the non-linearity of the function $H$. The Riemannian metric of the $\bmtheta$-coordinate system is given as the $\chi$-Fisher information matrix $\FIM(\bmtheta)=\boldsymbol{\nabla}_{\bmtheta}\boldsymbol{\nabla}^\top_{\bmtheta}\psi_\chi(\bmtheta)$ and its explicit form is given in Proposition~\ref{prop:FIM} in the Appendix. %, which recovers ordinarily Fisher-information matrix in $\chi(t)=t$~\citep{Ghalamkari2023Many}.

%\textbf{Projection theory in information geometry}
The $\chi$-divergence from $\mP\in\family_\chi$ to $\mK\in\family_\chi$ is defined as a Bregman divergence generated by the convex function $\phi_{\chi}(\bmeta)$ as 
%\begin{align*}
    $D_\chi[\mP;\mK] = \phi_\chi(\bmeta^\mP) - \phi_\chi(\bmeta^\mK) - \boldsymbol{\nabla}_{\bmeta} \phi_\chi(\bmeta^\mK)^\top (\bmeta^\mP - \bmeta^\mK)$,
%\end{align*}
%for $\mP, \mK\in\family_\chi$, 
where its explicit form can be found in Equation~\eqref{eq:chi_div}~\citep{Bregman1967relaxation}. %, which can be obtained by the closed form of the $\chi$-entropy $\varphi_\chi(\bmeta)=\sum_{ij}\cEsc{\mP_{ij}}\cLog{\mP_{ij}}$. 
We also define $e_\chi$-geodesic and $m_\chi$-geodesic from $\tP\in\family_\chi$ to $\tQ\in\family_\chi$ as $\{ t\bmtheta^{\mP} + (1-t)\bmtheta^{\mQ} \mid 0 \leq t \leq 1 \}$ and $\{ t\bmeta^{\mP} + (1-t)\bmeta^{\mQ} \mid 0 \leq t \leq 1 \}$, respectively. A manifold $\subfamily\subseteq\family_\chi$ is called $e_\chi$-flat if the $e_\chi$-geodesic between any two points in $\subfamily$ is contained entirely within $\subfamily$. Similarly, $\subfamily$ is called $m_\chi$-flat if the $m_\chi$-geodesic between any two points in $\subfamily$ is contained entirely within $\subfamily$. Any manifold defined by linear constraints in the $\bmtheta$- (resp. $\bmeta$-) coordinates is $e_\chi$-flat (resp. $m_\chi$-flat)~\citep{amari2016information}. For $m_\chi$-geodesic from $\tP$ to $\tQ$ and $e_\chi$-geodesic from $\tQ$ to $\tR$, they are said to be orthogonal if it holds that $(\bmtheta^{\tR}-\bmtheta^{\tQ})\cdot(\bmeta^{\tP}-\bmeta^{\tQ})=0$. In this case, the extended Pythagorean theorem $D_{\chi}[\mP;\mQ]+D_{\chi}[\mQ;\mR]=D_{\chi}[\mP;\mR]$ holds~\citep{Amari2011Geometry}.

The orthogonal projection of a point $\mP$ onto an $e_\chi$-flat manifold $\subfamily$ along an $m_\chi$-geodesic is called the $m_\chi$-projection, which is a convex optimization minimizing $D_{\chi}[\mP;\mQ]$ for $\mQ\in\subfamily$. Similarly, the orthogonal projection of a point $\mP$ onto an $m_\chi$-flat manifold $\subfamily$ along an $e_\chi$-geodesic is called the $e_\chi$-projection, which is a convex optimization problem that minimizes $D_{\chi}[\mQ;\mP]$ for $\mQ\in\subfamily$. 

The $\bmtheta$- and $\bmeta$-coordinate systems are orthogonal to each other, which means that we can define a mixture coordinate system by combining a part of the $\bmtheta$-coordinate and the $\bmeta$-coordinate system~\citep{amari2021information}. For example, a tuple of one-body expectation parameter $\bmeta_1=(\eta_{21},\dots,\eta_{1J})^\top$ and two-body natural parameter $\bmtheta_2=(\theta_{22},\theta_{23},\dots,\theta_{IJ})^\top$, which is referred to as $(\bmeta_1,\bmtheta_2)$, can uniquely identify the corresponding distribution $\tP$. In particular, the distribution $\tP$ is a singleton cross point of an $m_\chi$-flat manifold $\subfamily_1=\{ \tP \mid  \bmeta_1^\mP = \bmeta_1 \}$ and an $e_\chi$-flat manifold $\subfamily_2=\{ \tP \mid \bmtheta_2^\mP = \bmtheta_2 \}$.
\begin{figure*}[t]
  \centering
  \begin{minipage}{0.6\linewidth}
    \includegraphics[width=\linewidth]{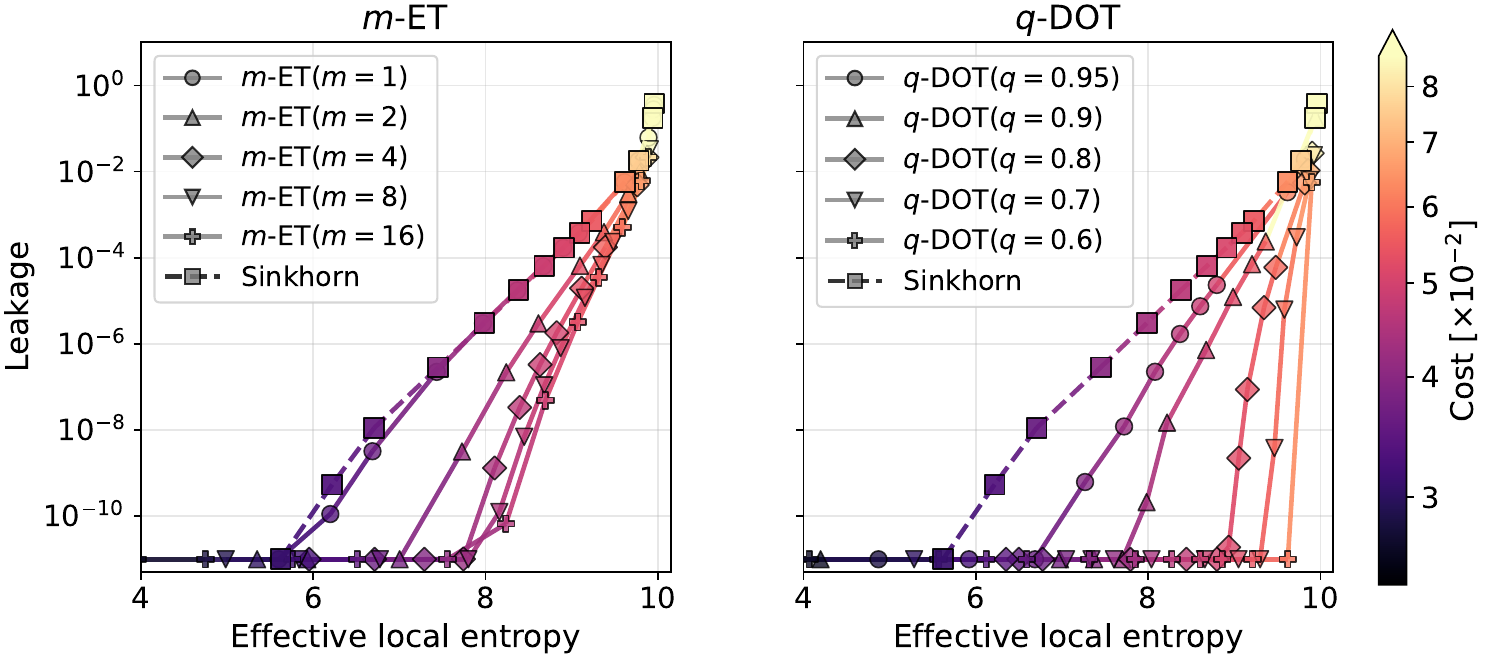}
  \end{minipage}
  \hfill
  \begin{minipage}{0.38\linewidth}
    \caption{
    Trade-off between leaked mass in transportation among inter-clusters and local softness evaluated by effective local entropy of Weibull-ET (left) and $q$-DOT (right). Colors indicate the transport cost. The color scale is clipped at the 80th percentile to improve visibility; values above this threshold are saturated.}
    \label{fig:q_dot_vs_m_et}
  \end{minipage}
\end{figure*}

\begin{figure*}[t]
  \centering
  \includegraphics[width=0.99\linewidth]{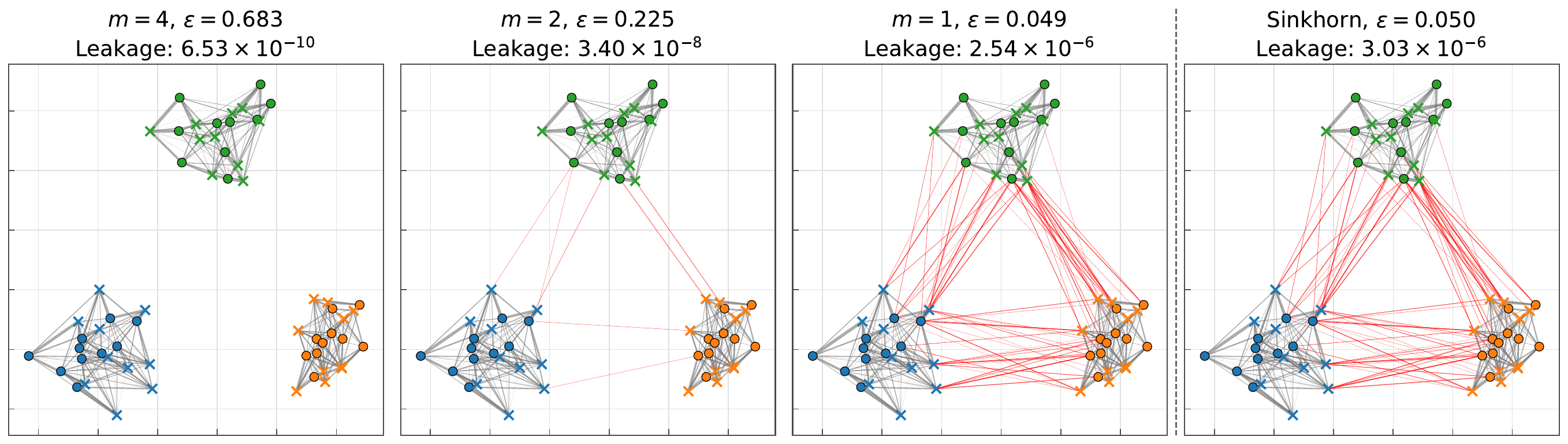}
  \caption{
    Transport plans obtained using the Sinkhorn method (rightmost panel) and the proposed Weibull-ET with various values of $m$ (other panels). For each method, the regularization strength $\reg$ is tuned so that the effective local entropy is approximately $8.0$. The hyperparameter $m$ controls the leakage level while maintaining approximately the same local softness, which is consistent with the results in Figure~\ref{fig:q_dot_vs_m_et}.
  }
  \label{fig:softness}
\end{figure*}

%\subsection{Gradient method for \texorpdfstring{$\chi$}{Chi}-ET}\label{subsec:natural_gradient}
\subsection{Optimization for \texorpdfstring{$\chi$}{Chi}-ET}\label{subsec:natural_gradient}
Given the $\chi$-deformed kernel matrix $\mK$, we consider the optimization in Equation~\eqref{eq:reform} and derive the algorithm presented in Section~\ref{sec:algorithm}. %given $\mK$ since it is equivalent with $\chi$-entropic escort transport. 
The key observation is that the feasible manifold $\csimplex{\ba}{\bb}$ in Equation~\eqref{eq:simplex}
is an $m_\chi$-flat manifold because the $\chi$-balancing condition can be written as a linear condition on the expectation parameters: 
\begin{align}\label{eq:balancing_condition_with_eta}
    \csimplex{\ba}{\bb} = \set{\mP \mid 
    \eta^{\mP}_{i1} = \sum_{i'=i}^{I} a_{i^\prime}, \ 
    \eta^{\mP}_{1j} = \sum_{j'=j}^{J} b_{j^\prime} \ 
    }.
\end{align}
Therefore, the $\chi$-ET is an $e_\chi$-projection onto an $m_\chi$-flat manifold, which is a convex optimization that minimizes $D_\chi[\mP,\mK]$ for $\mP \in \csimplex{\ba}{\bb}$. This statement is a natural generalization of the flatness of the feasible set of the ordinary optimal transport, discussed by~\citet{amari2018information}. As seen in Figure~\ref{fig:sketch_optim}(a), we update only the two-body expectation parameters $\bmeta_2$ while fixing the one-body expectation parameters $\bmeta_1$ by the constraint in Equation~\eqref{eq:balancing_condition_with_eta}, thus the number of parameters to be optimized is $L=(I-1)(J-1)$. We recall that $\eta_{11}$ and $\theta_{11}$ are not free-parameters as discussed in Section~\ref{sec:algorithm}. As seen in Proposition~\ref{prop:grad1} in the Appendix, the derivative of the $\chi$-divergence is given as $\boldsymbol{\nabla}_{\bmeta_2} D_{\chi}(\mP,\mK) = \bmtheta_2 - \bmtheta^{\mK}_2$. Thus, we obtain the update rule as
\begin{align}\label{eq:prime_update}
    \bmeta_{2} \gets \bmeta_{2} - \Tilde{\FIM}^{-1} \left(\bmtheta_2 - \bmtheta^{\mK}_2 \right),
\end{align}
where $\Tilde{\FIM}$ is the appropriately obtained second-order derivative of the objective function, namely the $L\times L$ Hessian matrix, and its inverse has complexity of $O(I^3J^3)$. 

To reduce the number of parameters to be optimized, we consider a dual problem using a mixture coordinate system. We focus on the critical condition $\bmtheta^{\mP^*}_2=\bmtheta^{\mK}_2$ at the optimal $\chi$-transport plan $\mP^*$ assuming the gradient $\boldsymbol{\nabla}_{\bmeta_2} D_{\chi}(\mP^*,\mK)$ is $0$ at $\mP^*$ because Proposition~\ref{prop:barrier} in Appendix~\ref{sec:proofs} ensures a relative-interior solution. Thus, the plan $\mP^*$ is identified with a mixture representation $(\bmeta_1,\bmtheta^{\mK}_2)$ of one-body expectation parameters $\bmeta_{1}$ (the $\chi$-balancing condition) and two-body natural parameters $\bmtheta^{\mK}_2$ (the critical condition). In other words, the optimal escort plan $\mP^*$ is uniquely determined as the intersection of the $m_\chi$-flat manifold $\csimplex{\ba}{\bb}$ and the $e_\chi$-flat manifold $\dualdomain{\mK}$, where $\dualdomain{\mK}$ denotes the set of distributions whose two-body natural parameters are constrained to match those of the matrix $\mK$, i.e., 
\begin{align}\label{eq:dualdomainconstrain}
    \dualdomain{\mK}=\set{\mP \mid \bmtheta_2^{\mP} = \bmtheta_2^{\mK}}.
\end{align}
The $e_\chi$-flatness of the manifold $\dualdomain\mK$ is straightforward since its constraint is linear with respect to the natural parameter $\bmtheta$. Therefore, we update $\mP$ along the $e_\chi$-geodesic within $\dualdomain\mK$ to find its intersection with $\csimplex{\ba}{\bb}$ as seen in Figure~\ref{fig:sketch_optim}(b). Specifically, we update $\bmtheta_1$ while fixing the two-body natural parameters $\bmtheta_2$ at $\bmtheta^{\mK}_2$. For the distribution $\mP$ at iteration $t$, the extended Pythagorean theorem shows that
\begin{align}
    D_{\chi}[\mQ; \mP^{*}] + D_{\chi}[\mP^{*}; \mP^{t}] = D_{\chi}[\mQ;\mP^t],
\end{align}
for any $\mQ\in\csimplex{\ba}{\bb}$~\citep{amari2000methods}. It is evident from Figure~\ref{fig:sketch_optim}(b) that minimizing $D_\chi[\mQ;\mP^t]$ for $\mP^t\in\dualdomain\mK$ makes the plan $\mP^t$ converge to the optimal plan $\mP^*$. Therefore, the optimization is surrogated into 
\begin{align*}
    \mP^* = \argmin_{\mP\in\dualdomain\mK} D_\chi[\mQ; \mP].
\end{align*}
As seen in Proposition~\ref{prop:grad2}, the first-order derivative of the objective is given as $\boldsymbol{\nabla}_{\bmtheta_1} D_{\chi}[\mQ;\mP]=\bmeta^{\mP}_{1}-\bmeta^{\mQ}_{1}=\bmeta^{\mP}_{1}-\hat{\bmeta}_{1}$ where we recall the constraint on the one-body expectation parameter of $\mQ\in\csimplex{\ba}{\bb}$ given in Equation~\eqref{eq:onebody_etahat}. The second-order derivative of the $\chi$-divergence is given in Equation~\eqref{eq:FIM}, which is derived in Proposition~\ref{prop:FIM}, and thus, we finally obtain the update rule given in Equation~\eqref{eq:update_rule}. The resulting complexity is $O((I+J)^3)$ due to the inverse of the $\FIM$. The definition of the Bregman divergence guarantees the positive definiteness of the matrix $\FIM$~\citep[Definition~1.1]{amari2016information}. Moreover, each element of the matrix $\FIM$ in Equation~\eqref{eq:FIM} is bounded. Therefore, the first-order derivative $\bmeta^{\mP}_{1}-\hat{\bmeta}_{1}$ is Lipschitz continuous, which, together with the convexity of the objective, ensures the convergence of the proposed algorithm to the global optimum~\citep{boyd2004convex}.
%ensuring the global optimality of the proposed algorithm

Approximating a matrix while fixing some of its natural parameters is known as the Legendre decomposition~\citep{Sugiyama2018Legendre, Zhou2025Optimal, Zhou2026Generalized}. Since the constraint on $\dualdomain{\mK}$ in Equation~\eqref{eq:dualdomainconstrain} fixes the two-body natural parameters, we solve $\chi$-ET by the generalized Legendre decomposition under an escort function. 

\textbf{Convergence and stopping criterion \ \ }
We again emphasize that the proposed $\chi$-ET finds the globally optimal $\mP$ that minimizes Equation~\eqref{eq:reform} since it is a projection onto a flat manifold. %Our method, based on a second-order derivative, requires fewer steps to convergence than first-order methods, such as the Sinkhorn method~\citep{Cuturi2013Sinkhorn}. 
The computational bottleneck of the proposed method stems from the inverse of the matrix $\FIM$, which as noted requires $O((I+J)^3)$. This bottleneck can be removed by the LBFGS method without losing the global optimality guarantees~\citep{liu1989limited}. %, but it also increases the number of iterations required for convergence. 
When the update rule in Equation~\eqref{eq:prime_update} is used, the iterate $\mP$ evolves on the feasible manifold $\csimplex{\ba}{\bb}$ as seen in Figure~\ref{fig:sketch_optim}(a). In particular, the $\chi$-balancing condition holds at every iteration. In contrast, when the update rule in Equation~\eqref{eq:update_rule} is used \GK{to reduce the computational complexity}, the feasibility is not preserved in each iteration, as seen in Figure~\ref{fig:sketch_optim}(b). Therefore, we need to run the algorithm until the constraint residual becomes small, i.e.,
$
\bigl\|\bmeta_1 - \hat{\bmeta}_1\bigr\| < \delta,
$
for a prescribed tolerance $\delta>0$. \GK{For clarity, both updates in Equations~\eqref{eq:update_rule} and~\eqref{eq:prime_update} find the identical unique optimal solution $\mP^*$. }%, which is a well-known result in convex analysis as duality of the optimization.}

%Rather than naively optimizing $D_\chi(\mP,\mK)$ given matrix $\mK$, the above trick reduces the its complexity by monotonically decreasing $D_\chi(\mR,\mP)$ for $\mP$ where $\mR$ is any matrix in $\dualdomain{\mK}$, which is formally shown in Proposition X using generalized Pythagorean theorem. 

%\textbf{Remark on normalization \ \ }
%Since our framework relies on information geometry, which is the geometry used for probability distributions, we need the normalization of both $\mP$ and $\mK$ in the objective function $D_\chi[\mP;\mK]$ in Equation~\eqref{eq:reform}. If we follow the \citet{Cuturi2013Sinkhorn}'s formulation and define the Kernel matrix as $\mK_{ij}=\cExp{-\cost_{ij}/\reg}$, the normalizing condition does not hold. If we naively normalize it by dividing all elements of $\mK$ by the total sum of $\mK$, the optimal solution $\mP$ is not guaranteed to be that of the original objective. On the other hand, our Theorem~\ref{th:solution_invariance} in the Appendix shows that the constant $g$ in Equation~\eqref{eq:Kernel} enables the normalization without changing the optimal $\mP$, which is also our contribution. %. To our best knowledge, the fact that $m$-normalization does not change the optimization results has not yet been reported.

\begin{figure*}[t]
  \centering
  \includegraphics[width=0.99\linewidth]{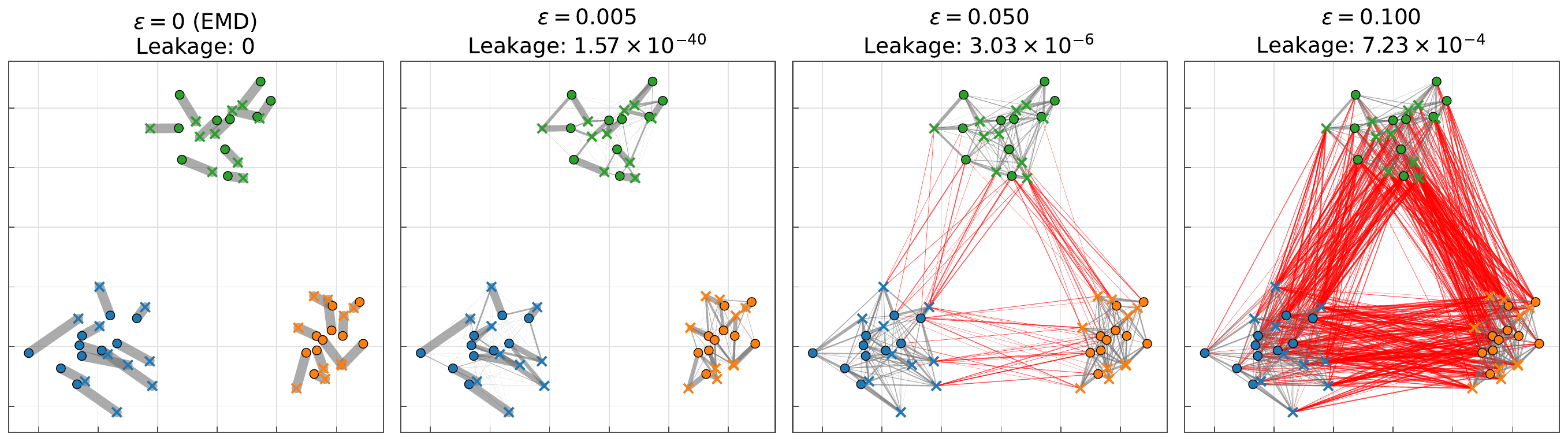}
  \caption{
  The effect of the value of $\reg$ in Shannon's entropic optimal transport from the three synthesized clustered sources to the targets. %, the optimal transport without regularizer (EMD) as a limit of $\varepsilon\rightarrow0$. 
  Gray and red lines represent intra- and inter-cluster transport paths, respectively, and line width corresponds to the amount of transported mass. Smaller $\reg$ makes the transport globally sparse, and it cannot maintain local softness. %While only the top 50\% of the transported mass is shown to improve visibility, Both Sinkhorn ($\reg>0$) and $\chi$-ET offer only full support plans, which are unable to prevent inter-cluster transport completely.
  }
  \label{fig:sinkhorn_plans}
\end{figure*}

\begin{figure*}[p]
  \centering
  \includegraphics[width=0.99\linewidth]{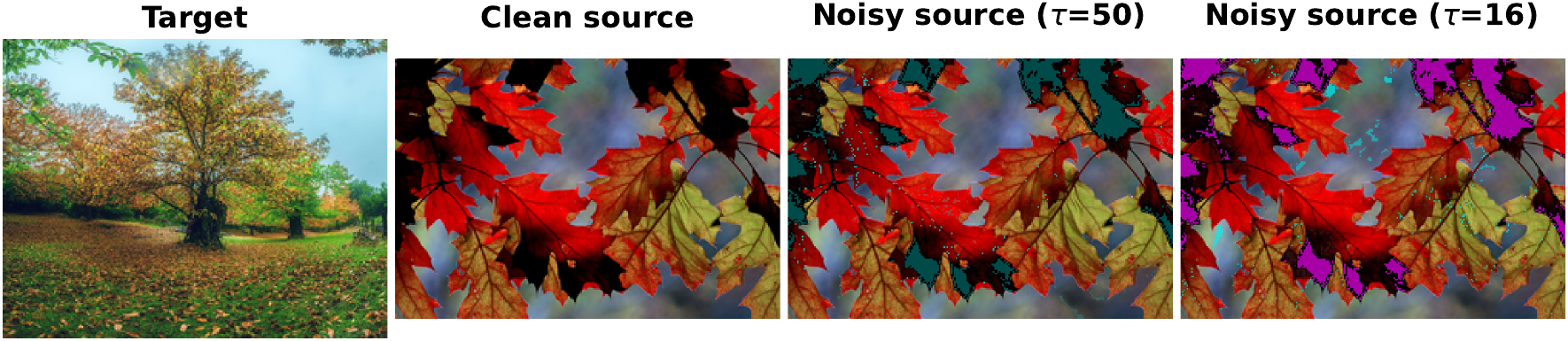}
  \caption{
  Target and source images used for color transfer. Source image overlaid with Dirichlet noise-induced changes in the color histogram, where opacity reflects the magnitude of histogram mass change.}
  \label{fig:input_imgs}
\end{figure*}

\begin{figure*}[p]
  \centering
  \includegraphics[width=0.99\linewidth]{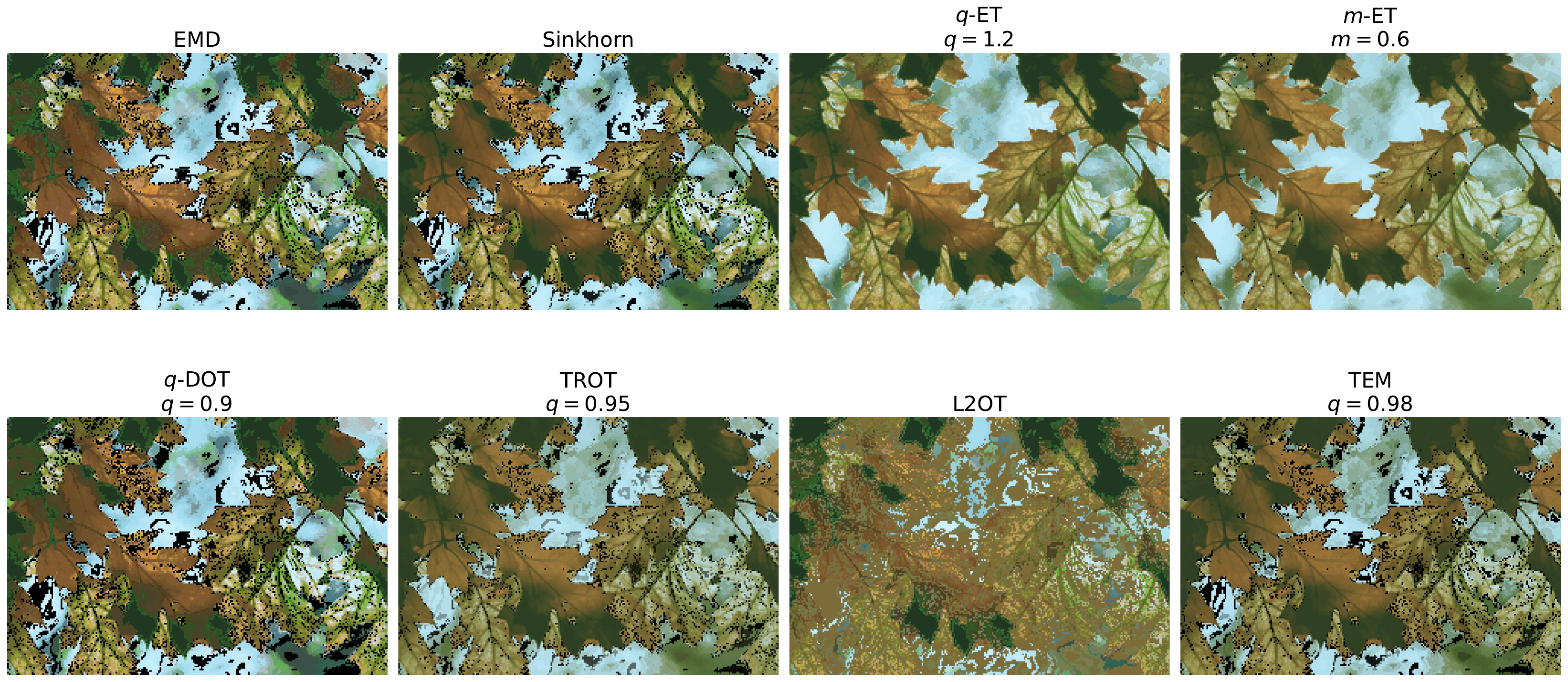}
  \caption{Color-transferred source images with Dirichlet noise $\tau=50$ and chosen hyper parameters}
  \label{fig:color_transport50}
\end{figure*}

\begin{figure*}[p]
  \centering
  \includegraphics[width=0.99\linewidth]{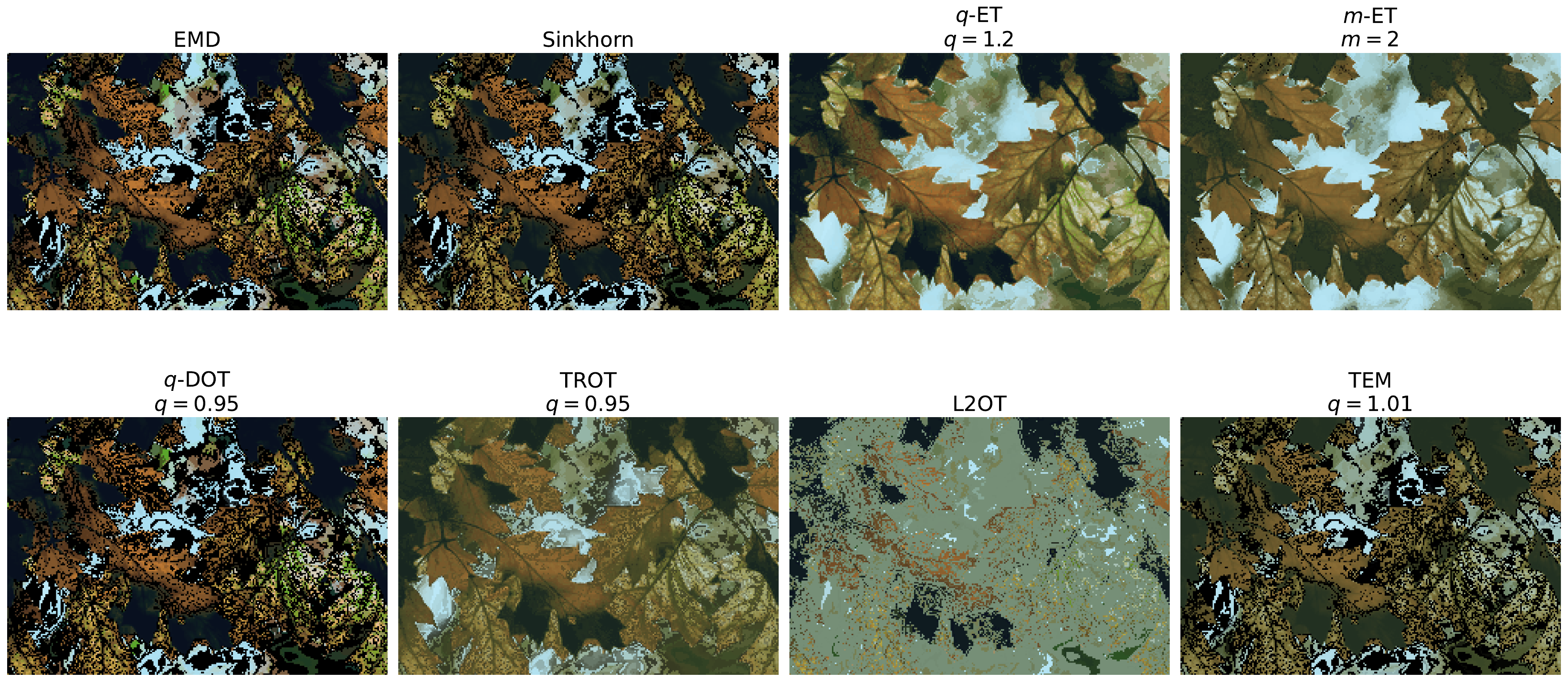}
  \caption{Color-transferred source images with Dirichlet noise $\tau=16$ and chosen hyper parameters}
  \label{fig:color_transport16}
\end{figure*}

\begin{figure*}[tbp]
  \centering
  \includegraphics[width=0.99\linewidth]{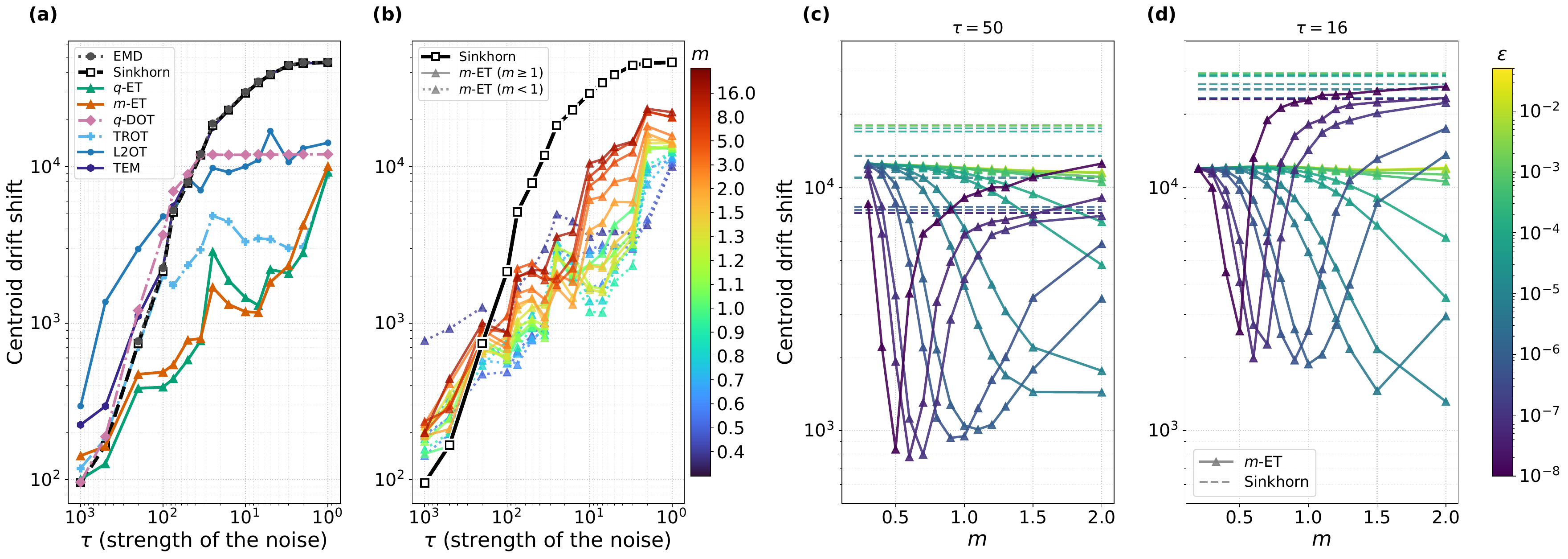}
  \caption{Effect of noise in color histogram $\ba$ on the color-transfer stability, measured by the centroid drift shift of color clusters in the color-transferred source images. (\textbf{a}) Dependence on the Dirichlet noise parameter $\tau$ for different optimal transport methods. (\textbf{b}) Corresponding results for the proposed Weibull-ET with varying values of $m$. (\textbf{c},\textbf{d}) Summary of how the centroid drift shift varies with the tail-control parameter $m$ and the regularization strength $\varepsilon$ in Weibull-ET.}
  \label{fig:main_exp}
\end{figure*}

\section{Numerical evaluation}\label{sec:experiments}

We evaluate the local adaptivity of Weibull-ET using a synthesized dataset in Section~\ref{ref:exp_local_softness}, directly observe the decay profile of the obtained plan in Section~\ref{sec:exp_decay}, and verify the robustness of the barycentric projection based on $\chi$-ET using color images in Section~\ref{sec:color_transport}. %We again emphasize that our contribution is not merely introducing Weibull-ET but also developing a general algorithm for $\chi$-ET and evaluating the practical benefits of a full-support tunable decay plan for $\chi$-ET. 
Please refer to the discussion in Sections~\ref{sec:escort} and~\ref{sec:tail} for the motivation that we focus on Tsallis-ET and Weibull-ET, but not Kaniadakis-ET. The basic properties of the baseline methods can be found in Table~\ref{tab:comparison}. Once again, while Tsallis-ET has already been formulated in~\cite{kurose2022optimal}, its practical advantages and feasible algorithm have not been discussed yet; thus, our contribution includes analyzing the existing Tsallis-ET.

\subsection{Local softness induced by larger \texorpdfstring{$m$}{m}}\label{ref:exp_local_softness}

To observe the property of the Weibull-ET, we synthesized three-clustered targets and sources in two-dimensional space by three normal distributions with different mean values. In particular, we sample $10$ points from each normal distribution to generate each cluster. We set uniform weights on each point as $a_i=b_j=1/N$ for the number of targets and sources $N=30$. %The dataset generation can be found in Appendix~\ref{sec:implement_detail}. 
The cost matrix $\cost$ is defined with squared distance in the two-dimensional space. We then performed optimal transport with Weibull-ET using Algorithm~\ref{alg:ET}, varying the hyperparameter $m$. To evaluate the local softness of the obtained transport plans quantitatively, we normalized row-wise each cluster block of the plan and computed the Shannon entropy within each cluster. We then used the exponential of the obtained entropy %as an effective number of intra-cluster destinations 
and averaged it over points and clusters, named \emph{effective local entropy}. 
\begin{align*}
  H_{\mathrm{eff}}(\mR) = \frac{1}{I} \sum_{i=1}^{I}
  \exp\Bigl( -\sum_{j \in \mathcal{T}(i)} p_{ij} \log p_{ij} \Bigr),
  \qquad
  p_{ij} = \frac{\mR_{ij}}{\sum_{j' \in \mathcal{T}(i)} \mR_{ij'}},
\end{align*}
where $\mathcal{T}(i)$ is the set of targets in the same cluster as source $i$. In Figure~\ref{fig:q_dot_vs_m_et}(left), we plot the effective local entropy while varying $\reg$ and $m$. The vertical axis represents the total mass transported between different clusters, defining the leakage. The proposed Weibull-ET with $m>1$ performs a better trade-off between the effective local softness and leakage when compared to the entropic optimal transport (Sinkhorn). We also note that the local softness can be improved further by existing $q$-DOT because it admits zero value in the plan, while Weibull-ET provides a full support plan. The trade-off between leakage and local softness of $q$-DOT varying $\reg$ and $q$ can be found in Figure~\ref{fig:q_dot_vs_m_et}(right). Once again, the cutoff function in $q$-DOT may prevent noise robustness in the barycentric projection, as discussed in Section~\ref{sec:tail}, which is also numerically demonstrated in Section~\ref{sec:color_transport}. Our contribution is balancing the wider tunability of the shape of decay, ensuring full-support of the plan without a cutoff function, but not solely achieving local softness.

We also determined the regularization strength $\reg$ by bisection such that the effective local entropy is approximately $8.0$ and visualized the resulting transport plans in Figure~\ref{fig:softness}, where the gray lines indicate transport within clusters and the highlighted red lines indicate transport between clusters. The results demonstrate that the leakage can be controlled by varying $m$ while maintaining approximately the same local softness. The results on the same dataset using entropic optimal transport with the Sinkhorn algorithm are presented in Figure~\ref{fig:sinkhorn_plans} with various $\reg$. The results demonstrate that the tuning of $\reg$ affects the transportation globally, i.e., smaller values of $\reg$ prevents leakage but also enhances sparseness (hard transport) within the cluster, which is consistent with their constant local decay ratio $\rho(z)=1$ as discussed in Section~\ref{sec:method}. %The obtained plan $\mR$ varying $m$ but fixing $\reg$ is visualized in Figure~\ref{fig:headline}(\textbf{b}), which suggests that a larger $m$ reduces leakage but keeps dense (soft) transport within each cluster. 

%These results demonstrate that the leakage can be controlled by varying $m$ while maintaining approximately the same local softness, whereas the conventional Sinkhorn method cannot control the leakage independently of the local softness.

\subsection{Rapid and slow decay controlled by \texorpdfstring{$m$}{m}}\label{sec:exp_decay}

\begin{figure*}[t]
  \centering\vspace{-2em}
  \includegraphics[width=0.959\linewidth]{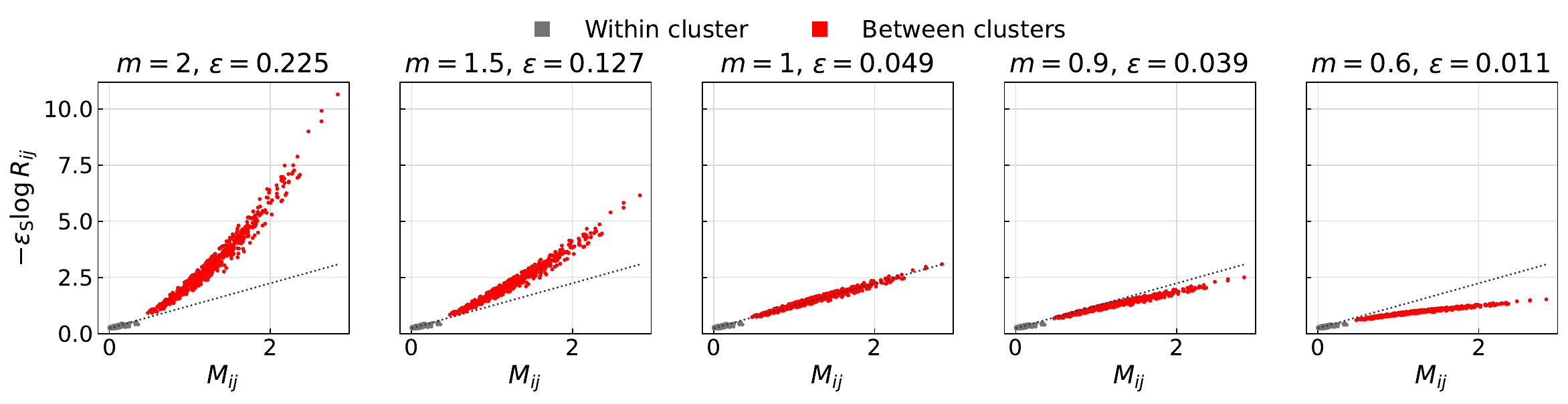}
\caption{Decay of the Weibull-ET plan $\mathbf{R}$ against the cost $\mathbf{M}$ for various $m$, with $\varepsilon$ tuned so that the effective local entropy is approximately 8.0. The dotted line represents the exponential decay of Sinkhorn at the same effective local entropy. Entries above and below this line indicate rapid ($m > 1$) and slow ($m < 1$) decay, respectively.}\vspace{-0.5em}
  \label{fig:mET_hist_part}
\end{figure*}

To directly observe how the regularization shapes the decay profile of the entries of the optimal plan $\mR$, we additionally ran the proposed Weibull-ET on the same synthesized datasets described above and plot $- \reg_S \log \mR_{ij}$ against $\cost_{ij}$ in Figure~\ref{fig:mET_hist_part} with various $m$ values, where each regularization strength $\reg$ is determined such that the effective local entropy is approximately $8.0$ and $\reg_S=0.05$ is the regularization strength at which Sinkhorn attains the same effective local entropy. To compare with the results by the entropic regularization, the plot includes a dotted line of unit slope whose intercept we fit to the within-cluster entries of the optimal plan obtained by the Sinkhorn algorithm. Since Proposition~\ref{prop:decay} states $-\reg_S \log{\mR_{ij}} = \cost_{ij} + \reg_S(\alpha_i + \beta_j)$, the entries below the dotted line for $m < 1$ and above it for $m > 1$ indicate slower and more rapid decay than the exponential decay, respectively. The vertical scatter at a fixed cost reflects the Lagrange multipliers $\boldsymbol{\alpha}$ and $\boldsymbol{\beta}$. The complete comparison of the tail of the Weibull-ET to the entropic regularization including a detailed discussion can be found in  Figures~\ref{fig:sinkhorn_hist} and~\ref{fig:mET_hist} in  Appendix~\ref{sec:add_exp}.

%Finally, in Figures~\ref{fig:sinkhorn_hist} and~\ref{fig:mET_hist} in Appendix~\ref{sec:add_exp}, we illustrate on the above synthetic dataset that the hyperparameter $m$ controls the decay of the plan from slow to rapid, while changing $\reg$ in Sinkhorn keeps it exponential.

\subsection{Noise robust barycentric projection for color transfer}\label{sec:color_transport}

We evaluate the robustness of the proposed method to noise in the input histogram $\ba$ through a color transfer task, which is a classical application of optimal transport under domain shift~\citep{Pitie2007linear}. We follow the standard barycentric projection framework commonly used for color transfer. As in previous work in \citep{Blondel2018Smooth}, we use two public-domain images: Fall Foliage by Bernard Spragg and Communion by Abel Maestro Garcia as seen in Figure~\ref{fig:input_imgs}. Each pixel is represented as a point in the RGB space, and we apply $k$-means clustering with $k = 512$ to obtain color centroids. From the cluster assignments, we construct a 512-dimensional source color histogram $\ba$ and target color histogram $\bb$. We define the cost matrix $\cost$ as the squared distance between centroids in the RGB space.

To assess robustness, we perturb the source histogram $\ba$ by adding Dirichlet noise $\tau$, where a larger $\tau$ value indicates small noise effects. We obtain the regularized transport plan using Algorithm~\ref{alg:ET} and baselines, and recolor the source image by shifting its color clusters via barycentric projection. The shifted color centroid follows Equation~\eqref{eq:Barycenter}. For the quantitative discussion of the effect of the noise, we define centroid drift shift as
\begin{align}\label{eq:CDS}
    \Delta^{\tau}_{\mathrm{method}}
    = \frac{1}{K}\sum_{k=1}^K a_k \| \by^{(\mathrm{clean},k)}_{\rm EMD} - \by^{(\tau,k)}_{\mathrm{method}} \|^2
\end{align}
where $\by^{(\tau,k)}_{\mathrm{method}}$ is the RGB value of the centroid of the $k$-th color, obtained by barycentric projection by the plan obtained by the method with a noisy source image with the noise $\tau$, and $a_k$ is the normalized number of pixels associated with the $k$-th color center. Our setup fixes $K$ as $512$. Similarly, $\by^{(\mathrm{clean},k)}_{\mathrm{EMD}}$ denotes the obtained RGB value of the centroid of the $k$-th color by EMD with a clean source image. The smaller drift shift implies less effect of the noise in the barycentric projection. While our experiments evaluate the centroid drift shift in Equation~\eqref{eq:CDS} to investigate the noise robustness of barycentric projection, the clean barycenter $\by^{(\mathrm{clean},k)}$ is not available in a practical setting. Thus, the hyperparameter is expected to be tuned based on an appropriately defined score of the downstream task. 
%As a remark, Equation~\eqref{eq:CDS} is identical to the pixel-wise squared error between the image color-shifted by EMD without noise and that by each method with noise.
As a remark, Equation~\eqref{eq:CDS} is identical, up to a multiplicative constant, to the pixel-wise squared error between the image color-transferred by noise-free EMD and that color-transferred by each method under noise.

The color-transferred source images are visualized in Figures~\ref{fig:color_transport50} and~\ref{fig:color_transport16}, where the hyperparameters of each method are chosen to minimize the centroid drift shift. Only the Tsallis-ET and Weibull-ET produce visually stable results, while baselines with the cutoff function demonstrate sensitivity to the perturbation. We also change the value of noise strength $\tau$ and observe its effects on the centroid drift shift in Figure~\ref{fig:main_exp}(\textbf{a}), which consistently shows strong robustness of Tsallis-ET and Weibull-ET against noise. While TROT also demonstrates comparable performance in extremely noisy settings ($\tau<10$), we emphasize that, to the best of our knowledge, it has no guarantee of global convergence, as discussed in Section~\ref{sec:relatework}. The hyperparameter tuning is extensively performed as described in Appendix~\ref{sec:implement_detail}. We also investigate the optimal hyperparameter $m$ of Weibull-ET that minimizes the centroid drift shift $\Delta^{\tau}_{m\mathrm{-ET}}$, varying the noise strength $\tau$. The results in Figure~\ref{fig:main_exp}(\textbf{b}) demonstrate that the favorable decay regime varies with the noise level. In particular, Weibull-ET outperforms Tsallis-ET for $16<\tau<35$ using full-support rapid decay, which lies outside the scope of Tsallis-ET. %This supports the benefit of tuning the decay across both slow- and rapid-decay regimes rather than universally favoring broader averaging.
%The results in Figure~\ref{fig:main_exp}(\textbf{b}) demonstrate that Weibull-ET offers better noise robustness than baselines in both regimes of rapid decay ($m>1$) and slow decay ($m<1$), and the optimal $m$ depends on the noise strength $\tau$. In the ranges $\tau<10$ and $35<\tau<100$, the optimal $m$ is larger than one, indicating that slow decay improves robustness. In contrast, in the range $16<\tau<35$, the optimal $m$ is smaller than one, indicating that local softness induced by rapid decay may effectively enhance robustness. 

For fixed noise strength $\tau$, we also observed how the strength of the regularization $\reg$ effects the optimal value of $m$ for fixed $\tau=50$ in Figure~\ref{fig:main_exp}(\textbf{c}) and for $\tau=16$ in Figure~\ref{fig:main_exp}(\textbf{d}). The full results with various values of $\tau$ can be found in Figure~\ref{fig:all_mET} in Appendix~\ref{sec:add_exp}. It shows that the optimal $m$ increases with $\reg$. We also emphasize that Weibull-ET with $m=1$ in Figure~\ref{fig:main_exp} does not recover Sinkhorn's result because the $m=1$ approach does not yield the ordinary exponential and logarithm functions as discussed in Section~\ref{sec:escort}.

This noise robustness discussed above cannot be attributed solely to the full-support of the optimal plan, as entropic optimal transport (Sinkhorn) also guarantees full-support, yet it does not demonstrate the noise robustness. We note that robustness can also be introduced by defining centroid distances with alternatives to the squared Euclidean distance. However, in such cases, the barycentric projection no longer admits a closed-form solution, leading to additional complexity. %Moreover, a heuristic of powering $q$ of the obtained transport plan fails to preserve the marginal constraints. As a result, this approach becomes unstable and performs poorly in high-noise regimes.

\section{Conclusion}\label{sec:conclude}

Our derived algorithm solves $\chi$-ET as a $\chi$-divergence optimization problem. The proposed Weibull escort $\chi_m$ enables full-support optimal transport with not only a slow but also a rapid-decay regime, which is unavailable to the Tsallis and Kaniadakis escorts, while keeping closed forms of the associated deformed exponential and logarithm functions. Our empirical results verify that the tunable decay brings practical benefits such as noise-robust barycentric projections and local softness, and our first-order perturbation analysis identifies a mechanism for this modulation. 
%Moreover, we introduce a Weibull escort $\chi_m$ that reaches a rapid-decay full-support regime unavailable to the Tsallis and Kaniadakis escorts. We also emphasize that the introduced Weibull escort function ensures the closed-forms of associated deformed exponential and logarithm functions. % and evaluate the resulting methods systematically.
%This work provides a practical algorithm of $\chi$-ET, firstly formulated by~\citet{kurose2022optimal}, and introduces a novel escort function $\chi_m$ that enables us to leverage both rapid and slow decays of the optimal plan without a cut-off function while ensuring global optimality and closed-forms of deformed exponential and logarithm functions. 
%In addition, we investigate how the tunable decay brings practical benefits such as noise robust barycentric projections and local softness. These effects arise from $\chi$-entropic regularization, which exposes the deformed rank-$1$ structure on the $\chi$-inverse escort plan $\mP$ in the cost-free limit of $\reg\rightarrow\infty$. %Notably, our contribution includes pointing out that the naive formulation for Tsallis-ET encounters a domain issue and prohibits a slower decay plan.
%Our work attempts to bridge the gap between the elegant information-geometric structure induced by escort entropic regularization, discovered by~\citet{kurose2022optimal}, and its practical usefulness in the machine learning community, and we hope that the robust properties highlighted will provide a foundation for the wider adoption of Escort Entropic Optimal Transport.
This work takes a step toward bridging the gap between the elegant information-geometric structure of escort entropic regularization, discovered by~\citet{kurose2022optimal}, and its practical utility for the machine learning community. We hope that the robustness demonstrated in this work provides a foundation for the broader adoption of escort entropic optimal transport.

\bibliography{main}
\bibliographystyle{tmlr}

\newpage
\onecolumn
\title{Supplementary Material for Escort Transport}
{\centering\huge\bfseries Supplementary Material}
\addcontentsline{toc}{part}{Supplementary Material}
\maketitle
\appendix

\begin{table}[h!]
\centering
\renewcommand{\arraystretch}{1.2}
\caption{List of the notations used throughout the paper}\label{tb:notation}
\begin{tabular}{ll}
\toprule
\textbf{Symbol} & \textbf{Description} \\
\midrule
$ [n] $ & The set of natural number smaller than $n$, i.e., $[n]=\{1,2,\dots,n\}$ \\
$\mathcal{X}=[I]\times[J]$ & Index set of the optimal plan $\mR$ \\
$\mathbb{R}, \mathbb{R}_{> 0},  \mathbb{R}_{\geq 0}$ & Sets of real, strictly positive, and non-negative real numbers, respectively \\
$\ba\in\RR_{\geq 0}^{I}, \bb\in\RR_{\geq 0}^{J}$ & Positive normalized source and target marginals\\
$\mM\in\RR_{\geq 0}^{I\times J}$ & Cost matrix, i.e., $\mM_{ij}=d(\bx^{(i)},\by^{(j)})$\\
$\langle \mM, \mP\rangle$ & Frobenius inner product \\
$\reg > 0$ & Regularization strength \\
$\mP\in\RR_{\geq 0}^{I\times J}$ & Inverse escort plan, optimized in the information-geometric formulation \\
$\mR\in\RR_{\geq 0}^{I\times J}$ & Actual transport plan, i.e., $\mR_{ij}=\Tilde{\chi}[\mP]_{ij}$ \\
$\mK\in\RR_{\geq 0}^{I\times J}$ & $\chi$-deformed kernel matrix, i.e., $\mK_{ij} = \cExp{-\mM_{ij}/\varepsilon+g}$ \\
$g\in\RR$ & $e$-normalizing constant chosen so that $\sum_{ij}\mK_{ij}=1$ \\
$\simplex{\ba}{\bb}$ & Transportation polytope \\
$\csimplex{\ba}{\bb}$ & $\chi$-transportation polytope\\
$\tilde{\chi}[\mP]$ & $\chi$-escort of $\mP$, i.e., $\tilde{\chi}[\mP]_{ij}=\chi[\mP_{ij}]/\sum_{ij}\chi[\mP_{ij}]$ \\
%$\mR_{(\ell)}$ & The $\ell$-th largest entry of $\mR$\\
$\Omega(\mR)$ & Regularizer in a regularized OT objective.\\
$\chi:\RR_{>0}\rightarrow\RR_{>0}$ & Differentiable, increasing escort function. \\
$\kappa, m, q$ & Kaniadakis, Weibull, and Tsallis deformation parameter \\
$\circ$ & Function composition. \\
$D_{\chi}[\mP;\mQ]$ & $\chi$-divergence from $\mP$ to $\mQ$ \\
$\cExp{\cdot}, \cLog{\cdot}$ & $\chi$-deformed exponential and logarithm function, respectively \\
$[x]_+$ & Cutoff function, i.e., $[x]_+=\max(x,0)$ \\
$\dtimes_\chi$ & deformed product\\
$\family_\chi$ & $\chi$-deformed exponential family \\
$\bmtheta, \bmeta$ & Natural and expectation parameters of the $\chi$-deformed exponential family, respectively\\
$\bmtheta_1, \bmeta_1$ & One-body natural and expectation parameters \\
$\bmtheta_2, \bmeta_2$ & Two-body natural and expectation parameters \\
$\psi_\chi(\bmtheta), \phi_\chi(\bmeta)$ & Deformed free energy and its Legendre dual\\
$\varphi_\chi(\mP) \coloneqq \phi_\chi \circ  H(\mP)$ & Deformed entropy\\
$\theta_{11}=-\psi_\chi(\bmtheta)$ & Normalization parameter \\
$\rho_\chi$ & Local decay rate of the transport-plan\\
$S^\chi$ & Response weight\\
$w$ & Normalized row-wise transport weight, i.e.,$w_{ij}=\mR_{ij}/a_i$\\
\bottomrule
\end{tabular}
\end{table}
\section{Remarks}\label{app:remarks}

\begin{figure*}[t]
  \centering
  \includegraphics[width=0.99\linewidth]{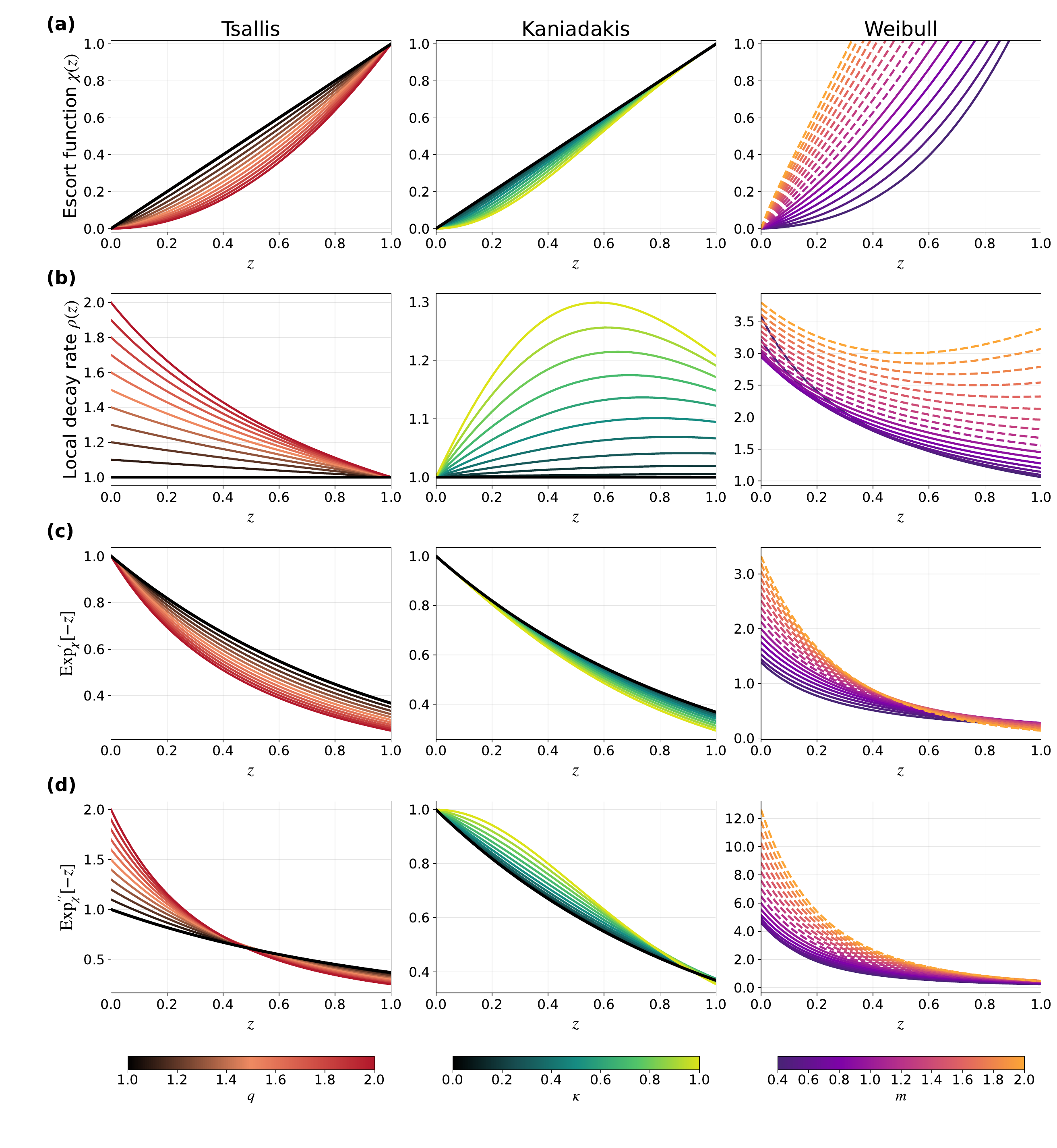}
\caption{\textbf{(a)} Escort function $\chi(z)$, \textbf{(b)} local decay rate $\rho_\chi(z)$, \textbf{(c)} the quantity governing the decay shape, and \textbf{(d)} the quantity governing the response weights $S_\chi(z)$, for the Tsallis, Kaniadakis, and proposed Weibull escorts. Black lines correspond to $q=1$ (Tsallis) and $\kappa=0$ (Kaniadakis), which recover ordinary entropic optimal transport. In the Weibull column, solid lines denote $m \le 1$ (slow decay) and dashed lines $m > 1$ (rapid decay). Tsallis with $q < 1$ is omitted because it causes a domain issue, as discussed in Appendix~\ref{sec:domain_issue}.}
\label{fig:full_plots}
\end{figure*}

\subsection{Domain issue in Tsallis-ET}\label{sec:domain_issue}
Since the domain of the deformed logarithm function $\cLog{\, \cdot \,}$ is $(0,\infty)$, %and the positive Kernel matrix $\mK$ or positive transport plan $\mP$ are positive matrices, 
our formulation does not violate the domain of the deformed logarithm function. However, we need to be aware of the domain of the deformed exponential function to guarantee the existence of a normalizer $g$. 

We assume the domain of $\cExp{\, \cdot \,}$ as $(c_{\rm min},c_{\rm max})$ for $c_{\rm min}\in\mathbb{R}\cup\{-\infty\}$ and $c_{\rm max}\in\mathbb{R}\cup\{\infty\}$. We defined the function $F$ as 
\begin{align*}
    F(g)=\sum_{ij} \cExp{-E_{ij}+g},
\end{align*}
where the appropriate $e_\chi$-normalizer $g^*$ satisfies $F(g^*)=1$. The domain conditions $-E_{ij}+g<c_{\rm max}$ and $c_{\rm min}<-E_{ij}+g$ lead the range of the possible value of $g\in(g_{\rm min},g_{\rm max})$ as
\begin{align*}
    g_{\rm min} = c_{\rm min} + \max_{ij} E_{ij}, \\
    g_{\rm max} = c_{\rm max} + \min_{ij} E_{ij}.
    %c_{\rm min} + \max_{ij} E_{ij} \eqqcolon g_{\rm min} < g < g_{\rm max} \coloneqq c_{\rm max} + \min_{ij} E_{ij}.
\end{align*}
If and only if condition of existence of $g^*$ in the interval $(g_{\rm min},g_{\rm max})$ is given as
\begin{align*}
    F(g_{\rm min}) < 1 < F(g_{\rm max}),
\end{align*}
for
\begin{align*}
    F(g_{\rm min}) = \lim_{g\rightarrow g^{+}_{\rm min}} F(g), \quad
    F(g_{\rm max}) = \lim_{g\rightarrow g^{-}_{\rm max}} F(g).
\end{align*}

Due to the positivity of $\cExp{\ \cdot \ }$, we obtain
\begin{align*}
    F(g_{\rm max}) = \sum_{ij} \cExp{-E_{ij} + c_{\rm max} +  \min_{ij} E_{ij}} 
    \geq \cExp{c_{\rm max}} = \infty,
\end{align*}
where $\cExp{c_{\rm max}}=\infty$ can be shown by the monotonicity of $\cExp{\, \cdot \,}$. Please note that if $\cExp{c_{\rm max}}$ is bounded above, it is inconsistent with the domain of its inverse function $\mathrm{dom}\left( \Log\right)=(0,\infty)$. Thus,  the fact $F(g_{\rm max})>1$ is guaranteed regardless of whether the domain of $\cExp{\ \cdot \ }$ is bounded above. On the other hand, we also obtain the bound as
\begin{align*}
    F(g_{\rm min}) = \sum_{ij} \cExp{-E_{ij} + c_{\rm min} +  \max_{ij} E_{ij}} \in (0,\infty).
\end{align*}
If the domain of $\cExp{\,\cdot \,}$ is not bounded below ($c_{\rm min}=-\infty$), then we can guarantee $F(g_{\rm min})\rightarrow 0$, which together with $F(g_{\rm max})>1$ ensures the existence of $g^*$ via the intermediate value theorem. On the other hand, for a finite lower-bound $c_{\min}<-\infty$, we cannot ensure the $F(g_{\rm min}) = 0$, thus $g^*$ may not exist in the possible domain.

\begin{figure*}[t]
  \centering
  \includegraphics[width=0.9\linewidth]{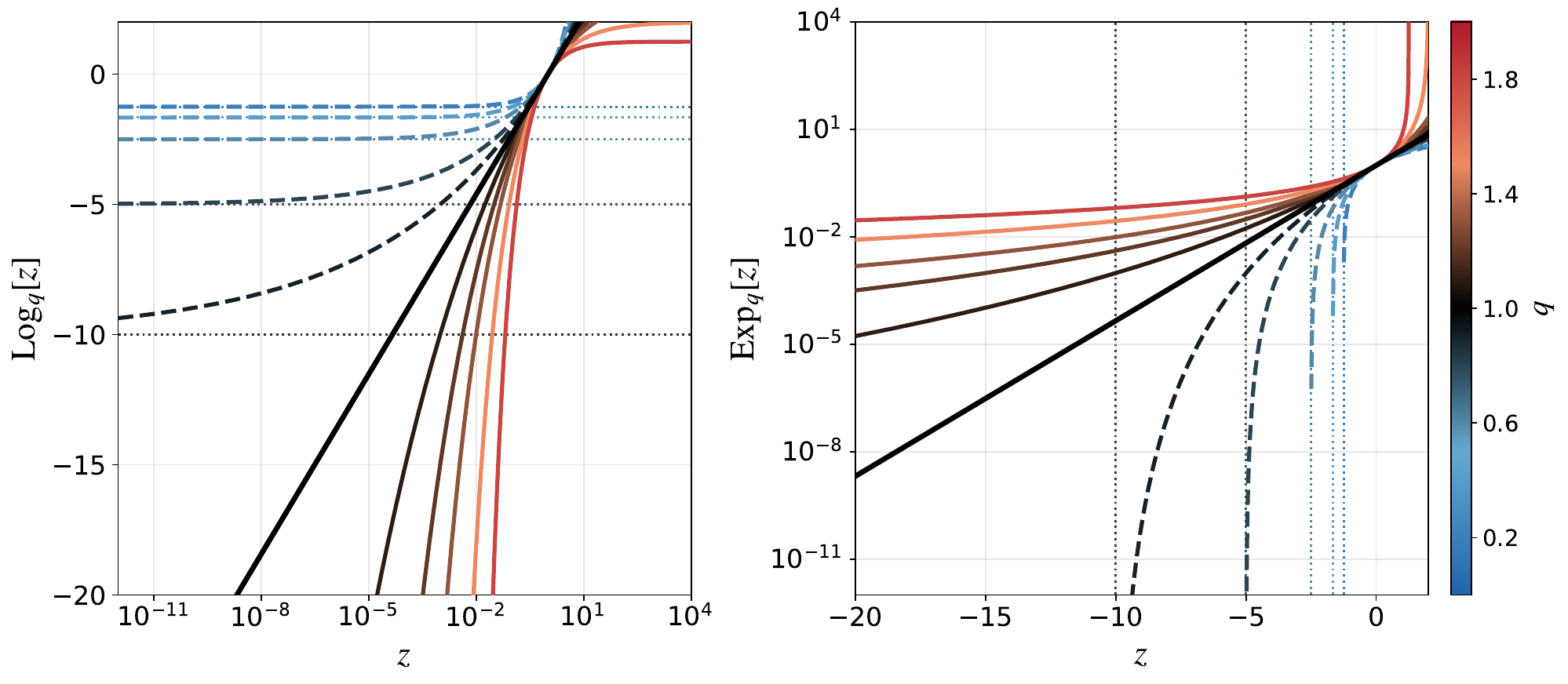}
\caption{Tsallis deformed logarithm and exponential function defined in Equation~\eqref{eq:q_log_q_exp} for varying $q$. The black solid line corresponds to $q=1$, which recovers the ordinary logarithm and exponential function, respectively. Dashed lines denote $q<1$ and solid lines $q\geq1$. For $q<1$, the Tsallis deformed logarithm function is bounded below by $-1/(1-q)$ as $q\rightarrow0^+$, shown as dotted horizontal lines, which causes a domain issue discussed in Appendix~\ref{sec:domain_issue}.}
\label{fig:Tsallis_log}
\end{figure*}

For example, the domain of Tsallis-exponential $\qExp{\, \cdot \,}$ is bounded below by $\qLog{0^+}\rightarrow\frac{-1}{1-q}$ for $q<1$ as seen in Figure~\ref{fig:Tsallis_log}, there is no guarantee the $e_q$-normalizer $g^*$ in Equation~\eqref{eq:chi_div} exists, which restricts Tsallis-ET for slow decay regime ($q\geq1$).

%\begin{remark}
%    In the reformulation in Equation~\eqref{eq:reform}, we assume $\cLog{\mK_{ij}}$ can be appropriately defined. For Tsallis deformation $\chi_{q}(t)=t^q$ with $q<1$, this assumption can not be guaranteed. In particular, 
%    \begin{align*}
%        \cExp{\mK_{ij}}=\left[1+(1-q)\left(-\frac{\mM_{ij}}{\reg}\right)\right]_+^{\frac{1}{1-q}}
%    \end{align*}
%    becomes $0$ if $\reg<(1-q)\mM_{ij}$, where its inverse function $\cLog{\mK_{ij}}$ can not be identified uniquely. Thus, our paper focuses on only $q\geq1$. We note that the inequality $\reg<(1-q)\mM_{ij}$ is not satisfied when we assume $q\geq1$ and $\reg>0$, which make us identify $\cLog{\mK_{ij}}$ uniquely. This difficulty essentially comes from the fact that the Tsallis-deformed logarithm function is bounded below (See Remark~\ref{rmk:bounded_among}). In contrast, $\cLog{\mK_{ij}}$ can be defined for any $\kappa$ for Kaniadakis deformation as emphasized in~\citep{Matsuzoe2015Deformed}.
%\end{remark}

%\begin{remark}\label{rmk:bounded_among}
%    There is no problem even if the deformed logarithm function is bounded above.
%\end{remark}

\subsection{Deformed products}\label{sec:deformed_product}

The $\chi$-deformed product $\dtimes_\chi$ for the Tsallis escort $\chi_q$ and the Weibull escort $\chi_m$ is given as 
\begin{align*}
    a \dtimes_q b = (a^{1-q}+b^{1-q}-1)^{1/(1-q)}, 
\end{align*}
and
\begin{align*}
    a \dtimes_m b = \frac{1}{\exp{\left[ \left\{ \sqrt[m]{\log\frac{1+a}{a}} + \sqrt[m]{\log\frac{1+b}{b}} -\sqrt[m]{\log 2} \right\}^m \right]}-1},
\end{align*}
respectively. As particular cases, the Tsallis product $\dtimes_q$ recovers additive with a constant shift $a \dtimes_q b \rightarrow a+b-1$ for $q\rightarrow0$, standard product $a \dtimes_q b \rightarrow ab$ for $q\rightarrow1$, and $\min$ operator $a \dtimes_q b \rightarrow \min(a,b)$ for $q\rightarrow\infty$. While the Weibull product $\dtimes_m$ also recovers $a \dtimes_m b \rightarrow \min(a,b)$ for $m\rightarrow0$, it does not cover the standard product. As a remark, the Weibull product with $m=1$ can be given as
\begin{align*}
    a \dtimes_m b = \frac{2ab}{(1+a)(1+b)-2ab}, \quad(m=1).
\end{align*}

\section{\texorpdfstring{$\chi$}{Chi}-ET and \texorpdfstring{$\chi$}{Chi}-divergence optimization}\label{sec:proofs}

To discuss the relaxation of the $\chi$-ET in Equation~\eqref{eq:ET} into the $\chi$-divergence optimization in Equation~\eqref{eq:chi_div}, we here review $e_\chi$-normalization~\citep{Matsuzoe2015Deformed}.

\begin{definition}[$e_\chi$-normalization]
    We assume each element of a non-negative normalized matrix $\hat{\mK}$ can be written as $\hat{\mK}_{ij}=\cExp{\mW_{ij}}$ using a real valued matrix $\mW\in\RR^{I\times J}$. We define the $e_\chi$-normalization of the matrix $\hat{\mK}$ by ${\mK}_{ij}=\cExp{\mW_{ij}+g}$ where the constant $g$ is identified to satisfy the normalization $\sum_{ij}{\mK}_{ij}=1$.
\end{definition}

The ordinary normalization that divides each element by the total sum of the matrix, i.e.,
\begin{align*}
    \mK_{ij} = \frac{\hat\mK_{ij}}{\sum_{ij}\hat\mK_{ij}},
\end{align*}
is referred to as $m_\chi$-normalization~\citep{Fujimoto2008generalized, Sakamoto2015Generalization}. When the $\chi$-escort function is the identity function $\chi(t)=t$, the $m_\chi$-normalization and $e_\chi$-normalization are equivalent due to the exponential laws. The escort distribution is defined with $m_\chi$-normalization of $\chi[\mP]$, i.e.,
\begin{align*}
    \cEsc{\mP}_{ij} \coloneqq \frac{\Esc{\mP_{ij}}}{\sum_{ij}\Esc{\mP_{ij}}}.
\end{align*}

The following theorem shows that the $e_\chi$-normalization for the matrix $\mK$ does not change the optimal solution $\mP$ of minimizing the $\chi$-divergence $D_\chi[\mP;\mK]$.
\begin{theorem}\label{th:solution_invariance}
 For given non-negative matrix $\hat{\mK}$ and its $e_\chi$-normalization $\mK$, it holds that
 \begin{align*}
      \argmin_{\mP\in\subfamily} D_\chi[\mP; \hat{\mK}]=
      \argmin_{\mP\in\subfamily} D_\chi[\mP; \mK],
 \end{align*}
where $\subfamily$ is an appropriately defined feasible manifold.
\begin{proof}
    We represent the given matrix $\hat{\mK}$ by real valued matrix $\mW$ as 
    \begin{align*}
        \hat{\mK}_{ij} = \cExp{\mW_{ij}},
    \end{align*}
    and its $e_\chi$-normalization $\mK$ can be written as 
    \begin{align*}
        \mK_{ij} = \cExp{\mW_{ij}+g},
    \end{align*}
    where the constant $g\in\RR$ is defined so as to satisfy $\sum_{ij}\mK_{ij}=1$.
    \begin{align*}
        &\argmin_{\mP\in\subfamily} D_\chi[\mP; \mK] \\
        =&\argmin_{\mP\in\subfamily} \sum_{ij}\cEsc{\mP_{ij}}\left(\cLog{\mP_{ij}}-\cLog{\mK_{ij}}\right) \\
        =&\argmin_{\mP\in\subfamily} \sum_{ij}\cEsc{\mP_{ij}}\left(\cLog{\mP_{ij}}-\cLog{ \cExp{\mW_{ij}+g}}\right) \\
        =&\argmin_{\mP\in\subfamily} \sum_{ij}\cEsc{\mP_{ij}}\left(\cLog{\mP_{ij}}-\mW_{ij}-g\right) \\
        =&\argmin_{\mP\in\subfamily} \sum_{ij}\cEsc{\mP_{ij}}\left(\cLog{\mP_{ij}}-\mW_{ij}\right) - g \sum_{ij} \cEsc{\mP}_{ij}\\
        =&\argmin_{\mP\in\subfamily} \sum_{ij}\cEsc{\mP_{ij}}\left(\cLog{\mP_{ij}}-\mW_{ij}\right)\\
        =&\argmin_{\mP\in\subfamily} \sum_{ij}\cEsc{\mP_{ij}}\left(\cLog{\mP_{ij}}-\cLog{\hat{\mK}_{ij}}\right)\\
        =&\argmin_{\mP\in\subfamily} D_\chi[\mP; \hat{\mK}],
    \end{align*}
    where we used the normalization $\sum_{ij} \cEsc{\mP}_{ij}=1$.
\end{proof}
\end{theorem}
%We emphasize that there is no guarantee that the statement in Theorem~\ref{th:solution_invariance} also holds for $m_\chi$-normalization. 
The above result leads to the following reformulation of the $\chi$-ET as discussed in Section~\ref{sec:ET}.% and motivates introducing $e$-normalization in $\chi$-ET. 

\begin{proposition}\label{prop:ET_and_divergence}
For given cost matrix $\cost$ and strength of the regularization $\reg$, the $\chi$-escort entropic transport in Equation~\eqref{eq:ET} is equivalent to minimizing the $\chi$-divergence $D_{\chi}[\mP; \mK]$ for $\mP\in\csimplex{\ba}{\bb}$, i.e.,
\begin{align*}
    \min_{\mP\in\csimplex{\ba}{\bb}} \inner{\cost}{\cEsc{\mP}} + \varepsilon \varphi_\chi(\mP) 
    = \min_{\mP\in\csimplex{\ba}{\bb}}  \varepsilon D_{\chi}[\mP; \mK],
\end{align*}
where the $\chi$-deformed kernel matrix $\mK$ is an $e_\chi$-normalization of the matrix 
\begin{align*}
    \hat{\mK}_{ij} = \cExp{-\frac{\cost_{ij}}{\reg}}.
\end{align*}
\end{proposition}
\begin{proof}
\begin{align*}
    & \min_{\mP\in\csimplex{\ba}{\bb}} \inner{\cost}{\cEsc{\mP}} + \varepsilon \varphi_\chi(\mP) \\
    &= \min_{\mP\in\csimplex{\ba}{\bb}}  \sum_{ij} \cost_{ij} \cEsc{\mP_{ij}} + \varepsilon\sum_{ij} \cEsc{\mP_{ij}} \cLog{\mP_{ij}} \\
    &= \min_{\mP\in\csimplex{\ba}{\bb}}  \varepsilon \sum_{ij} \cEsc{\mP_{ij}} \left(\cLog{\mP_{ij}}-\cLog{\hat{\mK}_{ij}}\right) \\
    &= \min_{\mP\in\csimplex{\ba}{\bb}}  \varepsilon D_{\chi}[\mP; \hat{\mK}] \\
    &= \min_{\mP\in\csimplex{\ba}{\bb}}  \varepsilon D_{\chi}[\mP; \mK]
\end{align*}
where the last equality follows from Theorem~\ref{th:solution_invariance} and we use $\cost_{ij} = -\varepsilon \cLog{\hat{\mK}_{ij}}$ and the definition of the $\chi$-entropy $\varphi_\chi(\mP)=\sum_{ij} \cEsc{\mP}_{ij} \cLog{\mP_{ij}}$.
\end{proof}

\begin{proposition}\label{prop:limit}
In the cost free limit $\reg\rightarrow\infty$, the optimal plan in Equation~\eqref{eq:ET} can be given as a form of 
\begin{align*}
    \cEsc{\mP}_{ij}=\cEsc{X_i \dtimes_\chi Y_j}.
\end{align*}
\end{proposition}
\begin{proof}
    In the cost free limit $\reg\rightarrow\infty$, the kernel matrix $\mK$ in Equation~\eqref{eq:chi_div} can be given as 
    \begin{align*}
        \mK_{ij} = \cExp{-\frac{\mM_{ij}}{\reg}+g} \longrightarrow \cExp{g}.
    \end{align*}
    In this limit, the normalizing condition $\sum_{ij}\mK_{ij}=1$ leads to
    \begin{align*}
        \sum_{ij} \mK_{ij} 
        = \sum_{ij} \cExp{g}
        = IJ \cExp{g} = 1.
    \end{align*}
    Thus, the $e_\chi$-normalizer $g$ is given as $g = \cLog{1/IJ}$ and the kernel matrix is
    \begin{align*}
        \mK_{ij} = \cExp{g} = \cExp{ \cLog{\frac{1}{IJ}}} = \frac{1}{IJ}.
    \end{align*}
    Its two-body parameters $\bmtheta^{\mK}_2$ can be obtained by Equation~\eqref{eq:update_thetaK} as $\bmtheta^{\mK}_2 = \mathbf{0}$. As discussed in Section~\ref{sec:ET}, the two-body parameters are invariant in the projection onto the $m_\chi$-flat manifold $\csimplex{\ba}{\bb}$, i.e., $\bmtheta^{\mP}_2 = \mathbf{0}$. Therefore, we obtain
    \begin{align*}
    \mP_{ij}&=\cExp{\sum_{i'=1}^i\sum_{j'=1}^j\theta^{\mP}_{i'j'}} \\
        &=\cExp{\theta^{\mP}_{11}+\sum_{i'=2}^i\theta^{\mP}_{i'1}+\sum_{j'=2}^j\theta^{\mP}_{1j'}} \\
        &=\cExp{x_i+y_j} =\cExp{x_i} \dtimes_\chi \cExp{y_j} = X_i \dtimes_\chi Y_j,
    \end{align*}
    where $\chi$-deformed product $\dtimes_\chi$ is defined through the generalized exponential law $\cExp{x_i}\dtimes_\chi\cExp{y_j}=\cExp{x_i+y_j}$ and 
    we define 
    $x_i=\theta^{\mP}_{11}/2+\sum_{i'=2}^i\theta^{\mP}_{i'1}$,
    $y_j=\theta^{\mP}_{11}/2+\sum_{j'=2}^j\theta^{\mP}_{1j'}$,
    $X_i=\cExp{x_i}$, and $Y_j=\cExp{y_j}$. Thus, we obtain $\cEsc{\mP_{ij}}=\cEsc{X_i \dtimes_\chi Y_j}$.
\end{proof}

\begin{proposition}\label{prop:decay}
The optimal $\chi$-escort plan of $\chi$-ET takes the form:
\begin{align*}
    \cEsc{\mP}_{ij} \propto \cExpp{-\frac{\mM_{ij}}{\reg} - \alpha_i - \beta_j}
\end{align*}
where $\cExpp{t}$ is the derivative of the $\chi$-exponential function $\cExp{t}$ and scalars $(\alpha_i)_i, (\beta_j)_j$ are determined by the $\chi$-balancing condition and the normalization $\sum_{ij}\mP_{ij}=1$.
\end{proposition}
\begin{proof}
    Let $\mP^*$ be the optimal plan that minimizes Equation~\eqref{eq:ET}. As discussed in Section~\ref{sec:ET}, the two-body natural parameters do not change in this optimization, i.e., $\bmtheta_2^{\mP^*}=\bmtheta_2^{\mK}$. Therefore, we obtain
    \begin{align*}
        \cLog{\mP_{ij}^*} - \cLog{\mK_{ij}} 
        &= \sum_{i'=1}^i\sum_{j'=1}^j
        \theta_{i'j'}^{\mP^*} - \theta_{i'j'}^{\mK} \\
        &= \theta^{\mP^*}_{11} - \theta^{\mK}_{11} 
        + \sum_{i'=2}^i \left( \theta^{\mP^*}_{i'1} - \theta^{\mK}_{i'1} \right)
        + \sum_{j'=2}^j \left( \theta^{\mP^*}_{1j'} - \theta^{\mK}_{1j'} \right).
    \end{align*}
    Thanks to Equation~\eqref{eq:chi_div}, it holds that
    \begin{align*}
     \cLog{\mK_{ij}} = -\frac{\mM_{ij}}{\reg} + g.
    \end{align*}
    We define each element of the vectors $\boldsymbol{\alpha}=(\alpha_i)_i$ and $\boldsymbol{\beta}=(\beta_j)_j$ as follows:
    \begin{align*}
        -\alpha_i=\frac{\theta^{\mP^*}_{11} - \theta^{\mK}_{11} + g}{2}+ \sum_{i'=2}^i\theta^{\mP^*}_{i'1} - \theta^{\mK}_{i'1}, \  \quad \ 
        -\beta_j=\frac{\theta^{\mP^*}_{11} - \theta^{\mK}_{11}+g}{2}+ \sum_{j'=2}^j  \theta^{\mP^*}_{1j'} - \theta^{\mK}_{1j'},
    \end{align*}
    then, 
    \begin{align*}
        \cLog{\mP_{ij}^*} = -\frac{\mM_{ij}}{\reg}  -\alpha_i - \beta_j.
    \end{align*}
    Thus, we obtain $\mP_{ij}^*=\cExp{-z_{ij}}$ for $ z_{ij} = \mM_{ij}/{\reg} + \alpha_i + \beta_j$ and its escort can be given as
    \begin{align*}
     \cEsc{\mP}_{ij} = \frac{ \cExpp{-z_{ij}} }{\sum_{kl} \cExpp{-z_{kl}}},
    \end{align*}
    where we used the relation 
    \begin{align}\label{eq:deff_chi_relation}
        \cExpp{ t } = \Esc{\cExp{t}},
    \end{align}
    which can be obtained by differentiating $\cLog{\cExp{t}}=t$ with respect to $t$.
\end{proof}

\begin{proposition}\label{prop:barrier}
    In the optimization of $\chi$-ET in Equation~\eqref{eq:ET}, if the escort function $\chi$ is defined such that $\cLog{0^+}=-\infty$, the corresponding regularizer $\Omega(\mR)=\varphi_\chi(\invEsc{\mR})$ generates the barrier at the boundary of the support, i.e.,
    \begin{align*}
           \left. \frac{\partial \Omega(\mR)}{\partial \mR_{ij}}\right|_{\mR_{ij}=0}\rightarrow\infty,
    \end{align*}
    and guarantees the full support of the optimal plan $\mR_{ij}>0$ for all $(i,j)\in\mathcal{X}=[I]\times[J]$.
\end{proposition}
\begin{proof}
    We define $\mR_{ij}=\cEsc{\mP}_{ij}$ and $\mP_{ij}=\invEsc{\mR}_{ij}$. The derivative of the regularizer is given as  
    \begin{align*}
        \frac{\partial \varphi_\chi(\mP)}{\partial\mR_{st}} 
        &= \frac{\partial }{\partial\mR_{st}}  \sum_{ij} \cEsc{\mP_{ij}}\cLog{\mP_{ij}} \\
        &= \frac{\partial }{\partial\mR_{st}}  \sum_{ij} \mR_{ij}\cLog{\mP_{ij}} \\
        &=\cLog{\mP_{st}} + \sum_{ij} \mR_{ij}  \frac{\partial \cLog{\mP_{ij}}}{\partial\mR_{st}} \\
        &=\cLog{\mP_{st}},
    \end{align*}
    where the normalizing condition $\sum_{ij}\mP_{ij}=1$ makes the second term vanished as
    \begin{align*}
        \sum_{ij} \mR_{ij}  \frac{\partial \cLog{\mP_{ij}}}{\partial\mR_{st}}
        &= \sum_{ij} \mR_{ij} \frac{\partial \mP_{ij}}{\partial \mR_{st}} \frac{\partial }{\partial \mP_{ij}} \cLog{\mP_{ij}} \\
        &= \sum_{ij} \mR_{ij} \frac{\partial \mP_{ij}}{\partial \mR_{st}} \frac{1}{\Esc{\mP_{ij}}} \\
        &= \sum_{ij} \frac{\Esc{\mP_{ij}}}{\sum_{nm}\Esc{\mP_{nm}}} \frac{\partial \mP_{ij}}{\partial \mR_{st}}  \frac{1}{\Esc{\mP_{ij}}} \\
        &= \frac{1}{\sum_{nm}\Esc{\mP_{nm}}} \frac{\partial}{\partial \mR_{st}} \sum_{ij} \mP_{ij} \\
        &= \frac{1}{\sum_{nm}\Esc{\mP_{nm}}} \frac{\partial}{\partial \mR_{st}} 1 \\
        &= 0.
    \end{align*}
    
    Because of the monotonicity of $\chi$-function and definition of the escort, i.e., $\mR_{ij}=\Esc{\mP_{ij}}/\sum_{kl}\Esc{\mP_{kl}}$, the limit  
    $\mR_{ij}\rightarrow0^+$ is equivalent with 
    $\Esc{\mP_{ij}}\rightarrow0^+$. Lemma~\ref{lemma:escort0} ensures this limit is equivalent with $\mP_{ij}\rightarrow0^+$. Thus, the assumption $\cLog{0^+}=-\infty$ leads to
    \begin{align*}
    \left. \frac{\partial \Omega(\mR)}{\partial \mR_{ij}}\right|_{\mR_{ij}=0}= 
        \left. 
        \cLog{\mP_{st}} \right|_{\mR_{ij}\rightarrow0^+}
        =
\left. 
        \cLog{\mP_{st}} \right|_{\mP_{ij}\rightarrow0^+}
        =-\infty.
    \end{align*}
    Therefore, $\cLog{0^+}=-\infty$ generates the barrier at the boundary of support and ensures the full support of the optimal plan $\mR_{ij}>0$ for all $(i,j)\in\mathcal{X}=[I]\times[J]$.
\end{proof}

\begin{lemma}\label{lemma:escort0}
    $\cLog{0^+}=-\infty$ ensures $\chi(0^+)=0$.
\end{lemma}
\begin{proof}
    By definition in Equation~\eqref{eq:clog_cExp}, the assumption $\cLog{0^+}=-\infty$ is equivalent to
    \begin{align*}
        \int_{0}^1\frac{ds}{\chi(s)} = \infty.
    \end{align*}
    Suppose, for contradiction, that $\chi(0^+)>0$. Then there exist $c>0$ and a sufficiently small $\delta>0$ such that $\chi(s) \geq c$ for all $s\in(0,\delta)$. Hence,
    \begin{align*}
        \int_{0}^{\delta} \frac{ds}{\chi(s)} \leq 
        \int_{0}^{\delta} \frac{ds}{c} =\frac{\delta}{c} < \infty.
    \end{align*}
    Moreover, since $\chi$ is positive and increasing on $(0,\infty)$, we also have
    \begin{align*}
        \int_{\delta}^{1} \frac{ds}{\chi(s)} \leq 
        \int_{\delta}^{1} \frac{ds}{\chi(\delta)} =
        \frac{1-\delta}{\chi(\delta)} < \infty.
    \end{align*}
    Therefore, we obtain
    \begin{align*}
        \int_{0}^{1} \frac{ds}{\chi(s)} = 
        \int_{0}^{\delta} \frac{ds}{\chi(s)}
        +
        \int_{\delta}^{1} \frac{ds}{\chi(s)} < \infty,
    \end{align*}
    which contradicts with $\cLog{0^+} =-\infty$. Thus, we conclude that $\chi(0^+)=0$.
\end{proof}

\begin{proposition}\label{prop:Kani_tail}
    %Kaniadakis-ET does not provide a light-tailed regime. 
    For any hyperparameter $\kappa\in(0,1)$, its optimal transport plan of Kaniadakis-ET exhibits a slow decay.
\end{proposition}
\begin{proof}
    Proposition~\ref{prop:decay} shows the derivative of the Kaniadakis exponential function $\kExp{-z}=(-\kappa z + \sqrt{1+\kappa^2z^2})^{1/\kappa}$ describes the decay of the optimal plan $\mR$. The derivative of the Kaniadakis exponential function is given as follows~\citep{Pistone2023Kani}:
    \begin{align*}
        \kExpp{-z} = \frac{1}{\sqrt{1+\kappa^2z^2}}\kExp{-z}.
    \end{align*}
    Assuming $z$ is large, we obtain the dominant term of the tail shape as
    \begin{align}\label{eq:Kani_decay}
        \kExpp{-z} \simeq C_\kappa z^{-\frac{1+\kappa}{\kappa}},
    \end{align}
    where we used the binomial expansion
    \begin{align*}
        \sqrt{1+\kappa^2z^2} \simeq \kappa z + \frac{1}{2\kappa z}, \quad \frac{1}{\sqrt{1+\kappa^2z^2}} \simeq \frac{1}{\kappa z} \left(  1 - \frac{1}{2\kappa^2 z^2} \right),
    \end{align*}
    and the constant $C_\kappa$ is given as $C_\kappa=2^{-1/\kappa}\kappa^{-\frac{1+\kappa}{\kappa}}$. %Since the absolute value of the power in Equation~\eqref{eq:Kani_decay} is always larger than $1$, i.e., $(1+\kappa)/\kappa>1$, 
    This asymptotic form is polynomial rather than exponential or super-exponential. Therefore, the optimal plan by Kaniadakis-ET exhibits slow decay.
\end{proof}

When considering the effect of a small change $z \rightarrow z+\Delta z$ on a positive function $R(z)$, a first-order Taylor expansion of  $R$ with respect to $z$ gives 
\begin{align*}
    \frac{R(z+\Delta z)-R(z)}{R(z)} = \frac{R'(z)}{R(z)}\Delta z + O\left((\Delta z)^2\right).
\end{align*}
We therefore focus on $-{R'(z)}/{R(z)}$ as the local decay rate, which characterizes the decay property of the function $R$. The following proposition provides the closed-form expression for the local decay ratio.

\begin{proposition}\label{prop:decay_rate}
    Assuming the plan $\mR(z_{ij})$ as a function of $z_{ij}=\cost_{ij}/\reg+\alpha_i+\beta_j$ with Lagrange multipliers $\alpha_i$ and $\beta_j$, its local decay rate $\rho(z_{ij}) \coloneq -R'(z_{ij})/R(z_{ij})$ is given as
    \begin{align}\label{eq:def_of_sensti}
        \rho_\chi(z_{ij}) = \chifp{\cExp{-z_{ij}}}.
    \end{align}
    \begin{proof}
        %The local decay of $\mR$ is defined as relative shift of the limit $\Delta z\rightarrow 0$ of the perturbation $z \rightarrow z + \Delta z$ as
        %\begin{align}\label{eq:decayratio}
        %    \frac{R(z_{ij}+\Delta z_{ij})}{R(z_{ij})} = \frac{\tilde{R}(z_{ij}+\Delta z_{ij})}{\tilde{R}(z_{ij})} \xlongrightarrow{\Delta z_{ij} \to 0}\frac{1}{\tilde{R}(z_{ij})} \frac{d \tilde{R}(z_{ij})}{dz_{ij}}=\frac{d}{dz_{ij}} \log{\tilde{R}(z_{ij})}\eqcolon  -\rho_\chi(z_{ij})
        %\end{align}
        %where $\tilde{R}$ is non-normalized plan, i.e., 
        %\begin{align*}
        %    \mR_{ij}=\frac{\tilde{\mR}_{ij}}{\sum_{ij}\tilde{\mR}_{ij}} .
        %\end{align*}
        %According to Proposition~\ref{prop:decay}, the form of the non-normalized plan $\tilde{R}$ is $\tilde{R}(z)=\cExpp{-z}$.%, the right-hand side of Equation~\eqref{eq:decayratio} can be obtained as follows:
        \begin{align*}
            \rho_\chi(z_{ij}) &\coloneq -\frac{R^\prime(z_{ij})}{R(z_{ij})} 
            = -\frac{1}{R(z_{ij})}\frac{dR(z_{ij})}{dz_{ij}} 
            = -\frac{d}{dz_{ij}} \log{R(z_{ij})} \\
            %&= -\frac{d}{dz} \left(\log{\cExp{-z}}  - \log{\sum_{ij \cExp{-z_{ij}}} }\right)\\
            &= -\frac{d}{dz_{ij}} \log{\cExpp{-z_{ij}}} \\
            %&= -\frac{d}{dz} \log{\chif{\cExp{-z}}} \\
            &= -\frac{1}{\chif{\cExp{-z_{ij}}}} \frac{d}{dz_{ij}} \chif{\cExp{-z_{ij}}} \\
            &= \frac{1}{\chif{\cExp{-z_{ij}}}} \chif{\cExp{-z_{ij}}} \frac{d\chif{\cExp{-z_{ij}}}}{d \cExp{-z_{ij}}} \\
            &= \chifp{\cExp{-z_{ij}}},
        \end{align*}
        where we use Equation~\eqref{eq:deff_chi_relation} and Proposition~\ref{prop:decay}.
    \end{proof}
\end{proposition}

\begin{theorem}[]\label{th:EGRF}
    The perturbation effect on the barycenter $\Delta\hat{\bx}^{(i)}=\sum_j \Delta w_{ij} \by^{(j)}$ depends on the escort function $\chi$ as follows:
\begin{align}
    \Delta\hat{\bx}^{(i)[\chi]}(\bz_i) =
    -\sum_{j=1}^J w_{ij}(\bz_i) \rho_\chi(z_{ij})(\by^{(j)}-\hat{\bx}^{(i)})\Delta z_{ij} + O(\|\Delta \bz\|^2).
\end{align}
\end{theorem}
\begin{proof}
    We expand $\cExpp{-z_{ij}-\Delta z_{ij}}$ around $\Delta z_{ij}=0$ as
    \begin{align*}
        \cExpp{-z_{ij}-\Delta z_{ij}} &= 
        \cExpp{-z_{ij}} - \cExppp{-z_{ij}}\Delta z_{ij} + O(\|\Delta \bz\|^2) \\
        &=\cExpp{-z_{ij}} \left(
            1- \rho_\chi(z_{ij})\Delta z_{ij}
        \right) +  O(\|\Delta \bz \|^2),
    \end{align*}
    where we use $\cExppp{-z_{ij}}=\rho_\chi(z_{ij})\cExpp{-z_{ij}}$ which can be obtained from Equation~\eqref{eq:def_of_sensti} and one more derivative on both sides of Equation~\eqref{eq:deff_chi_relation}. Thus, the perturbation on the $\chi$-generalized softmax in Equation~\eqref{eq:chi_softmax} is given by
    \begin{align*}
        w_{ij}^{[\chi]}(\bz_i+\Delta \bz_i)=
        \frac
        {
        \cExpp{-z_{ij}} \left(
            1- \rho_\chi(z_{ij})\Delta z_{ij}
        \right) +  O(\|\Delta \bz \|^2)
        }{
        \sum_{k=1}^J
        \cExpp{-z_{ik}} \left(
            1- \rho_\chi(z_{ik})\Delta z_{ik}
        \right) +  O(\|\Delta \bz \|^2)
        }.
    \end{align*}
    The denominator can be transformed as follows:
    \begin{align*}
        &\sum_{k=1}^J
        \cExpp{-z_{ij}} \left(
            1- \rho_\chi(z_{ik})\Delta z_{ik}
        \right) +  O(\|\Delta \bz \|^2) \\
        &=\sum_{k=1}^J
        \cExpp{-z_{ik}}
            - \sum_{k=1}^J \rho_\chi(z_{ik})\Delta z_{ik}
         +  O(\|\Delta \bz \|^2) \\
        &=
        \left(\sum_{k=1}^J
        \cExpp{-z_{ik}}\right)\left[
        1-\frac{\sum_{k=1}^J \cExpp{-z_{ik}\rho_\chi(z_{ik})\Delta z_{ik}}}
        {\sum_{k=1}^J\cExpp{-z_{ik}}}\right]+  O(\|\Delta \bz \|^2) \\
        &=
        \left(\sum_{k=1}^J
        \cExpp{-z_{ik}}\right)\left[
        1-\sum_{k=1}^J w_{ik}(\bz_i)\rho_\chi(z_{ik})\Delta z_{ik}\right]+  O(\|\Delta \bz \|^2).
    \end{align*}
    As a result, we obtain
    \begin{align*}
        w_{ij}(\bz_i+\Delta \bz_i)
        &= w_{ij}(\bz_i)
        \frac{1-\rho_\chi(z_{ij})\Delta z_{ij}}
        {1-\sum_{k} w_{ik}(\bz_i)\rho_\chi(z_{ik})\Delta z_{ik}} + O(\|\Delta \bz\|^2) \\
        &=w_{ij}(\bz_i)\left(
        1-\rho_\chi(z_{ij})\Delta z_{ij}
        \right)
        \left(
        1+\sum_{k=1}^J
        w_{ik}(\bz_i)\rho_\chi(z_{ik})\Delta z_{ik}
        \right) + O(\|\Delta \bz\|^2),
    \end{align*}
    where we have used Equation~\eqref{eq:chi_softmax} and $(1-u)^{-1}=1+u+O(u^2)$ with $u=\sum_k w_{ik}(z_i)\rho_\chi(z_{ik})\Delta z_{ik}$. Therefore,
    \begin{align*}
        \Delta w_{ij} &\coloneqq 
        w_{ij}(\bz_i+\Delta z_{i}) - w_{ij}(\bz_i)\\
        &= -w_{ij}(\bz_i)\rho_\chi(z_{ij})\Delta z_{ij} + w_{ij}(\bz_{i})\sum_{k=1}^J
        w_{ik}(\bz_i)\rho_\chi(z_{ik})\Delta z_{ik} + O(\|\Delta \bz\|^2).
    \end{align*}
    Then, the shift of the centroid can be given as
    \begin{align*}
        \Delta \hat{\bx}^{(i)}
        &= \sum_{j=1}^J \Delta w_{ij} \by^{(j)} \\
        &= -\sum_{j=1}^J w_{ij}(\bz_i)\rho_\chi(z_{ij})\Delta z_{ij}\by^{(j)} + \sum_{j=1}^J w_{ij}(\bz_{i})\by^{(j)}\sum_{k=1}^J
        w_{ik}(\bz_i)\rho_\chi(z_{ik})\Delta z_{ik} + O(\|\Delta \bz\|^2) \\
        &=
        -\sum_{j=1}^J w_{ij}(\bz_i)\rho_\chi(z_{ij})\Delta z_{ij}\by^{(j)} + \hat{\bx}^{(i)} \sum_{j=1}^J
        w_{ij}(\bz_i)\rho_\chi(z_{ij})\Delta z_{ij} + O(\|\Delta \bz\|^2) \\
        &=
            -\sum_{j=1}^J w_{ij}(\bz_i) \rho_\chi(z_{ij})(\by^{(j)}-\hat{\bx}^{(i)})\Delta z_{ij} + O(\|\Delta \bz\|^2),
    \end{align*}
    where we use $\hat{\bx}^{(i)}=\sum_j w_{ij}\by^{(j)}$.
\end{proof}

\section{Proof for update rules}

\begin{proposition}\label{prop:grad1}
$\boldsymbol{\nabla}_{\bmeta^{\mP}_2} D_{\chi}[\mP;\mK] = \bmtheta^{\mP}_2 - \bmtheta^{\mK}_2$.
\end{proposition}
\begin{proof}
The definition of the Bregman divergence generated by the convex function $\phi_{\chi}(\bmeta)$~\citep{Bregman1967relaxation} is given by
\begin{align*}
        D_\chi[\mP; \mK] = \phi_\chi(\bmeta^\mP) - \phi_\chi(\bmeta^\mK) - \boldsymbol{\nabla}_{\bmeta} \phi_\chi(\bmeta^\mK)^\top (\bmeta^\mP - \bmeta^\mK).
\end{align*}
We use the relation $\bmtheta^{\mP}_2=\boldsymbol{\nabla}_{\bmeta_2}\phi_{\chi}(\bmeta^{\mP})$ to obtain
\begin{align*}
        \boldsymbol{\nabla}_{\bmeta^{\mP}_2} D_\chi[\mP; \mK] = \boldsymbol{\nabla}_{\bmeta^{\mP}_2}\phi_\chi(\bmeta^\mP) -  \boldsymbol{\nabla}_{\bmeta^{\mP}_2} \phi_\chi(\bmeta^\mK)  
        =\bmtheta^{\mP}_2 - \bmtheta^{\mK}_2.
\end{align*}
\end{proof}

\begin{proposition}\label{prop:grad2}
    $\boldsymbol{\nabla}_{\bmtheta_1} D_{\chi}[\mQ;\mP]=\bmeta^{\mP}_{1}-\bmeta^{\mQ}_{1}$.
\end{proposition}
\begin{proof}
    %We used the duality of the Bregman divergence, that is, $D_\chi(\mR,\mP)$ is a Bregman divergence generated by a Legendre transformation of the convex function generating $D_\chi(\mP,\mR)$~\citep{Amari2010Information}. 
    
    We use that the duality of Bregman divergences
    $D_\chi[\mQ;\mP]$ is the Bregman divergence generated by the Legendre transformation of the convex function that generates $D_\chi[\mP;\mQ]$~\citep{Amari2010Information}. Supposing the convex function $\psi_\chi(\bmtheta)$ is the Legendre transform of $\phi_\chi(\bmeta)$, it holds that 
    \begin{align*}
        D_\chi[\mQ;\mP] = \psi_\chi(\bmtheta^\mP) - \psi_\chi(\bmtheta^\mQ) - \boldsymbol{\nabla}_{\bmtheta} \psi_\chi(\bmtheta^\mQ)^\top (\bmtheta^\mP - \bmtheta^\mQ).
    \end{align*}
    Using the relation $\bmeta^{\mP}_1=\boldsymbol{\nabla}_{\bmtheta_1}\psi_{\chi}(\bmtheta^{\mP})$, we obtain
    \begin{align}\label{eq:1stderiv}
        \boldsymbol{\nabla}_{\bmtheta^{\mP}_1} D_\chi[\mQ;\mP] = \boldsymbol{\nabla}_{\bmtheta^{\mP}_1}\psi_\chi(\bmeta^\mP) -  \boldsymbol{\nabla}_{\bmtheta^{\mP}_1} \psi_\chi(\bmeta^\mQ)  
        =\bmeta^{\mP}_1 - \bmeta^{\mQ}_1.
    \end{align}
\end{proof}

We assume $\mQ\in\csimplex{\ba}{\bb}$ in Section~\ref{sec:optim}, thus its one-body natural-parameters $\bmeta^\mQ_1$ satisfies the $\chi$-balancing condition in Equation~\eqref{eq:balancing_condition_with_eta}. As a results, it holds that $\boldsymbol{\nabla}_{\bmtheta_1} D_{\chi}[\mQ;\mP]=\bmeta^{\mP}_{1}-\bmeta^{\mQ}_{1} = \bmeta^{\mP}_1 - \hat{\bmeta}_1$.

While the following statement can be regarded as a particular case of Theorem 1 in~\citep{Ghalamkari2026Deformed}, we include its derivation here for completeness of this paper.
\begin{figure*}[t]
  \centering
  \includegraphics[width=0.99\linewidth]{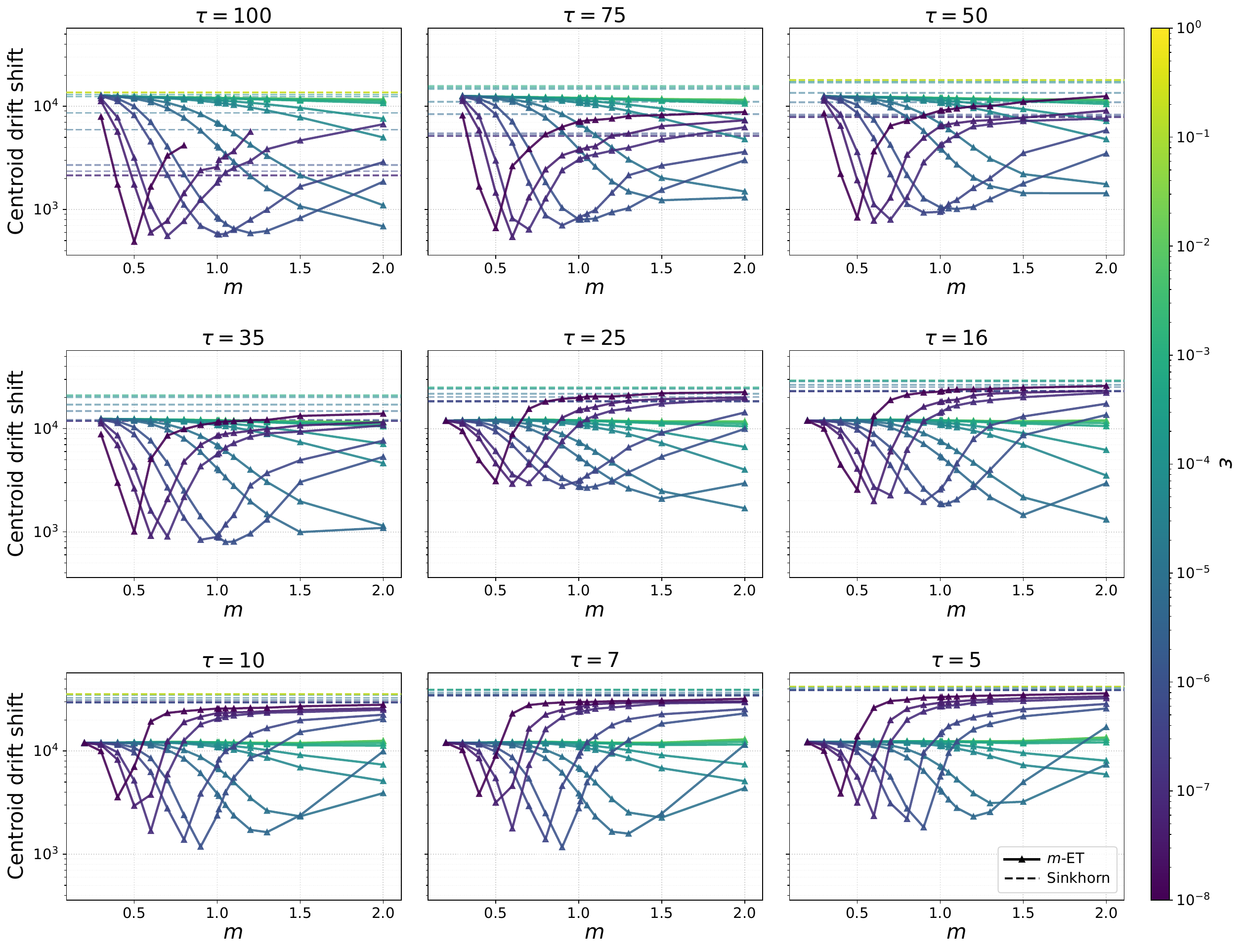}
  \caption{The full version of Figure~\ref{fig:main_exp}(\textbf{c},\textbf{d}) with various noise level $\tau$. While larger noise (smaller $\tau$) makes the centroid drift shift larger, the proposed Weibull-ET with appropriately chosen $\reg$ and $m$ consistently achieves less centroid drift shift than that of Sinkhorn.}
  \label{fig:all_mET}
\end{figure*}
\begin{proposition}\label{prop:FIM}
    The second-order derivative of the function $D_\chi[\mQ;\mP]$ with respect to $\bmtheta^{\mP}$ can be given as 
    \begin{align*}
       \FIM_{(ij),(vw)} = \frac{\partial \eta_{vw}}{\partial \theta_{ij}} = \sum_{k=1}^I \sum_{l=1}^J \cEsc{\mP}_{kl}\chifp{\mP_{kl}}
    \Delta_{ij}^{kl}\Delta_{vw}^{kl},
        \end{align*}
where the term $\Delta_{ij}^{kl}$ is defined as $\Delta_{ij}^{kl}=\zeta_{ij}^{kl}-\eta_{ij}$ and the zeta-function $\zeta:\mathcal{X}\times\mathcal{X}\rightarrow\{0,1\}$ is defined as 
$\zeta_{ij}^{kl}=1$ if $i \leq k, j \leq l$ and $0$ otherwise.
\end{proposition}
\begin{proof}
The first-order derivative of each element in $\mP$ with respect to the natural parameter 
$\bmtheta_{ij}$ is obtained as follows:
\begin{align*}
        \frac{\partial \mP_{kl}}{\partial \theta_{ij}}
        &= \frac{\partial }{\partial \theta_{ij}} \cExp{\sum_{m=1}^k \sum_{n=1}^l \theta_{mn}}\\
        &=\Esc{\mP_{kl}}\left(  \frac{\partial \theta_{11}}{\partial \theta_{ij}} + \zeta_{ij}^{kl} \right) \\
       & = \Esc{\mP_{kl}} \left(  - \frac{\partial \psi_{\chi}(\bmtheta)}{\partial \theta_{ij}} + \zeta_{ij}^{kl} \right) \\
       & = \Esc{\mP_{kl}} \left(  -\eta_{ij} + \zeta_{ij}^{kl} \right),
\end{align*}
where we use $\theta_{11}=-\psi_{\chi}(\bmtheta)$ and $\bmeta=\boldsymbol{\nabla}_{\bmtheta}\psi_{\chi}(\bmtheta)$.
The second-order derivative is obtained as follows:
\begin{align*}
        \frac{\partial }{\partial \theta_{vw}}\frac{\partial \mP_{kl}}{\partial \theta_{ij}} &=
        \frac{\partial }{\partial \theta_{vw}}\left\{\chi\left[ \mP_{kl} \right] \left(  -\eta_{ij} + \zeta_{ij}^{kl} \right)\right\} \\
        &=  \frac{\partial \Esc{\mP_{kl}} }{\partial \theta_{vw}} \left(  -\eta_{ij} + \zeta_{ij}^{kl} \right) 
        -\Esc{\mP_{kl}}\frac{\partial \eta_{ij}}{\partial \theta_{vw}} \\
        &=  \frac{\partial \Esc{\mP_{kl}}}{\partial \mP_{kl}} \frac{\partial \mP_{kl} }{\partial \theta_{vw}} \left(  -\eta_{ij} + \zeta_{ij}^{kl} \right) 
        -\Esc{\mP_{kl}}\frac{\partial \eta_{ij}}{\partial \theta_{vw}} \\
        &= \chifp{\mP_{kl}} \Esc{\mP_{kl}} \left(  -\eta_{vw} + \zeta_{vw}^{kl} \right)\left(  -\eta_{ij} + \zeta_{ij}^{kl} \right) - \Esc{\mP_{kl}}\frac{\partial \eta_{ij}}{\partial \theta_{vw}}.
\end{align*}
Because of the normalization condition $\sum_{ij} \mP_{ij}=1$, it holds that
\begin{align*}
    \sum_{kl} \frac{\partial }{\partial \theta_{vw}}\frac{\partial \mP_{kl}}{\partial \theta_{ij}} = 0.
\end{align*}
Thus, we obtain 
\begin{align}\label{eq:2ndgrd}
    \frac{\partial \eta_{vw}}{\partial \theta_{ij}} = \sum_{k=1}^I \sum_{l=1}^J \cEsc{\mP}_{kl}\chifp{\mP_{kl}}
    \Delta_{ij}^{kl}\Delta_{vw}^{kl}.
\end{align}
We can confirm that Equation~\eqref{eq:2ndgrd} provides the second derivative of $D_\chi[\mQ;\mP]$ by differentiating Equation~\eqref{eq:1stderiv} once more again with respect to $\theta_{ij}$, i.e., 
\begin{align*}
    \frac{\partial}{\partial \theta_{ij}} \frac{\partial}{\partial \theta_{vw}} D_\chi[\mQ;\mP]
    =
    \frac{\partial}{\partial \theta_{ij}} \left( \eta_{vw}-\hat{\eta}_{vw} \right) =   \frac{\partial \eta_{vw}}{\partial \theta_{ij}}.
\end{align*}
\end{proof}
We note that Equation~\eqref{eq:2ndgrd} with $\chi(t)=t$ recovers the closed form of the Fisher information matrix that appeared in~\citep{Ghalamkari2023Many}.
\begin{figure*}[t]
  \centering
  \includegraphics[width=0.99\linewidth]{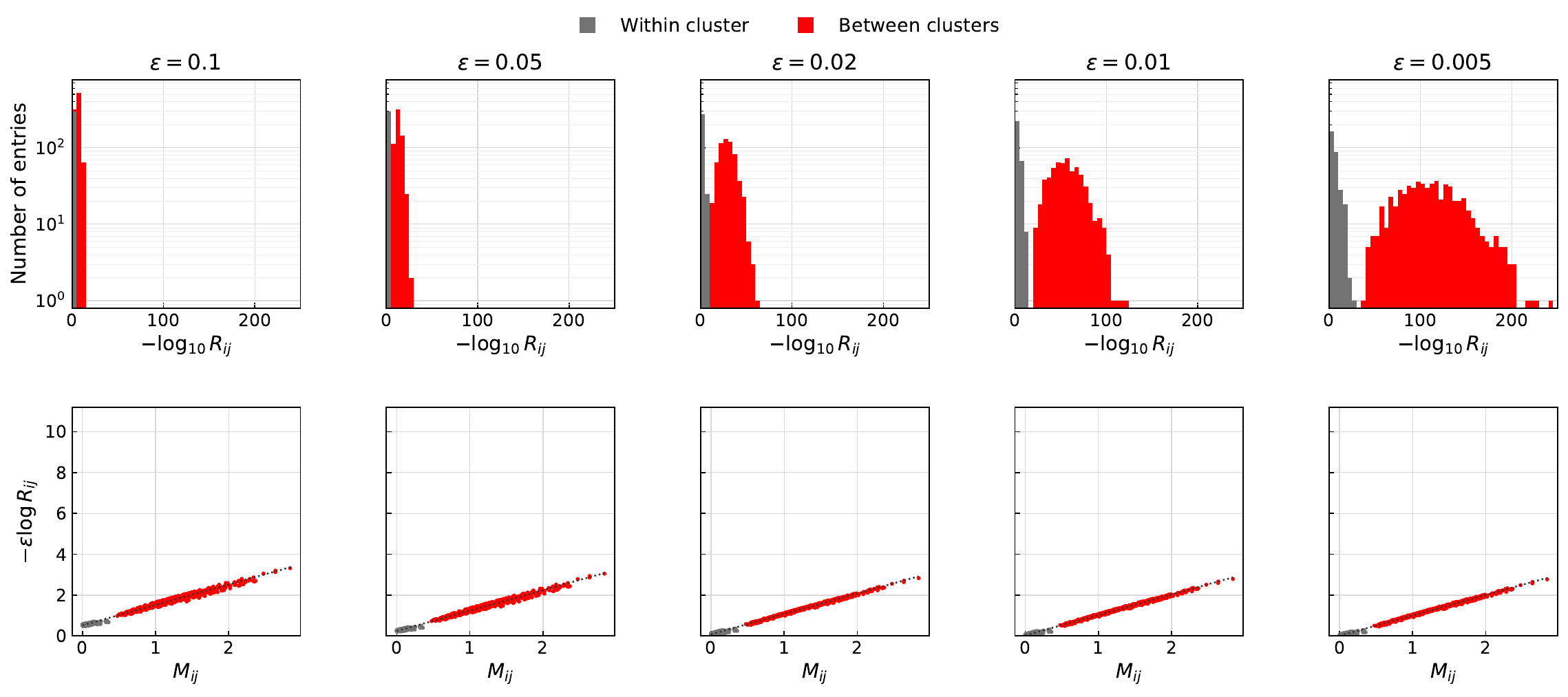}
  \caption{Histograms on a log scale (top) and scatter plots against the cost (bottom) of the entries of the optimal plans $\mR$ obtained by Sinkhorn with various $\reg$. The dotted line shows the exponential decay. }
  \label{fig:sinkhorn_hist}
\end{figure*}

\begin{figure*}[t]
  \centering
  \includegraphics[width=0.99\linewidth]{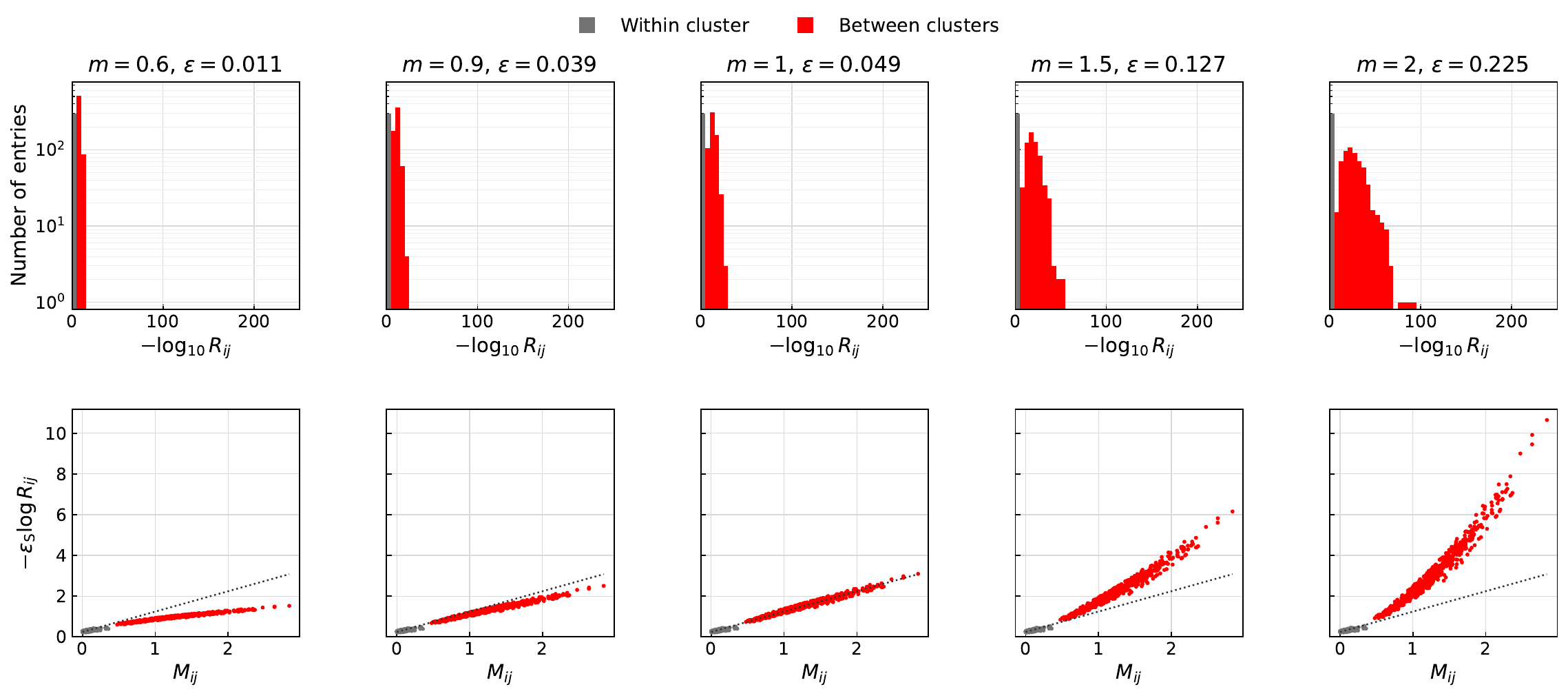}
  \caption{Histograms on a log scale (top) and scatter plots against the cost (bottom) of the entries of the optimal plans $\mR$ obtained by Weibull-ET, with $\reg$ tuned so that the effective local entropy is approximately $8.0$. The dotted line shows the exponential decay that Sinkhorn follows at the same effective local entropy. The entries deviate below this line for $m < 1$ and above it for $m > 1$, showing that $m$ switches the decay from slow to rapid.
}
  \label{fig:mET_hist}
\end{figure*}
\section{Additional experimental evaluation}\label{sec:add_exp}

We show the full results of Figure~\ref{fig:main_exp}(\textbf{c},\textbf{d}) varying the noise level $\tau$ in Figure~\ref{fig:all_mET}. As a sanity check, we can confirm that the larger noise shows a larger centroid drift. The results also reveal two regimes where rapid decay ($m > 1$) and slow decay ($m < 1$) each demonstrate improved robustness.

We examine how the regularization strength $\reg$ shapes the entries of the optimal plan $\mR$, using the three-cluster synthetic dataset of Section~\ref{ref:exp_local_softness}. The top row of Figure~\ref{fig:sinkhorn_hist} shows the histograms of the obtained plan $\mR$ by the Sinkhorn algorithm. A larger $\reg$ confines the entries of $\mR$ to a narrower range of magnitudes.  A smaller $\reg$ widens this range: a small fraction of entries carry most of the mass, while the remaining entries become orders of magnitude smaller. The bottom row of Figure~\ref{fig:sinkhorn_hist} plots $-\reg \log \mR_{ij}$ against the cost $\cost_{ij}$, together with a dotted line of unit slope whose intercept we fit to the within-cluster entries. For every $\reg$, the entries lie close to the dotted line, including the between-cluster entries that we do not use in the fit. This indicates that the plan decays exponentially in the cost, with a vertical scatter caused by $\boldsymbol{\alpha}$ and $\boldsymbol{\beta}$. The result is consistent with Proposition~\ref{prop:decay} with $\chi(t) = t$, which gives $-\reg \log \mR_{ij} = \cost_{ij} + \reg(\alpha_i + \beta_j)$.

We apply the same analysis to the proposed Weibull-ET with various $m$ in Figure~\ref{fig:mET_hist}. For each $m$, we tune $\reg$ so that the effective local entropy is approximately $8.0$. With its two parameters, Weibull-ET keeps the histogram of the within-cluster entries (gray) nearly unchanged through $\reg$
and changes that of the between-cluster entries (red) through $m$, while Sinkhorn, with $\reg$ alone, shifts both histograms together, as seen in Figure~\ref{fig:sinkhorn_hist}. In the bottom row, we scale the vertical axis by $\reg_{\mathrm{S}}=0.05$, the value of the regularization strength at which Sinkhorn attains the same effective local entropy $8.0$, so that the dotted line of unit slope represents the exponential decay of the corresponding Sinkhorn plan. The between-cluster entries bend below the dotted line for $m < 1$ and above it for $m > 1$, indicating slow- and rapid decay, respectively.

\section{Implementation detail}\label{sec:implement_detail}

For $q$-DOT, we contacted the authors and obtained their implementation. 
%and used the default hyperparameters of the L-BFGS optimizer. 
EMD and Sinkhorn were computed using POT~\citep{POT}, a Python library for optimal transport. We implemented Tsallis-ET and Weibull-ET following Algorithm~\ref{alg:ET} with $\delta=10^{-7}$. L$^2$OT was solved with the convex optimization library CVXPY using the OSQP solver~\citep{diamond2016cvxpy}. TEM was implemented based on the pseudocode provided in the original paper~\citep{Amid2024optimal} and TROT was evaluated using the code from the official repository~\footnote{\url{https://github.com/BorisMuzellec/TROT}}. 

For the color-transfer experiments, hyperparameters were tuned as follows. For the regularization parameter $\reg$, we considered 17 values: $ 1, 0.5, 0.1, 0.05, \ldots, 10^{-8}.$
For $q$-ET, we varied
$
q \in \{0.1, 0.2, 0.4, 0.6, 0.8, 0.9, 0.95, 0.98, 1.01, 1.05, 1.075, 1.1, 1.125, 1.15, 1.175, 1.2, 1.3, 1.4, 1.5, 1.6, 1.7, 1.8, 1.9\}.
$
For $m$-ET, we varied
$
m \in \{0.2, 0.3, 0.4, 0.5, 0.6, 0.7, 0.8, 0.9, 1.0, 1.01, 1.05, 1.1, 1.2, 1.3, 1.5, 2.0, 3.0, 5.0, 8.0, 16.0\}.
$
For $q$-DOT, we varied
$
q \in \{0.99, 0.95, 0.9, 0.8, 0.7, 0.6, 0.5, 0.4, 0.3, 0.2, 0.1, 0.01\}.
$
Each algorithm was run for at most $10^6$ iterations, or until its convergence criterion was satisfied. In all experiments, the obtained plans that did not satisfy the feasibility condition were discarded. Feasibility was assessed using the \texttt{get\_err\_marginal} function in \texttt{deform\_ot.py}.

%For Kaniadakis, we implement $\Exp_{\kappa}$ and $\Log_{\kappa}$ using the numerically stable $\operatorname{sinh}$ and $\operatorname{sinh}^{-1}$ forms.
%\begin{align*}
%    \Exp_{\kappa}{t} = \exp
%    \left[
%    \frac{\operatorname{sinh}^{-1}(\kappa t)}
%    { \kappa}\right], \quad
%    \Log_{\kappa}{t} = \frac{\sinh{\left[\kappa \log{t}\right]}}{\kappa}
%\end{align*}
%We evaluate $\operatorname{sinh}$ and $\operatorname{sinh}^{-1}$ using NumPy (\texttt{np.arcsinh}, \texttt{np.sinh}) in float64, which avoids catastrophic cancellation present in the direct power form.

\paragraph{Details for visualization}
In Figures~\ref{fig:headline}(b), \ref{fig:softness}, and~\ref{fig:sinkhorn_plans}, gray and red lines represent within- and between-cluster entries $\mR_{ij}$ of the transport plan, respectively. Gray line width is proportional to the value $\mR_{ij}$. Since between-cluster entries span several orders of magnitude, we draw red lines only for entries above a threshold and scale their width linearly with $\log \mR_{ij}$. All three figures share the width normalization and the threshold, set to $10^{-6}$ times the largest between-cluster entry over all their panels; thus, line widths are comparable across figures.

\paragraph{Additional information for reproducibility} 
All experiments were executed on a university HPC cluster using CPU-only resources. Each compute node runs Linux kernel 6.1.149 on an \verb|x86_64| platform and is equipped with dual AMD EPYC 9354 processors, providing a total of 64 CPU cores and 251 GB of RAM. The implementation was written in Python 3.12.7, and jobs were scheduled via \verb|bsub|. Peak memory usage during all experiments remained below 19 GB. To support reproducibility, the complete source code is included in the supplementary material.

\end{document}